\documentclass[11pt]{article}

\usepackage{acl}

\usepackage{microtype}
\usepackage{graphicx}
\graphicspath{{figures/}}
\usepackage{pifont}
\usepackage{colortbl}
\usepackage{array}
\usepackage{booktabs}
\usepackage{amsmath,amssymb}
\usepackage{multirow}
\usepackage{multicol}
\usepackage{siunitx}
\usepackage{hyperref}
\usepackage{xcolor}
\usepackage{placeins}
\usepackage{caption}
\usepackage{tabularx}
\usepackage{ragged2e}
\usepackage{dblfloatfix}
\usepackage{float}
\usepackage{needspace}
\usepackage{makecell}

\makeatletter
\AtBeginDocument{%
  \setlength{\@fptop}{0pt}%
  \setlength{\@fpsep}{12pt}%
  \setlength{\@fpbot}{0pt plus 1fil}%
  \setlength{\@dblfptop}{0pt}%
  \setlength{\@dblfpsep}{12pt}%
  \setlength{\@dblfpbot}{0pt plus 1fil}%
}
\makeatother

\newsavebox{\fitbox}
\long\def\fitcol#1{%
  \sbox{\fitbox}{#1}%
  \ifdim\wd\fitbox>\columnwidth
    \resizebox{\columnwidth}{!}{\usebox{\fitbox}}%
  \else
    \usebox{\fitbox}%
  \fi}
\long\def\fitpage#1{%
  \sbox{\fitbox}{#1}%
  \ifdim\wd\fitbox>\textwidth
    \resizebox{\textwidth}{!}{\usebox{\fitbox}}%
  \else
    \usebox{\fitbox}%
  \fi}

\definecolor{perfA}{HTML}{E8C4C4}   % poor
\definecolor{perfB}{HTML}{F2DCC0}
\definecolor{perfC}{HTML}{EDE7C2}
\definecolor{perfD}{HTML}{D9E5C4}
\definecolor{perfE}{HTML}{C2DCC4}   % strong
\newcommand{\pc}[1]{%
  \ifdim #1pt>95pt\cellcolor{perfE}#1\else
  \ifdim #1pt>90pt\cellcolor{perfD}#1\else
  \ifdim #1pt>80pt\cellcolor{perfC}#1\else
  \ifdim #1pt>50pt\cellcolor{perfB}#1\else
  \cellcolor{perfA}#1\fi\fi\fi\fi}

\title{DeBERTa-ConPara: Attack-Aware and Deployment-Realistic Detection of AI-Generated Text}

\author{
  Mohamed Mady\textsuperscript{1,2} \quad
  Yupei Li\textsuperscript{3} \quad
  Johannes Reschke\textsuperscript{2} \quad
  Bj\"{o}rn W. Schuller\textsuperscript{1,3} \\[0.4em]
  \textsuperscript{1}CHI, Chair of Health Informatics,
  Technical University of Munich, Germany \\
  \textsuperscript{2}Smart Embedded Systems Lab,
  OTH Regensburg, Germany \\
  \textsuperscript{3}GLAM, Group on Language, Audio \& Music,
  Imperial College London, UK \\[0.3em]
  \small\texttt{mohamed.mady@tum.de} \quad
  \texttt{yupei.li22@imperial.ac.uk} \quad
  \texttt{johannes.reschke@oth-regensburg.de} \quad
  \texttt{schuller@tum.de}
}

\begin{document}

\maketitle

\begin{abstract}
Robust detection of AI-generated text under deployment conditions is
challenging: distribution shifts across domains and generators, adversarial
perturbations of the input surface, and the absence of target-domain labels
for threshold calibration all degrade detectors that perform well in-domain.

We present \textbf{DeBERTa-ConPara}, a deployment-oriented detector combining
attack-aware Unicode preprocessing with a contextual transformer encoder
trained over HC3~Plus, M4, MAGE and RAID. Our central finding is that preprocessing acts in opposite
directions depending on where it is applied: normalising the \emph{training}
corpus deduplicates it, collapsing $35.4\,\%$ of RAID rows into copies of
their clean siblings and deleting the adversarial supervision, whereas
normalising at \emph{inference} is an effective defence. A factorial varying
the two placements independently identifies raw training with normalised
inference as the best configuration, reaching \textbf{99.61\,\% AUROC},
\textbf{99.01\,\%} TPR@5\,\% FPR and \textbf{96.57\,\%} TPR@1\,\% FPR on the
official RAID hidden test, alongside \textbf{93.14\,\%} average balanced
accuracy across HC3~Plus and MAGE under a fixed threshold. The gain is
confined to two of twelve attack classes: homoglyph and zero-width-space
insertion rise from $11.05\,\%$ and $1.12\,\%$ to $96.98\,\%$. The same
signature reproduces in a zero-shot detector of different architecture,
showing the effect belongs to the attacks rather than to our model.
We additionally report two negative results: semantic-invariance augmentation
through paraphrasing and supervised contrastive learning (\textsc{ConPara})
does not improve the best configuration, and the handcrafted feature-fusion
branch is inert in distribution and harmful outside it.
\end{abstract}

\section{Introduction}
\label{sec:introduction}

Recent advances in large language models (LLMs) have dramatically improved the fluency and realism of AI-generated text, making machine-written content increasingly difficult to distinguish from human writing across domains such as education, journalism, scientific publishing, and online communication~\citep{Achiam23-GPT4,Gemini23,StanfordAIIndex25}. While these capabilities enable powerful generative applications, they also introduce major concerns including misinformation propagation, academic dishonesty, synthetic reviews, and coordinated inauthentic behaviour~\citep{Turnitin24-Wrapped,Chakraborty24-Position, Goldstein23-GenAI, Stokel-Walker22-ChatGPT}. Consequently, robust AI-generated text detection has become a critical challenge for trustworthy deployment of generative AI systems.

Existing detection approaches fall into three families, surveyed in Section~\ref{sec:related_work}: supervised classifiers~\citep{li2025discourse}, zero-shot statistical detectors~\citep{Bao24-FastDetectGPT}, and watermarking~\citep{Wouters24-Watermarks}. Watermarking requires cooperation from the generator and leaves already-generated text untouched, so post-hoc detection remains the practical setting and is the one we address. Although many detectors achieve strong in-domain performance, robustness degrades substantially under unseen generators, domain shifts, decoding variations, and adversarial perturbations~\citep{Dugan24-RAID,Chakraborty24-Position}. Lightweight surface-form attacks such as homoglyph substitutions and zero-width Unicode characters can severely disrupt tokenisation-sensitive pipelines and distort linguistic and statistical feature distributions while preserving semantic content.

Despite recent progress, two deployment challenges remain insufficiently addressed. 
First, most evaluation protocols implicitly assume access to target-domain labels by independently tuning decision thresholds for each dataset or domain at test time, an assumption that rarely holds in practical deployment. Second, many adversarial attacks operate primarily at the character and formatting level, exploiting weaknesses in Unicode handling and tokenisation before semantic modelling begins. Existing detectors therefore often focus on representation learning while overlooking vulnerabilities in preprocessing and tokenisation pipelines.

We introduce \textbf{DeBERTa-ConPara}, a deployment-oriented detector built on \texttt{microsoft/deberta-v3-large}~\citep{He21-DeBERTaV3} that combines attack-aware preprocessing with a multi-dataset adversarial training corpus.

On the official RAID hidden-test benchmark, DeBERTa-ConPara reaches $99.01\,\%$ TPR@5\,\% FPR while sustaining $93.14\,\%$ average balanced accuracy across HC3 Plus and MAGE under a fixed threshold. Several leaderboard systems score higher on RAID alone; our claim is the combination, restricted to the English domains, generators and attacks evaluated here.

Our contributions are:

\begin{itemize}
    \item A controlled $2\times2\times2$ factorial, all eight cells submitted
    to the official RAID hidden test, showing that Unicode preprocessing
    defends at inference time but removes adversarial supervision when
    applied to the training corpus.

    \item An attribution of that robustness to two of twelve attack classes,
    reproduced in a zero-shot detector of different architecture: the effect
    belongs to the attacks, not to our model.

    \item A multi-dataset curation strategy that keeps cross-dataset
    generalisation alongside that robustness, where systems tuned to one
    benchmark do not.

    \item Two documented negative results: \textsc{ConPara} augmentation does
    not improve the best configuration, and the feature-fusion branch is inert
    in distribution and harmful out of it.
\end{itemize}

\section{Related Work}
\label{sec:related_work}

AI-generated text detection has rapidly evolved alongside advances in large language models (LLMs), leading to a broad spectrum of detection paradigms including zero-shot statistical methods, supervised detectors, watermarking approaches, and adversarially robust detection frameworks~\citep{Wu25-Survey}. Despite substantial progress, existing methods continue to struggle under realistic deployment conditions involving unseen generators, domain shifts, and adversarial perturbations.

\subsection{Zero-Shot and Statistical Detection}

Zero-shot detectors identify machine-generated text without task-specific supervised training by exploiting statistical irregularities in language-model outputs. Representative approaches include GLTR~\citep{Gehrmann19-GLTR}, DetectGPT~\citep{Mitchell23-DetectGPT}, Fast-DetectGPT~\citep{Bao24-FastDetectGPT}, and Binoculars~\citep{Hans24-Binoculars}, which analyse token distributions, probability curvature, or cross-model likelihood patterns. These methods are computationally lightweight and avoid costly supervised training, making them attractive for rapid deployment and unseen-generator evaluation. Feature-based approaches such as Ghostbuster~\citep{Verma23-Ghostbuster} additionally leverage stylometric and probabilistic signals for cross-domain generalisation. More recent work targets the proxy-model mismatch that arises when the scoring model differs from the unknown generator: SurpMark~\citep{Chen26-SurpMark} summarises a passage by the dynamics of its token surprisals and scores it against fixed human and machine references, avoiding the per-input contrastive generation that makes curvature-based methods expensive.

Although zero-shot methods avoid costly labeled training data and may generalise to unseen generators, prior work shows substantial degradation under domain shifts, decoding variations, and adversarial rewriting attacks~\citep{Chakraborty24-Position,Dugan24-RAID,Sadasivan23-CanAI}. These limitations become particularly pronounced under fixed-threshold evaluation where target-domain calibration is unavailable.

\subsection{Supervised Detection Methods}

Supervised detectors fine-tune pretrained language encoders on labeled human-versus-AI text pairs. Early work leveraged transformer architectures such as BERT~\citep{Devlin19-BERT} and RoBERTa~\citep{Liu19-RoBERTa} trained on benchmarks including HC3~\citep{Guo23-HC3} and HC3 Plus~\citep{Su23-HC3Plus}. More recent approaches improve robustness through adversarial training, representation learning, and generator-aware adaptation. For example, RADAR~\citep{Hu23-RADAR} trains detectors adversarially against paraphrasing attacks. RepreGuard~\citep{Chen25-RepreGuard} instead detects machine-generated text from hidden representation patterns rather than output probabilities, and MGT-Prism~\citep{Liu25-MGTPrism} aligns spectral features to improve generalisation across domains.

Recent studies further show that paraphrasing alone can evade many existing detectors~\citep{Krishna23-Paraphrase}, motivating both adversarial training approaches and semantic-invariance augmentation strategies.

Although supervised methods often achieve near-ceiling in-domain accuracy, large-scale benchmarks reveal substantial degradation under cross-generator and cross-domain evaluation~\citep{Wang24-M4,Li24-MAGE,Dugan24-RAID}. Many detectors implicitly overfit to generator-specific artefacts, stylistic cues, or decoding signatures that fail to generalise under distribution shift.

\subsection{Watermarking-Based Detection}

Watermarking approaches embed detectable signals directly during text generation rather than relying on post-hoc analysis~\citep{Kirchenbauer23-Watermark}. Subsequent work explored probabilistic and cryptographic watermarking strategies for LLM outputs~\citep{Wouters24-Watermarks,Zhang24-StrongWatermarking}. However, watermark robustness degrades substantially under paraphrasing and semantic-preserving rewriting attacks~\citep{Zhang24-StrongWatermarking}.

\subsection{Domain Generalisation and Adversarial Robustness}

Several recent works explicitly target robustness under domain shift and adversarial perturbations. EAGLE~\citep{Bhattacharjee24-EAGLE} employs domain-adversarial and contrastive learning objectives to encourage generator-invariant representations, while MGT-Prism~\citep{Liu25-MGTPrism} explores spectral-alignment techniques for improving robustness across domains and generators.

Recent benchmark efforts further highlight the importance of deployment-realistic evaluation. DetectRL~\citep{Wu24-DetectRL} showed that many detectors degrade substantially under natural distribution shifts and adversarial conditions. Similarly, RAID~\citep{Dugan24-RAID} established a large-scale adversarial robustness benchmark covering paraphrasing, homoglyph attacks, whitespace perturbations, zero-width Unicode characters, and structural manipulations. Most existing approaches primarily address semantic attacks through adversarial training or paraphrase augmentation, while comparatively little attention has been paid to explicit mitigation of surface-form perturbations before tokenisation.

\section{Datasets}
\label{sec:datasets}

DeBERTa-ConPara is trained and evaluated on four complementary large-scale benchmarks spanning semantic invariance, cross-domain generalisation, fluent multi-domain writing, and adversarial robustness: HC3 Plus~\citep{Su23-HC3Plus}, M4~\citep{Wang24-M4}, MAGE~\citep{Li24-MAGE}, and RAID~\citep{Dugan24-RAID}. Together they expose the detector to substantial variability across generators, domains, writing styles and attacks.

The training corpus holds $1{,}549{,}358$ rows, balanced exactly between
human and AI-generated text, with $172{,}826$ further rows held out for
validation. Table~\ref{tab:datasets-overview} gives its composition;
Appendix~\ref{app:m4_details} describes the human-written filler drawn from M4
to achieve that balance. Two subsets, PeerRead and WikiHow, are generated by
us rather than drawn from a published benchmark, and they matter out of
proportion to their size: the four public benchmarks use generators current
in 2023, so these $43{,}059$ rows are the corpus's only exposure to newer
models.

Detailed dataset descriptions, split configurations, attack categories, and per-domain statistics are provided in Appendix~\ref{app:dataset_overview}.

\begin{table}[!h]
\centering
\small
\fitcol{%
\begin{tabular}{lccc}
\toprule
Dataset & Rows & Domains & AI generators \\
\midrule
RAID                    & 530{,}534 & 8 & 11 \\
M4 (human filler)       & 438{,}305 & 5 & \phantom{0}0 \\
MAGE                    & 319{,}071 & 7 & 27 \\
HC3 Plus                & 148{,}040 & 4 & \phantom{0}1 \\
M4                      & \phantom{0}70{,}349 & 5 & \phantom{0}8 \\
PeerRead$^{\dagger}$    & \phantom{0}28{,}047 & 1 & \phantom{0}5 \\
WikiHow$^{\dagger}$     & \phantom{0}15{,}012 & 1 & \phantom{0}7 \\
\midrule
Total                   & 1{,}549{,}358 & & \\
\bottomrule
\end{tabular}}

\caption{Composition of the $1.55$M-row training corpus. RAID contributes
$398{,}834$ AI-generated and $131{,}700$ human rows; the M4 human filler
carries no AI text. Domain and generator counts characterise each benchmark,
not the sampled subset. A further $172{,}826$ rows are held out for validation.
$^{\dagger}$~Generated by us: PeerRead with ChatGPT, LLaMA and LLaMA-2-chat at
7B, 13B and 70B; WikiHow with Claude, GPT-4o, Grok, Qwen-14B, LLaMA-3-8B,
Mistral-7B and Cohere.}
\label{tab:datasets-overview}
\end{table}

\begin{figure*}[!t]
\centering
\includegraphics[width=0.97\textwidth]{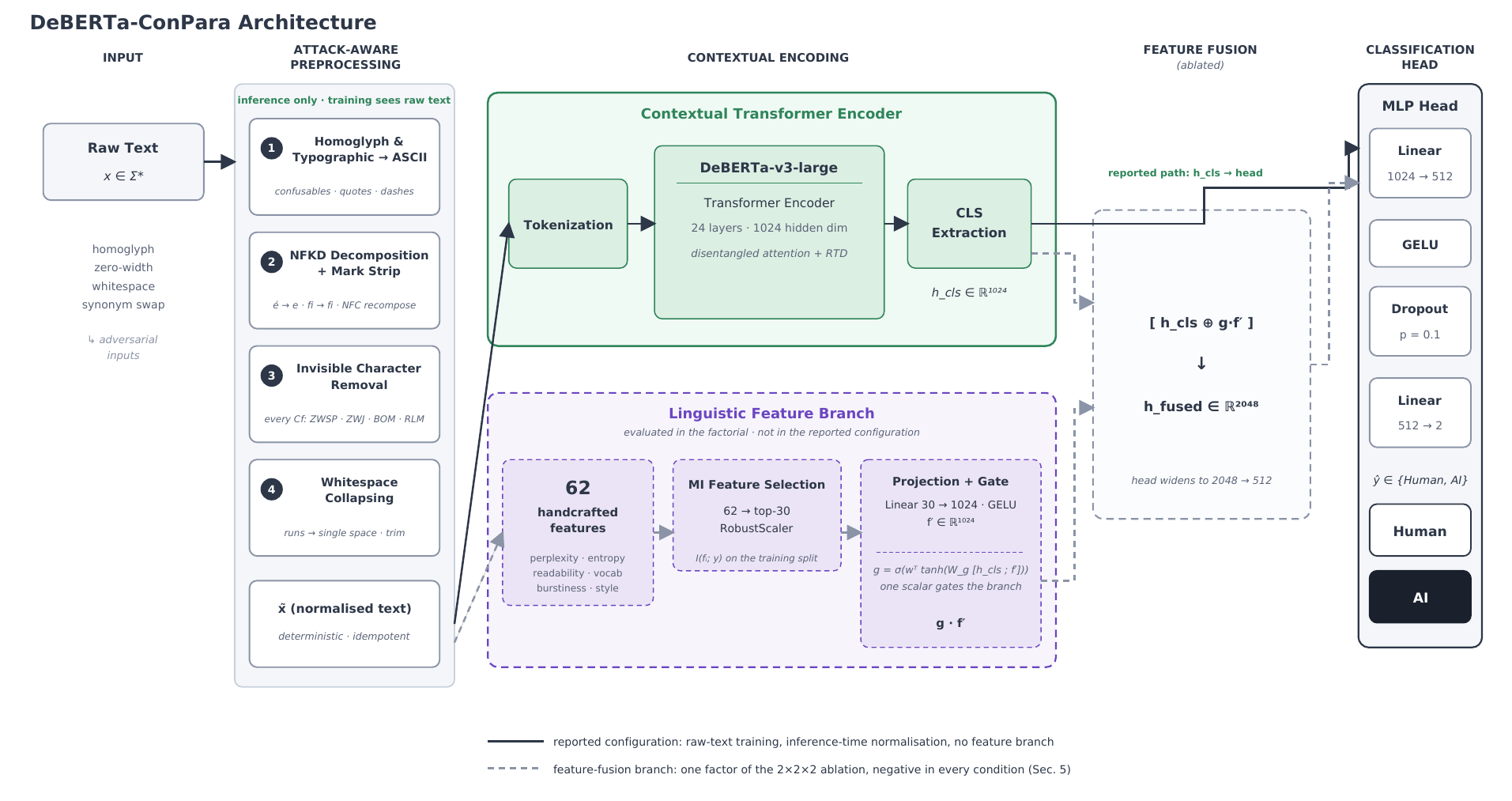}

\caption{
DeBERTa-ConPara architecture. Solid components form the reported
configuration: attack-aware preprocessing applied at inference only, a
DeBERTa-v3-large encoder, and a two-layer head on the classification
embedding. Dashed components are the linguistic feature branch and its gated
fusion, one factor of the $2\times2\times2$ ablation in
Section~\ref{sec:experiments}, which is negative in every condition and is
not part of the reported model.
}

\label{fig:architecture}
\end{figure*}

\section{Method}
\label{sec:method}

We propose \textbf{DeBERTa-ConPara}, a deployment-oriented detector combining
attack-aware preprocessing with contextual transformer representations. We
also describe a linguistic feature-fusion branch and semantic-invariance
augmentation (\textsc{ConPara}), both central to the submitted version and
both reported as negative results in Section~\ref{sec:experiments}. The
configuration we report trains on raw text, normalises at inference, and
carries no feature branch.

\subsection{Overall Architecture}
\label{subsec:architecture}

Figure~\ref{fig:architecture} presents an overview of DeBERTa-ConPara. The input text, normalised at inference by the pipeline of Section~\ref{subsec:preprocessing}, is encoded by a contextual transformer whose classification embedding feeds the head; the linguistic feature branch shown dashed is the ablated component described next.

\paragraph{Contextual Transformer Encoder.}

We employ \texttt{microsoft/deberta-v3-large}~\citep{He21-DeBERTaV3} as the
contextual encoder backbone. The submitted version used
\texttt{deberta-v3-base}, appropriate for the smaller corpus then available.
With the corpus extended to $1.55$M rows we
re-ran the comparison with every other factor held fixed (identical data,
hyperparameters and seed, five epochs), and the larger backbone is
consistently better: balanced accuracy at the optimal threshold
rises from $94.56\,\%$ to $96.58\,\%$ and AUROC from $0.98655$ to $0.99328$.
We therefore report the larger backbone throughout; backbone change and
corpus extension were adopted together. DeBERTa-v3 combines disentangled attention mechanisms with ELECTRA-style replaced-token detection (RTD) pretraining~\citep{Clark20-ELECTRA}, providing strong contextual representations and robust transfer performance across diverse language understanding tasks.

The final hidden-state representation corresponding to the special classification token is extracted as a $1024$-dimensional contextual embedding, denoted $\mathbf{h}_{\text{cls}}$ below; in the reported configuration it is the only input to the classifier.

\paragraph{Linguistic Feature Encoder.}

In parallel, we extract 62 handcrafted linguistic and statistical features designed to capture stylometric properties that may be weakly represented in transformer embeddings. The feature categories include probabilistic, stylometric, readability, and lexical-diversity signals.

To reduce redundancy and improve robustness under distribution shift, we select the top $k=30$ features using Mutual Information~\citep{CoverThomas06-InfoTheory} on the training set; the full feature list and the selection procedure are in Appendix~\ref{app:features}.

Selected features are normalised using RobustScaler, projected to the encoder
width, and combined with the contextual representation through a learned
gate.

Formally, given the selected feature vector $\mathbf{f}(x)\in\mathbb{R}^{30}$
and the encoder's classification embedding
$\mathbf{h}_{\text{cls}}\in\mathbb{R}^{1024}$, the branch computes

\begin{equation}
\begin{aligned}
\mathbf{f}' &= \mathrm{Dropout}\big(\phi(W_p\,\mathbf{f}(x))\big),
   \quad W_p\in\mathbb{R}^{1024\times 30},\\[2pt]
g &= \sigma\!\Big(\mathbf{w}^{\!\top}
   \tanh\!\big(W_g\,[\mathbf{h}_{\text{cls}};\mathbf{f}']\big)\Big),\\[2pt]
\mathbf{h} &= \big[\,\mathbf{h}_{\text{cls}}\;;\; g\,\mathbf{f}'\,\big]
   \in \mathbb{R}^{2048},
\end{aligned}
\label{eq:feature_attention}
\end{equation}
where $\phi$ is GELU and $\sigma$ the logistic function. Note that $g$ is a
\emph{single scalar}: the gate admits or suppresses the projected feature
vector as a whole rather than reweighting its coordinates. The gated features
are concatenated with the $1024$-dimensional DeBERTa representation, forming a
\textbf{2048-dimensional fused representation} passed to the classifier.

\paragraph{Binary Classification Head.}

The fused representation is processed using a lightweight two-layer MLP:

\begin{equation}
\begin{aligned}
2048
&\rightarrow 512
\rightarrow \mathrm{GELU} \\
&\rightarrow \mathrm{Dropout}(0.1)
\rightarrow 2.
\end{aligned}
\label{eq:mlp}
\end{equation}

In the configuration we report, which carries no feature branch, the
classification embedding is passed to the same head directly, giving
$1024 \rightarrow 512 \rightarrow \mathrm{GELU} \rightarrow
\mathrm{Dropout}(0.1) \rightarrow 2$. The released checkpoint implements
exactly this head.

\subsection{Attack-Aware Preprocessing Pipeline}
\label{subsec:preprocessing}

We introduce an attack-aware preprocessing pipeline applied prior to tokenisation to improve robustness against Unicode-based and formatting-based adversarial perturbations such as homoglyph substitutions and zero-width character insertions.

The pipeline applies, in order: targeted homoglyph substitution using a curated
confusables table spanning Cyrillic, Greek, fullwidth and stylised Latin
variants; typographic normalisation of smart quotes, dashes and fractions to
ASCII; NFKD decomposition with combining-mark removal, re-composed with NFC;
removal of every Unicode format character (category Cf), which covers
zero-width and directional marks; and whitespace collapsing. Homoglyphs are
handled explicitly because they are canonically distinct and decomposition
alone does not repair them. Full stage descriptions are given in
Table~\ref{tab:preprocessing_pipeline}.

The pipeline is deterministic and idempotent, with negligible overhead.

Where it is applied, however, matters more than whether it is applied: Section~\ref{subsec:ablation} shows that normalising
the training corpus removes the adversarial supervision that corpus provides,
whereas normalising at inference is an effective defence. The configuration we
report therefore trains on raw text and normalises only at inference, in
contrast to the submitted version, which applied the pipeline at both stages.

\subsection{Semantic-Invariance Augmentation}

To improve robustness against paraphrasing and semantic-preserving rewriting attacks, we investigate semantic-invariance augmentation through paraphrase generation and supervised contrastive learning.

\paragraph{Paraphrase Augmentation.}

For selected training samples, paraphrases are generated using \texttt{flan-t5-base}~\citep{Chung22-FlanT5}. Generated paraphrases are filtered using a cosine-similarity threshold of $\geq 0.85$, computed over sentence embeddings from a pretrained encoder~\citep{Krishna23-Paraphrase}. This threshold empirically retains paraphrases that preserve the original semantic intent while introducing sufficient lexical and syntactic variation; lower thresholds admitted meaning-altering rewrites, while higher thresholds eliminated useful diversity in preliminary experiments.

\paragraph{Supervised Contrastive Learning.}

We additionally experiment with supervised contrastive learning (SupCon)~\citep{Khosla20-SupCon} using paraphrased positive pairs and opposite-class negatives. The training objective combines weighted cross-entropy with SupCon loss weighted by $\lambda \in \{0.05, 0.1, 0.2\}$, selected via ablation (Equation~\ref{eq:total_loss}, Appendix~\ref{app:conpara}).

\subsection{Training Configuration}

Training uses AdamW~\citep{Loshchilov17-AdamW} with weight decay $0.01$, peak
learning rate $2\times10^{-5}$, batch size $32$ and maximum sequence length
$512$, under a OneCycleLR schedule with cosine annealing~\citep{Smith19-OneCycleLR};
the reported checkpoint is the epoch with the best validation balanced
accuracy. Full hyperparameters are listed in Appendix~\ref{app:hyperparameters}.
The development sequence that preceded the factorial, from Cohere-focused
oversampling through adversarial data scaling to a generator-difficulty
weighted loss, is recorded in Appendix~\ref{app:versions}; the controlled
comparisons of Section~\ref{sec:experiments} supersede it.

\section{Experiments and Results}
\label{sec:experiments}

We evaluate DeBERTa-ConPara under a deployment-realistic fixed-threshold protocol. Following the official RAID hidden-test evaluation procedure~\citep{Dugan24-RAID}, we report TPR@5\,\% FPR, TPR@1\,\% FPR, and AUROC. In addition to RAID leaderboard evaluation, we assess cross-dataset generalisation on HC3 Plus and MAGE, perform external academic-domain validation, analyse multi-seed stability, and compare against strong zero-shot and supervised baselines.

%%%%%%%%%%%%%%%%%%%%%%%%%%%%%%%%%%%%%%%%%%%%%%%%%%%%%%%%%%%%
\subsection{Evaluation Protocol}
\label{subsec:protocol}

We adopt a \textbf{fixed-threshold evaluation protocol} designed to better reflect realistic deployment conditions. A single decision threshold $\tau^*$ is calibrated on source validation data by maximising balanced accuracy~\citep{Youden50-Index} and is then held constant across all target distributions.

Unlike standard evaluation settings that independently re-tune thresholds for each target dataset, the fixed-threshold setting exposes operating-point trade-offs and generator-specific failure modes that emerge when target-domain labels are unavailable. This setup more closely reflects practical deployment scenarios encountered in moderation systems, educational platforms, and misinformation-filtering pipelines.

For RAID, we follow the official hidden-test leaderboard protocol~\citep{Dugan24-RAID}, where thresholds are calibrated to achieve exactly 5\,\% false positive rate (FPR) on human-written text within each domain. Performance is evaluated using TPR@5\,\% FPR (primary metric), TPR@1\,\% FPR, and AUROC~\citep{Fawcett06-ROC}.

%%%%%%%%%%%%%%%%%%%%%%%%%%%%%%%%%%%%%%%%%%%%%%%%%%%%%%%%%%%%
\subsection{Ablation Study: Preprocessing at Training Time and at Inference Time}
\label{subsec:ablation}

Preprocessing can be applied when the training corpus is built, when a test
document is scored, or both. The two are distinct interventions that, as we
show below, act in opposite directions, so we vary them independently.

\paragraph{Normalisation deduplicates the training corpus.}
RAID contains $13{,}371$ unique human articles under twelve attack variants.
Three attacks, homoglyph substitution, zero-width-space insertion and
whitespace manipulation, are pure character substitutions, so Unicode
normalisation returns the original text exactly and the attacked row becomes a
byte-identical duplicate that deduplication discards. On a matched
$1{,}892$-row sample, the normalised pipeline retains $1{,}223$ distinct
texts against $1{,}892$ for the raw one: normalisation
collapses $35.4\,\%$ of RAID and removes all three attack classes from the
corpus. Raw-text training is therefore implicitly adversarial;
normalising at training time deletes that supervision.

\paragraph{An eight-cell factorial.}
Crossing preprocessing at training time with preprocessing at inference time,
with and without the feature-fusion branch, gives eight configurations,
trained from the same source corpus with identical hyperparameters and a fixed
seed. The training factor is not normalisation alone: because normalisation
collapses attacked rows into duplicates, the four \emph{n}-trained cells see
roughly $35\,\%$ fewer RAID rows than the \emph{r}-trained cells. Tr should
therefore be read as normalisation \emph{and} the deduplication it causes.
All eight cells were submitted to the RAID hidden test, whose labels we never
observe; Table~\ref{tab:factorial} reports them.

\begin{table}[t]
\centering\small
\setlength{\tabcolsep}{3pt}
\fitcol{%
\begin{tabular}{@{}ccc|ccc|cc@{}}
\toprule
\textbf{Ft} & \textbf{Tr} & \textbf{Inf} &
\textbf{AUR} & \textbf{T5} & \textbf{T1} &
\textbf{T1$_{\text{cl}}$} & \textbf{pen} \\
\midrule
\ding{51} & n & n & \pc{99.48} & \pc{98.84} & \pc{87.45} & \pc{88.83} &  1.38 \\
\ding{51} & n & r & \pc{97.32} & \pc{92.76} & \pc{75.59} & \pc{88.38} & 12.79 \\
\ding{51} & r & r & \pc{99.39} & \pc{98.83} & \pc{93.64} & \pc{95.77} &  2.13 \\
\ding{51} & r & n & \pc{99.51} & \pc{99.02} & \pc{94.72} & \pc{95.28} &  0.56 \\
\midrule
\ding{55} & n & n & \pc{99.54} & \pc{98.50} & \pc{94.52} & \pc{95.44} &  0.92 \\
\ding{55} & n & r & \pc{95.77} & \pc{86.14} & \pc{79.48} & \pc{95.28} & 15.80 \\
\ding{55} & r & r$^\dagger$ & \pc{99.35} & \pc{98.39} & \pc{93.78} & \pc{97.17} &  3.39 \\
\ding{55} & \textbf{r} & \textbf{n} & \pc{99.61} & \pc{99.01} & \pc{96.57}
          & \pc{96.98} & \textbf{0.41} \\
\bottomrule
\end{tabular}}
\caption{The complete $2\times2\times2$ factorial on the $672{,}000$-row RAID
hidden test. Ft = feature-fusion branch; Tr/Inf = normalisation at training
and inference (n/r); AUR = AUROC; T5/T1 = TPR at $5\,\%$/$1\,\%$ FPR;
T1$_{\text{cl}}$ = TPR@1\,\% on unattacked rows; pen = attack penalty
T1$_{\text{cl}}-$T1. Shading is a fixed scale shared with
Appendix~\ref{app:factorial_full}. The last row is the configuration we
report. $^\dagger$~This cell is a vanilla fine-tuned DeBERTa-v3-large, with
neither attack-aware preprocessing nor handcrafted features; it is the
supervised baseline requested during review.}
\label{tab:factorial}
\end{table}

The $-$F\,r$\to$r cell is a vanilla fine-tuned DeBERTa-v3-large with neither
attack-aware preprocessing nor handcrafted features, and is the supervised
baseline requested during review. Every comparison below is therefore on
labels we never observe. Three main effects follow, each consistent
across the conditions it is averaged over.

\emph{Inference-time normalisation always helps}, by $+1.08$ to $+15.04$\,pp
of TPR@1\,\% FPR. The gain is largest exactly where the model was trained on
normalised text and so never met an attack ($+11.86$ and $+15.04$), and small
where raw training already supplied that exposure ($+1.08$ and $+2.79$).

\emph{Raw-corpus training always helps}, by $+2.05$ to $+18.05$\,pp, and most
when inference is also raw ($+14.30$, $+18.05$), where it is the only
defence present.

\emph{The feature branch never helps.} It costs $7.07$, $3.89$, $1.85$ and
$0.14$\,pp in the four conditions: negative in all of them, and most expensive
in the fully normalised setting the submitted version used. That aggregate
conceals a far larger domain-specific effect, which we return to below.

In the attack-penalty column, six of eight cells lose under $3.4$\,pp; the two that lose $12.79$ and
$15.80$\,pp are precisely the two trained on normalised text and evaluated on
raw. Attack robustness is therefore not a property of the model but of the
agreement between its training corpus and its inference pipeline.

\paragraph{The feature branch fails on one domain.}
Per-domain results (Appendix~\ref{app:factorial_full},
Table~\ref{tab:fact_domain_1}) show that the branch's cost is not spread
evenly. On seven of eight RAID domains it changes TPR@1\,\% FPR by at most a
few points, and on abstracts it is slightly beneficial. On \emph{poetry} it
costs $49.8$\,pp: $38.1\,\%$ with the branch against $87.9\,\%$ without, in the
otherwise identical fully normalised configuration. The penalty falls to
$7.8$\,pp once training uses raw text, so adversarial supervision partly
compensates for it, but poetry remains the domain where the branch is most
damaging in every condition.

\paragraph{The effect is confined to two attack classes.}
For ten of the twelve RAID categories the \emph{penalty relative to
unattacked text} is within roughly a point across every configuration;
the aggregate differences between cells are driven by homoglyph substitution
and zero-width-space insertion. Absolute rates still vary by attack, with
paraphrase ranging from $75.9$ to $90.4\,\%$, but that variation is
common to all eight cells. Without
inference-time normalisation a normalised-trained model detects $11.05\,\%$ of
homoglyph and $1.12\,\%$ of zero-width-space attacks at $1\,\%$ FPR; with it,
both reach $96.98\,\%$, identical to its unattacked score. Normalisation
does not reduce these attacks, it eliminates them.

Neither placement alone suffices: raw training lifts homoglyph detection to
$85.78\,\%$ and zero-width to $73.80\,\%$; inference-time normalisation
carries both to $96.98\,\%$.

\Needspace{3\baselineskip}
The mismatch penalty is also asymmetric, and the same two-attack signature reproduces in
Fast-DetectGPT~\citep{Bao24-FastDetectGPT}, which has a different tokeniser and
no trained head (Appendix~\ref{app:factorial}).

\paragraph{The feature-fusion branch.}
Four interventions applied to one checkpoint on an identical $20{,}000$-row
sample isolate the 30-feature branch; zeroing the projected representation and
forcing the gate to zero are mathematically equivalent and produce
bit-identical margins, confirming they act where intended. Removing the branch
shifts the logit margin by $1.751$ on average, yet AUROC is unchanged to four
decimals ($0.99326$ against $0.99336$) and TPR@1\,\% FPR falls $0.40$\,pp:
saturation absorbs the shift. A model trained \emph{without} the branch scores
$92.30\,\%$ against $92.31\,\%$, so its training signal is not needed either.
Out of distribution it is harmful, costing $23.4$\,pp of TPR@1\,\% FPR on
SemEval-2024 and $7.7$\,pp on HC3-SI. Note that zeroing the raw inputs is not
equivalent to removing the branch: the projection is affine, so a zero input
still emits its bias, worth $0.172$ logits.

\paragraph{Relation to the submitted version.}
This design supersedes the A1/A3 comparison in the submitted paper, whose two
arms shared a corpus that had already been normalised at construction time; their
difference therefore reflected a train/test mismatch rather than the effect of
preprocessing.

\subsection{Backbone Comparison}
\label{subsec:backbones}

Is the result specific to DeBERTa? We trained \texttt{bert-large-cased} and
\texttt{roberta-large} on the identical raw corpus with identical
hyperparameters and seed and submitted all three to the hidden test
(Table~\ref{tab:backbones}, Appendix~\ref{app:backbones}). All three
tokenisers fragment homoglyph-substituted words into subword pieces rather
than emitting unknown tokens, so the vulnerability is a property of subword
tokenisation rather than of one model. DeBERTa leads by six points of
TPR@1\,\% FPR over BERT and by twenty-five over RoBERTa ($99.61$, $99.07$
and $94.32\,\%$ AUROC), and the backbone ranking on our validation pool does
not survive the hidden test: RoBERTa leads BERT by nine points there and
trails it by nineteen on the hidden test.

\subsection{Impact of the Training Data Scaling and Validation Design}
\label{subsec:scaling}

\paragraph{Scaling to the full RAID corpus.}
To test whether the gains continue, we trained the best configuration on the
complete RAID training split ($9.38$M balanced rows, roughly six times the
adversarial exposure), selecting the checkpoint on validation AUROC, which
peaked at the second epoch. Hidden-test performance improves marginally, to
$99.63\,\%$ AUROC and $97.12\,\%$ TPR@1\,\% FPR, an increase of $0.55$\,pp over
the $1.55$M-row configuration. On a leakage-free evaluation restricted to
sources absent from both training corpora, however, the same model loses
$3.42$\,pp of TPR@1\,\% FPR on MAGE ($91.99 \rightarrow 88.57$, disjoint
$95\,\%$ intervals), and its cross-dataset average falls from $93.14$ to
$92.45\,\%$ (Table~\ref{tab:baseline_comparison}). Sixfold adversarial scaling
therefore buys about half a point on the benchmark and costs several points of
cross-dataset generalisation, which is the trade-off our multi-dataset design is intended to
avoid. We report the $1.55$M-row configuration as our primary system for this
reason.

Validation composition also matters: an early configuration used a validation
split in which MAGE was over-represented relative to training, biasing early
stopping and costing three points of HC3-SI. Correcting to proportional
validation restored it (Appendix~\ref{app:validation}).

\subsection{Cross-Dataset Generalisation}
\label{subsec:crossdataset}

Table~\ref{tab:baseline_comparison} evaluates the four no-feature cells on
three external corpora under the fixed-threshold protocol, with $\tau^\star$
calibrated once on the source validation split and held constant.

The result is that preprocessing placement, which separates the cells by
$12.9$\,pp on RAID, separates them by at most $0.41$\,pp here: the
cross-dataset averages are $93.35$, $93.55$, $93.17$ and $93.14\,\%$. Inference
normalisation in particular is inert: N-nn against N-nr, and N-rr against
N-rn, are the same weights differing only in inference, and their AUROC and
TPR@1\,\% FPR agree to four decimal places on every benchmark. This is what
the mechanism predicts: HC3 Plus and MAGE contain no Unicode perturbations, so
normalisation has nothing to remove. Attack-aware preprocessing is not a
general preprocessing benefit but a defence that is exactly inert in its
absence.

\paragraph{M4GT-Bench.}
On M4GT-Bench~\citep{Wang24-M4GTBench}, which was not part of training, the four cells reach
$97.67$, $94.55$, $96.70$ and $93.59\,\%$ TPR@1\,\% FPR for
$+$F\,n$\to$n, $-$F\,n$\to$n, $+$F\,r$\to$r and $-$F\,r$\to$n respectively.
This is the one external corpus on which the feature branch helps, by about
three points in both training conditions, the opposite sign to its effect
on SemEval-2024 and HC3-SI, and a reminder that the branch is not uniformly
harmful but unpredictable.

Raw-corpus training is close to neutral rather than free: it costs about two
points on HC3-SI and one on M4 while gaining one on MAGE, for a net change
within half a point on the average. Set against a $17.1$\,pp gain in RAID
TPR@1\,\% FPR, we regard that as an acceptable exchange, and we report it
rather than the average alone.

Further out-of-distribution evaluation (Appendix~\ref{app:external}) and
the development sequence behind the submitted version
(Appendix~\ref{app:versions}) are consistent with these findings.

%%%%%%%%%%%%%%%%%%%%%%%%%%%%%%%%%%%%%%%%%%%%%%%%%%%%%%%%%%%%
\begin{table}[t]
\centering\small
\setlength{\tabcolsep}{3pt}
\fitcol{%
\begin{tabular}{@{}lccccc@{}}
\toprule
Method & HC3-QA & HC3-SI & MAGE & Avg & RAID \\
\midrule
\multicolumn{6}{@{}l}{\textit{RAID leaderboard systems with public checkpoints}} \\
MELD                   & \pc{94.64} & \pc{66.86} & \pc{96.17} & \pc{85.89} & \pc{99.78} \\
ModernBERT$^{\ast}$    & \pc{95.40} & \pc{52.54} & \pc{93.51} & \pc{80.48} & \pc{94.14} \\
Desklib v1.01          & \pc{97.87} & \pc{56.38} & \pc{83.44} & \pc{79.23} & \pc{91.17} \\
SuperAnnotate          & \pc{99.04} & \pc{56.25} & \pc{60.01} & \pc{71.77} & \pc{64.87} \\
e5-small-lora          & \pc{87.29} & \pc{59.45} & \pc{67.12} & \pc{71.29} & \pc{85.69} \\
TMR                    & \pc{84.85} & \pc{55.92} & \pc{70.99} & \pc{70.59} & \pc{95.79} \\
BERT-tiny-4M           & \pc{69.84} & \pc{55.60} & \pc{61.49} & \pc{62.31} & \pc{84.18} \\
ADAL                   & \pc{59.89} & \pc{44.89} & \pc{61.23} & \pc{55.34} & \pc{96.25} \\
RADAR                  & \pc{53.32} & \pc{49.09} & \pc{60.40} & \pc{54.27} & \pc{63.91} \\
\midrule
\multicolumn{6}{@{}l}{\textit{Ours, no-feature cells (DeBERTa-v3-large)}} \\
$-$F\ \ n$\to$n         & \pc{99.72} & \pc{84.79} & \pc{95.55} & \pc{93.35} & \pc{98.50} \\
$-$F\ \ n$\to$r          & \pc{99.67} & \pc{85.77} & \pc{95.22} & \pc{93.55} & \pc{86.14} \\
$-$F\ \ r$\to$r           & \pc{99.68} & \pc{83.64} & \pc{96.20} & \pc{93.17} & \pc{98.39} \\
\textbf{$-$F\ \ r$\to$n} & \pc{99.69} & \pc{83.50} & \pc{96.23} & \pc{93.14} & \pc{99.01} \\
\midrule
\multicolumn{6}{@{}l}{\textit{Ours, other backbones (raw/norm)}} \\
BERT-large             & \pc{98.85} & \pc{81.39} & \pc{92.59} & \pc{90.94} & \pc{96.77} \\
RoBERTa-large          & \pc{99.28} & \pc{72.38} & \pc{92.70} & \pc{88.12} & \pc{87.08} \\
\midrule
\multicolumn{6}{@{}l}{\textit{Ours, trained on the full RAID split (Sec.~\ref{subsec:scaling})}} \\
$-$F\ \ r$\to$n & \pc{99.72} & \pc{83.71} & \pc{93.94} & \pc{92.45} & \pc{99.13} \\
\bottomrule
\end{tabular}}
\caption{Balanced accuracy at the fixed threshold $\tau^\star$ on the
external benchmarks, and RAID TPR@5\,\% FPR from the official
leaderboard~\citep{Dugan24-RAID}. Avg is the mean over HC3-QA, HC3-SI and
MAGE. Competitors are every RAID entry with a public checkpoint, each scored
under our protocol. The feature cells were not evaluated externally.
$^{\ast}$~Trained on MAGE, so that column is in-distribution for it.}
\label{tab:baseline_comparison}
\end{table}

\subsection{Comparison Against Strong Baselines}
\label{subsec:baselines}
\label{sec:baseline_comparison}

We compare against every RAID leaderboard system that publishes a checkpoint
(Table~\ref{tab:baseline_comparison}), each downloaded and evaluated under our
fixed-threshold protocol rather than at numbers reported elsewhere. MELD is the only open system above ours on RAID
alone ($99.78$ against $99.01$); on adversarial robustness in isolation it is
the stronger detector. And HC3, MAGE and M4 are training sources for our
system, so those columns are held-out splits for us but external data for
every competitor, which favours us by construction. What the table does show
is that no open system combines both: our configuration leads the
cross-dataset average by $7.3$ points over MELD, and the systems nearest us on
RAID (ADAL, TMR) fall to near chance on HC3-SI. Several commercial systems score above $99$ on RAID but publish no
checkpoint.

\section{Discussion and Conclusion}
\label{sec:discussion}

In-domain accuracy alone is an insufficient indicator of robustness. Among
systems with public checkpoints, ours is the only one that
is both within a point of the best RAID score and first on the cross-dataset
average.

Two findings generalise. Normalisation before tokenisation neutralises
zero-width and homoglyph attacks, but only at inference: applying it to the
training corpus deletes the adversarial supervision it provides. And
robustness here proved data-centric: the feature branch reduced TPR@1\,\%
FPR in every condition, by $49.8$\,pp on poetry alone.

Both findings are cheap to act on. Inference-time normalisation is a
deterministic pre-tokenisation step with no learned parameters, and the same
signature appears in a zero-shot detector of different architecture, so
existing detectors can adopt it without retraining; it cannot replace
adversarial exposure during training.

{\raggedright\noindent Code, checkpoint and live demo:\\
{\small\href{https://github.com/MohamedMady19/deberta-conpara}{\nolinkurl{github.com/MohamedMady19/deberta-conpara}}\\
\href{https://huggingface.co/mohamedmady/deberta-conpara}{\nolinkurl{huggingface.co/mohamedmady/deberta-conpara}}\\
\href{https://huggingface.co/spaces/mohamedmady/deberta-conpara}{\nolinkurl{huggingface.co/spaces/mohamedmady/deberta-conpara}}}\par}

\newpage
\clearpage
% === MANDATORY *ACL SECTIONS ===
\section*{Limitations}

While DeBERTa-ConPara demonstrates strong robustness across multiple benchmarks and adversarial settings, several limitations remain.

First, each configuration is a single training run. Three seeds of one
configuration spanned $10.68$ points of hidden-test TPR@1\,\% FPR. That
spread belongs to the metric as much as to the model: TPR@1\,\% FPR is fixed
by one percentile of the human score distribution, so small calibration
shifts displace it disproportionately, whereas balanced accuracy on the same
system varies by $\pm0.75$\,pp across five seeds
(Appendix~\ref{app:multiseed}). Our factorial comparisons are paired
(identical corpus and seed, one factor changed), so they are not subject to
the full between-run spread, but we could not quantify the paired variance
without repeating the factorial. We therefore rest our conclusions on effects
that are large relative to that spread (raw-corpus training, $+2.05$ to
$+18.05$\,pp; the poetry penalty, $49.8$\,pp) or consistent in sign across all
four conditions (the feature branch is negative in every one). Differences of
a few points between cells are indicative only; where we draw a conclusion in
that range, as in the scaling analysis, it rests on the cross-dataset loss
($3.42$\,pp, disjoint $95\,\%$ intervals), not on the small hidden-test gain.

Second, all experiments focus exclusively on English-language datasets, and all robustness claims should be read as applying only to the English domains, generators and attacks evaluated here. Extending the framework to multilingual settings remains an important direction, particularly since Unicode-based attack patterns and linguistic structures vary substantially across languages. Multilingual benchmarks such as DetectRL-X~\citep{Wu26-DetectRLX}, which spans eight languages and six domains, report that detector performance degrades unevenly across scripts and that short texts are substantially harder in morphologically richer languages than in English; whether attack-aware normalisation transfers to non-Latin scripts is therefore an open question our results do not address.

Third, although attack-aware preprocessing substantially improves robustness against surface-level perturbations, semantically preserving transformations such as sophisticated paraphrasing remain challenging for current detection systems.

Fourth, the human-written training data originates from pre-2022 sources, predating the widespread availability of large-scale generative AI. While this temporal boundary provides provenance guarantees, human writing style is evolving in response to AI use, and AI-assisted composition is emerging as a challenging intermediate category that binary detection frameworks do not address. We consider the development of three-class detection (human, AI-assisted, fully AI-generated) an important direction for future work.

Finally, robustness evaluation is necessarily bounded by currently available benchmarks and attack strategies. AI text generation is advancing rapidly, including the emergence of post-processing humanisation filters that deliberately reduce the statistical detectability of AI-generated text~\citep{Chakraborty24-Position}. A particularly challenging emerging scenario involves personally adapted AI systems that deliberately imitate an individual's writing style including their characteristic errors, vocabulary, and syntactic patterns making stylometric detection ineffective by design. Addressing such threats will require author-adaptive detection approaches and continued benchmark refreshment as both human writing and AI generation capabilities evolve.

\section*{Ethical Considerations}

Reliable AI-generated text detection can support important applications including academic integrity, misinformation mitigation, content moderation, and quality control in human-annotated dataset construction.

Detection systems carry false-positive risks, particularly for non-native speakers, minority dialects and stylistically atypical human writing, while false negatives cause harm in academic-integrity and misinformation settings. We therefore recommend that detectors serve as decision support rather than definitive evidence in high-stakes settings, accompanied by human oversight, transparent confidence reporting and clear appeal mechanisms.

The released system is intended for defensive detection research rather than surveillance or censorship applications.

\newpage
\clearpage
\bibliography{references}

@inproceedings{Chakraborty24-Position,
  author    = {Chakraborty, Souradip and Bedi, Amrit and
               Zhu, Sicheng and An, Bang and Manocha, Dinesh and
               Huang, Furong},
  title     = {Position: On the Possibilities of {AI}-Generated
               Text Detection},
  booktitle = {Proceedings of the 41st International Conference
               on Machine Learning},
  series    = {Proceedings of Machine Learning Research},
  volume    = {235},
  pages     = {6093--6115},
  publisher = {PMLR},
  address   = {Vienna, Austria},
  month     = jul,
  year      = {2024},
  url       = {https://proceedings.mlr.press/v235/chakraborty24a.html}
}

@misc{Guo23-HC3,
  author        = {Guo, Biyang and Zhang, Xin and Wang, Ziyuan and
                   Jiang, Minqi and Nie, Jinran and Ding, Yuxuan and
                   Yue, Jianwei and Wu, Yupeng},
  title         = {How Close is {ChatGPT} to Human Experts?
                   Comparison Corpus, Evaluation, and Detection},
  year          = {2023},
  eprint        = {2301.07597},
  archivePrefix = {arXiv},
  primaryClass  = {cs.CL},
  doi           = {10.48550/arXiv.2301.07597},
  howpublished  = {arXiv preprint arXiv:2301.07597}
}

@misc{Su23-HC3Plus,
  author        = {Su, Zhenpeng and Wu, Xing and Zhou, Wei and
                   Ma, Guangyuan and Hu, Songlin},
  title         = {{HC3} Plus: A Semantic-Invariant
                   Human {ChatGPT} Comparison Corpus},
  year          = {2023},
  eprint        = {2309.02731},
  archivePrefix = {arXiv},
  primaryClass  = {cs.CL},
  doi           = {10.48550/arXiv.2309.02731},
  howpublished  = {arXiv preprint arXiv:2309.02731}
}

@inproceedings{Wang24-M4,
  author    = {Wang, Yuxia and Mansurov, Jonibek and
               Ivanov, Petar and Su, Jinyan and
               Shelmanov, Artem and Tsvigun, Akim and
               Whitehouse, Chenxi and Mohammed Afzal, Osama and
               Mahmoud, Tarek and Sasaki, Toru and
               Arnold, Thomas and Aji, Alham Fikri and
               Habash, Nizar and Gurevych, Iryna and
               Nakov, Preslav},
  title     = {{M4}: Multi-generator, Multi-domain, and
               Multi-lingual Black-Box Machine-Generated
               Text Detection},
  booktitle = {Proceedings of the 18th Conference of the
               European Chapter of the Association for
               Computational Linguistics (Volume 1:
               Long Papers)},
  pages     = {1369--1407},
  publisher = {Association for Computational Linguistics},
  address   = {St. Julian's, Malta},
  month     = mar,
  year      = {2024},
  doi       = {10.18653/v1/2024.eacl-long.83},
  url       = {https://aclanthology.org/2024.eacl-long.83}
}

@inproceedings{Li24-MAGE,
  author    = {Li, Yafu and Li, Qintong and Cui, Leyang and
               Bi, Wei and Wang, Zhilin and Wang, Longyue and
               Yang, Linyi and Shi, Shuming and Zhang, Yue},
  title     = {{MAGE}: Machine-generated Text Detection
               in the Wild},
  booktitle = {Proceedings of the 62nd Annual Meeting of the
               Association for Computational Linguistics
               (Volume 1: Long Papers)},
  pages     = {36--53},
  publisher = {Association for Computational Linguistics},
  address   = {Bangkok, Thailand},
  month     = aug,
  year      = {2024},
  doi       = {10.18653/v1/2024.acl-long.3},
  url       = {https://aclanthology.org/2024.acl-long.3/}
}

@inproceedings{Dugan24-RAID,
  author    = {Dugan, Liam and Hwang, Alyssa and
               Trhl\'{i}k, Filip and Zhu, Andrew and
               Ludan, Josh Magnus and Xu, Hainiu and
               Ippolito, Daphne and Callison-Burch, Chris},
  title     = {{RAID}: A Shared Benchmark for Robust Evaluation
               of Machine-Generated Text Detectors},
  booktitle = {Proceedings of the 62nd Annual Meeting of the
               Association for Computational Linguistics
               (Volume 1: Long Papers)},
  pages     = {12463--12492},
  publisher = {Association for Computational Linguistics},
  address   = {Bangkok, Thailand},
  month     = aug,
  year      = {2024},
  doi       = {10.18653/v1/2024.acl-long.674},
  url       = {https://aclanthology.org/2024.acl-long.674/}
}

@inproceedings{Devlin19-BERT,
  author    = {Devlin, Jacob and Chang, Ming-Wei and
               Lee, Kenton and Toutanova, Kristina},
  title     = {{BERT}: Pre-training of Deep Bidirectional
               Transformers for Language Understanding},
  booktitle = {Proceedings of the 2019 Conference of the
               North American Chapter of the Association for
               Computational Linguistics: Human Language
               Technologies, Volume 1 (Long and Short Papers)},
  pages     = {4171--4186},
  publisher = {Association for Computational Linguistics},
  address   = {Minneapolis, Minnesota},
  month     = jun,
  year      = {2019},
  doi       = {10.18653/v1/N19-1423},
  url       = {https://aclanthology.org/N19-1423}
}

@misc{Liu19-RoBERTa,
  author        = {Liu, Yinhan and Ott, Myle and Goyal, Naman and
                   Du, Jingfei and Joshi, Mandar and Chen, Danqi and
                   Levy, Omer and Lewis, Mike and Zettlemoyer, Luke and
                   Stoyanov, Veselin},
  title         = {{RoBERTa}: A Robustly Optimized {BERT}
                   Pretraining Approach},
  year          = {2019},
  eprint        = {1907.11692},
  archivePrefix = {arXiv},
  primaryClass  = {cs.CL},
  url           = {https://arxiv.org/abs/1907.11692},
  howpublished  = {arXiv preprint arXiv:1907.11692}
}

@inproceedings{Clark20-ELECTRA,
  author    = {Clark, Kevin and Luong, Minh-Thang and
               Le, Quoc V. and Manning, Christopher D.},
  title     = {{ELECTRA}: Pre-training Text Encoders as
               Discriminators Rather Than Generators},
  booktitle = {8th International Conference on Learning
               Representations},
  address   = {Addis Ababa, Ethiopia},
  month     = apr,
  year      = {2020},
  url       = {https://openreview.net/forum?id=r1xMH1BtvB}
}

@inproceedings{He21-DeBERTaV3,
  author    = {He, Pengcheng and Gao, Jianfeng and Chen, Weizhu},
  title     = {{DeBERTaV3}: Improving {DeBERTa} using
               {ELECTRA}-Style Pre-Training with
               Gradient-Disentangled Embedding Sharing},
  booktitle = {The Eleventh International Conference on Learning
               Representations},
  address   = {Kigali, Rwanda},
  month     = may,
  year      = {2023},
  url       = {https://openreview.net/forum?id=sE7-XhLxHA}
}

@misc{Achiam23-GPT4,
  author        = {Achiam, Josh and Adler, Steven and Agarwal, Sandhini
                   and Ahmad, Lama and Akkaya, Ilge and others},
  title         = {{GPT-4} Technical Report},
  year          = {2023},
  eprint        = {2303.08774},
  archivePrefix = {arXiv},
  primaryClass  = {cs.CL},
  howpublished  = {arXiv preprint arXiv:2303.08774}
}

@misc{Gemini23,
  author        = {{Gemini Team}},
  title         = {Gemini: A Family of Highly Capable
                   Multimodal Models},
  year          = {2023},
  eprint        = {2312.11805},
  archivePrefix = {arXiv},
  primaryClass  = {cs.CL},
  howpublished  = {arXiv preprint arXiv:2312.11805}
}

@inproceedings{Gehrmann19-GLTR,
  author    = {Gehrmann, Sebastian and Strobelt, Hendrik and
               Rush, Alexander},
  title     = {{GLTR}: Statistical Detection and Visualization
               of Generated Text},
  booktitle = {Proceedings of the 57th Annual Meeting of the
               Association for Computational Linguistics:
               System Demonstrations},
  pages     = {111--116},
  publisher = {Association for Computational Linguistics},
  address   = {Florence, Italy},
  month     = jul,
  year      = {2019},
  doi       = {10.18653/v1/P19-3019},
  url       = {https://aclanthology.org/P19-3019}
}

@inproceedings{li2025discourse,
  author    = {Li, Yupei and Milling, Manuel and
               Specia, Lucia and Schuller, Bj\"{o}rn W.},
  title     = {Discourse Features Enhance Detection of
               Document-Level Machine-Generated Content},
  booktitle = {Proceedings of the 2025 International Joint
               Conference on Neural Networks (IJCNN)},
  pages     = {1--8},
  publisher = {IEEE},
  address   = {Rome, Italy},
  month     = jun,
  year      = {2025},
  doi       = {10.1109/IJCNN64981.2025.11228879}
}

@inproceedings{Mitchell23-DetectGPT,
  author    = {Mitchell, Eric and Lee, Yoonho and
               Khazatsky, Alexander and Manning, Christopher D.
               and Finn, Chelsea},
  title     = {{DetectGPT}: Zero-Shot Machine-Generated Text
               Detection using Probability Curvature},
  booktitle = {Proceedings of the 40th International Conference
               on Machine Learning},
  series    = {Proceedings of Machine Learning Research},
  volume    = {202},
  pages     = {24950--24962},
  publisher = {PMLR},
  address   = {Honolulu, Hawaii, USA},
  month     = jul,
  year      = {2023},
  url       = {https://proceedings.mlr.press/v202/mitchell23a.html}
}

@inproceedings{Hu23-RADAR,
  author    = {Hu, Xiaomeng and Chen, Pin-Yu and Ho, Tsung-Yi},
  title     = {{RADAR}: Robust {AI}-Text Detection via
               Adversarial Learning},
  booktitle = {Advances in Neural Information Processing
               Systems 36},
  pages     = {15077--15095},
  publisher = {Curran Associates, Inc.},
  address   = {New Orleans, Louisiana, USA},
  year      = {2023},
  doi       = {10.52202/075280-0662},
  url       = {https://proceedings.neurips.cc/paper_files/paper/2023/hash/30e15e5941ae0cdab7ef58cc8d59a4ca-Abstract-Conference.html}
}

@inproceedings{Bao24-FastDetectGPT,
  author    = {Bao, Guangsheng and Zhao, Yanbin and
               Teng, Zhiyang and Yang, Linyi and Zhang, Yue},
  title     = {Fast-{DetectGPT}: Efficient Zero-Shot Detection
               of Machine-Generated Text via Conditional
               Probability Curvature},
  booktitle = {12th International Conference on Learning
               Representations},
  address   = {Vienna, Austria},
  month     = may,
  year      = {2024},
  url       = {https://openreview.net/forum?id=Bpcgcr8E8Z}
}

@inproceedings{Hans24-Binoculars,
  author    = {Hans, Abhimanyu and Schwarzschild, Avi and
               Cherepanova, Valeriia and Kazemi, Hamid and
               Saha, Aniruddha and Goldblum, Micah and
               Geiping, Jonas and Goldstein, Tom},
  title     = {Spotting {LLMs} With Binoculars: Zero-Shot
               Detection of Machine-Generated Text},
  booktitle = {Proceedings of the 41st International Conference
               on Machine Learning},
  series    = {Proceedings of Machine Learning Research},
  volume    = {235},
  pages     = {17519--17537},
  publisher = {PMLR},
  address   = {Vienna, Austria},
  month     = jul,
  year      = {2024},
  url       = {https://proceedings.mlr.press/v235/hans24a.html}
}

@misc{Bhattacharjee24-EAGLE,
  author        = {Bhattacharjee, Amrita and Moraffah, Raha and
                   Garland, Joshua and Liu, Huan},
  title         = {{EAGLE}: A Domain Generalization Framework
                   for {AI}-Generated Text Detection},
  year          = {2024},
  eprint        = {2403.15690},
  archivePrefix = {arXiv},
  primaryClass  = {cs.CL},
  howpublished  = {arXiv preprint arXiv:2403.15690}
}

@inproceedings{Liu25-MGTPrism,
  author    = {Liu, Shengchao and Liu, Xiaoming and
               Li, Chengzhengxu and Zhang, Zhaohan and
               Ma, Guoxin and Lan, Yu and Xiao, Shuai},
  title     = {{MGT-Prism}: Enhancing Domain Generalization
               for Machine-Generated Text Detection via
               Spectral Alignment},
  booktitle = {Proceedings of the AAAI Conference on Artificial
               Intelligence},
  volume    = {40},
  number    = {38},
  pages     = {32132--32140},
  year      = {2026},
  doi       = {10.1609/aaai.v40i38.40485}
}

@inproceedings{Kirchenbauer23-Watermark,
  author    = {Kirchenbauer, John and Geiping, Jonas and
               Wen, Yuxin and Katz, Jonathan and Miers, Ian and
               Goldstein, Tom},
  title     = {A Watermark for Large Language Models},
  booktitle = {Proceedings of the 40th International Conference
               on Machine Learning},
  series    = {Proceedings of Machine Learning Research},
  volume    = {202},
  pages     = {17061--17084},
  publisher = {PMLR},
  address   = {Honolulu, Hawaii, USA},
  month     = jul,
  year      = {2023},
  url       = {https://proceedings.mlr.press/v202/kirchenbauer23a.html}
}

@inproceedings{Wouters24-Watermarks,
  author    = {Wouters, Bram},
  title     = {Optimizing Watermarks for Large Language Models},
  booktitle = {Proceedings of the 41st International Conference
               on Machine Learning},
  series    = {Proceedings of Machine Learning Research},
  volume    = {235},
  pages     = {53251--53269},
  publisher = {PMLR},
  address   = {Vienna, Austria},
  month     = jul,
  year      = {2024},
  url       = {https://proceedings.mlr.press/v235/wouters24a.html}
}

@inproceedings{Zhang24-StrongWatermarking,
  author    = {Zhang, Hanlin and Edelman, Benjamin L. and
               Francati, Danilo and Venturi, Daniele and
               Ateniese, Giuseppe and Barak, Boaz},
  title     = {Watermarks in the Sand: Impossibility of Strong
               Watermarking for Language Models},
  booktitle = {Proceedings of the 41st International Conference
               on Machine Learning},
  series    = {Proceedings of Machine Learning Research},
  volume    = {235},
  pages     = {58851--58880},
  publisher = {PMLR},
  address   = {Vienna, Austria},
  month     = jul,
  year      = {2024},
  url       = {https://proceedings.mlr.press/v235/zhang24o.html}
}

@inproceedings{Loshchilov17-AdamW,
  author    = {Loshchilov, Ilya and Hutter, Frank},
  title     = {Decoupled Weight Decay Regularization},
  booktitle = {7th International Conference on Learning
               Representations},
  address   = {New Orleans, Louisiana, USA},
  month     = may,
  year      = {2019},
  url       = {https://openreview.net/forum?id=Bkg6RiCqY7}
}

@inproceedings{Khosla20-SupCon,
  author    = {Khosla, Prannay and Teterwak, Piotr and Wang, Chen and
               Sarna, Aaron and Tian, Yonglong and Isola, Phillip and
               Maschinot, Aaron and Liu, Ce and Krishnan, Dilip},
  title     = {Supervised Contrastive Learning},
  booktitle = {Advances in Neural Information Processing
               Systems 33},
  pages     = {18661--18673},
  publisher = {Curran Associates, Inc.},
  address   = {Virtual},
  year      = {2020},
  url       = {https://proceedings.neurips.cc/paper/2020/hash/d89a66c7c80a29b1bdbab0f2a1a94af8-Abstract.html}
}

@misc{Smith19-OneCycleLR,
  author        = {Smith, Leslie N. and Topin, Nicholay},
  title         = {Super-Convergence: Very Fast Training of Neural
                   Networks Using Large Learning Rates},
  year          = {2019},
  eprint        = {1708.07120},
  archivePrefix = {arXiv},
  primaryClass  = {cs.LG},
  url           = {https://arxiv.org/abs/1708.07120},
  note          = {Also published in Proc. SPIE 11006, Artificial
                   Intelligence and Machine Learning for Multi-Domain
                   Operations Applications, 2019}
}

@article{Chung22-FlanT5,
  author  = {Chung, Hyung Won and Hou, Le and
             Longpre, Shayne and Zoph, Barret and
             Tay, Yi and Fedus, William and Li, Yunxuan and
             Wang, Xuezhi and Dehghani, Mostafa and
             Brahma, Siddhartha and others},
  title   = {Scaling Instruction-Finetuned Language Models},
  journal = {Journal of Machine Learning Research},
  volume  = {25},
  number  = {70},
  pages   = {1--53},
  year    = {2024},
  url     = {https://jmlr.org/papers/v25/23-0870.html}
}

@book{CoverThomas06-InfoTheory,
  author    = {Cover, Thomas M. and Thomas, Joy A.},
  title     = {Elements of Information Theory},
  publisher = {John Wiley \& Sons},
  address   = {Hoboken, New Jersey, USA},
  edition   = {2nd},
  year      = {2006},
  isbn      = {978-0-471-24195-9}
}

@article{Youden50-Index,
  author  = {Youden, W. J.},
  title   = {Index for Rating Diagnostic Tests},
  journal = {Cancer},
  volume  = {3},
  number  = {1},
  pages   = {32--35},
  year    = {1950},
  doi     = {10.1002/1097-0142(1950)3:1<32::aid-cncr2820030106>3.0.co;2-3}
}

@article{Fawcett06-ROC,
  author  = {Fawcett, Tom},
  title   = {An Introduction to {ROC} Analysis},
  journal = {Pattern Recognition Letters},
  volume  = {27},
  number  = {8},
  pages   = {861--874},
  year    = {2006},
  doi     = {10.1016/j.patrec.2005.10.010}
}

@misc{StanfordAIIndex25,
  author       = {{Stanford Institute for Human-Centered Artificial Intelligence (HAI)}},
  title        = {The 2025 {AI} Index Report},
  year         = {2025},
  howpublished = {\url{https://hai.stanford.edu/ai-index/2025-ai-index-report}},
  note         = {Accessed: 2025-05-01}
}

@misc{Turnitin24-Wrapped,
  author       = {{Turnitin LLC}},
  title        = {2024 Turnitin Wrapped},
  year         = {2024},
  howpublished = {\url{https://www.turnitin.com/blog/2024-turnitin-wrapped}},
  note         = {Accessed: 2025-05-01}
}

@article{Goldstein23-GenAI,
  author  = {Goldstein, Josh A. and Sastry, Girish and
             Musser, Micah and DiResta, Ren{\'e}e and
             Gentzel, Matthew and Sedova, Katerina},
  title   = {Generative Language Models and Automated
             Influence Operations: Emerging Threats and
             Potential Mitigations},
  journal = {arXiv preprint arXiv:2301.04246},
  year    = {2023},
  eprint        = {2301.04246},
  archivePrefix = {arXiv},
  primaryClass  = {cs.CY},
  url           = {https://arxiv.org/abs/2301.04246}
}

@article{Stokel-Walker22-ChatGPT,
  author  = {Stokel-Walker, Chris},
  title   = {{AI} Bot {ChatGPT} Writes Smart Essays ---
             Should Professors Worry?},
  journal = {Nature},
  year    = {2022},
  month   = dec,
  doi     = {10.1038/d41586-022-04397-7},
  url     = {https://www.nature.com/articles/d41586-022-04397-7},
  note    = {News feature, 9 December 2022}
}

@article{Sadasivan23-CanAI,
  author  = {Sadasivan, Vinu Sankar and Kumar, Aounon and
             Balasubramanian, Sriram and Wang, Wenxiao and
             Feizi, Soheil},
  title   = {Can {AI}-Generated Text be Reliably Detected?
             Stress Testing {AI} Text Detectors Under Various Attacks},
  journal = {Transactions on Machine Learning Research},
  year    = {2025},
  url     = {https://openreview.net/forum?id=NvSwR4IvLO}
}

@inproceedings{Verma23-Ghostbuster,
  author    = {Verma, Vivek and Fleisig, Eve and
               Tomlin, Nicholas and Klein, Dan},
  title     = {Ghostbuster: Detecting Text Ghostwritten by
               Large Language Models},
  booktitle = {Proceedings of the 2024 Conference of the
               North American Chapter of the Association for
               Computational Linguistics: Human Language
               Technologies (Volume 1: Long Papers)},
  pages     = {1702--1717},
  publisher = {Association for Computational Linguistics},
  address   = {Mexico City, Mexico},
  month     = jun,
  year      = {2024},
  doi       = {10.18653/v1/2024.naacl-long.95},
  url       = {https://aclanthology.org/2024.naacl-long.95}
}

@inproceedings{Krishna23-Paraphrase,
  author    = {Krishna, Kalpesh and Song, Yixiao and
               Karpinska, Marzena and Wieting, John and
               Iyyer, Mohit},
  title     = {Paraphrasing Evades Detectors of {AI}-Generated
               Text, but Retrieval is an Effective Defense},
  booktitle = {Advances in Neural Information Processing
               Systems 36},
  pages     = {27469--27500},
  publisher = {Curran Associates, Inc.},
  address   = {New Orleans, Louisiana, USA},
  year      = {2023},
  doi       = {10.52202/075280-1195},
  url       = {https://proceedings.neurips.cc/paper_files/paper/2023/hash/575c450013d0e99e4b0ecf82bd1afaa4-Abstract-Conference.html}
}

@inproceedings{Wu24-DetectRL,
  author    = {Wu, Junchao and Zhan, Runzhe and Wong, Derek F. and
               Yang, Shu and Yang, Xinyi and Yuan, Yulin and
               Chao, Lidia S.},
  title     = {{DetectRL}: Benchmarking {LLM}-Generated Text
               Detection in Real-World Scenarios},
  booktitle = {Advances in Neural Information Processing
               Systems 37 (NeurIPS 2024), Datasets and Benchmarks Track},
  pages     = {100369--100401},
  publisher = {Curran Associates, Inc.},
  address   = {Vancouver, BC, Canada},
  year      = {2024},
  doi       = {10.52202/079017-3186}
}

@article{Wu25-Survey,
  author  = {Wu, Junchao and Yang, Shu and Zhan, Runzhe and
             Yuan, Yulin and Chao, Lidia S. and Wong, Derek F.},
  title   = {A Survey on {LLM}-Generated Text Detection:
             Necessity, Methods, and Future Directions},
  journal = {Computational Linguistics},
  volume  = {51},
  number  = {1},
  pages   = {275--338},
  year    = {2025},
  doi     = {10.1162/coli_a_00549}
}

@misc{Mady26-FeatAug,
  author        = {Mady, Mohamed and Li, Yupei and
                   Reschke, Johannes and Schuller, Bj\"{o}rn W.},
  title         = {Feature-Augmented Transformers for Robust
                   {AI}-Text Detection Across Domains and Generators},
  year          = {2026},
  eprint        = {2605.03969},
  archivePrefix = {arXiv},
  primaryClass  = {cs.CL},
  url           = {https://arxiv.org/abs/2605.03969}
}

@misc{Mady25-HC3Gemini,
  author       = {Mady, Mohamed},
  title        = {{HC3}-{G}emini-{F}lash-{R}esponses},
  year         = {2026},
  howpublished = {Hugging Face Datasets},
  url          = {https://huggingface.co/datasets/mohamedmady/HC3-Gemini-Flash-Responses}
}

@misc{Mady25-AcademicText,
  author       = {Mady, Mohamed},
  title        = {{A}cademic-{T}ext-arxiv-gpt-gemini},
  year         = {2026},
  howpublished = {Hugging Face Datasets},
  url          = {https://huggingface.co/datasets/mohamedmady/Academic-Text-arxiv-gpt-gemini}
}

@inproceedings{Chen26-SurpMark,
  author    = {Chen, Shuangyi and Khisti, Ashish J.},
  title     = {Black-Box Detection of {LLM}-Generated Text Using
               Generalized {J}ensen {S}hannon Divergence},
  booktitle = {Proceedings of the 43rd International Conference on
               Machine Learning},
  series    = {Proceedings of Machine Learning Research},
  volume    = {306},
  pages     = {14913--14955},
  publisher = {PMLR},
  address   = {Seoul, South Korea},
  month     = jul,
  year      = {2026},
  url       = {https://proceedings.mlr.press/v306/chen26bf.html}
}

@inproceedings{Wu26-DetectRLX,
  author    = {Wu, Junchao and Liu, Yefeng and Zhu, Chenyu and
               Zhang, Hao and Wu, Zeyu and Shi, Tianqi and
               Du, Yichao and Wang, Longyue and Luo, Weihua and
               Su, Jinsong and Wong, Derek F.},
  title     = {{DetectRL-X}: Towards Reliable Multilingual and
               Real-World {LLM}-Generated Text Detection},
  booktitle = {Proceedings of the 64th Annual Meeting of the
               Association for Computational Linguistics
               (Volume 1: Long Papers)},
  year      = {2026},
  pages     = {38247--38294},
  address   = {San Diego, California, USA},
  publisher = {Association for Computational Linguistics},
  url       = {https://aclanthology.org/2026.acl-long.1773/}
}

@article{Chen25-RepreGuard,
  author  = {Chen, Xin and Wu, Junchao and Yang, Shu and
             Zhan, Runzhe and Wu, Zeyu and Luo, Ziyang and
             Wang, Di and Yang, Min and Chao, Lidia S. and
             Wong, Derek F.},
  title   = {{RepreGuard}: Detecting {LLM}-Generated Text by
             Revealing Hidden Representation Patterns},
  journal = {Transactions of the Association for Computational
             Linguistics},
  volume  = {13},
  year    = {2025},
  pages   = {1812--1831},
  url     = {https://aclanthology.org/2025.tacl-1.81/}
}

@inproceedings{Wang24-M4GTBench,
  author    = {Wang, Yuxia and Mansurov, Jonibek and Ivanov, Petar and
               Su, Jinyan and Shelmanov, Artem and Tsvigun, Akim and
               Mohammed Afzal, Osama and Mahmoud, Tarek and
               Puccetti, Giovanni and Arnold, Thomas and Aji, Alham Fikri and
               Habash, Nizar and Gurevych, Iryna and Nakov, Preslav},
  title     = {{M4GT-Bench}: Evaluation Benchmark for Black-Box
               Machine-Generated Text Detection},
  booktitle = {Proceedings of the 62nd Annual Meeting of the
               Association for Computational Linguistics
               (Volume 1: Long Papers)},
  pages     = {3964--3992},
  publisher = {Association for Computational Linguistics},
  address   = {Bangkok, Thailand},
  month     = aug,
  year      = {2024},
  doi       = {10.18653/v1/2024.acl-long.218},
  url       = {https://aclanthology.org/2024.acl-long.218/}
}
\newpage
\clearpage

\newpage
\appendix
\raggedbottom

\section{Appendix}
\label{app:appendix}

This appendix provides additional details on dataset composition, benchmark characteristics, preprocessing, feature engineering, and robustness evaluation settings used throughout this work. The supplementary material is intended to support reproducibility and provide deeper insight into the deployment-oriented robustness properties of DeBERTa-ConPara.

%%%%%%%%%%%%%%%%%%%%%%%%%%%%%%%%%%%%%%%%%%%%%%%%%%%%%%%%%%%%
\subsection{Detailed Dataset and Benchmarks Descriptions}
\label{app:dataset_overview}
\label{app:datasets}   % alias for main text cross-reference

DeBERTa-ConPara is trained and evaluated on four complementary large-scale datasets spanning semantic invariance, cross-domain generalisation, highly fluent writing, and adversarial robustness settings.

%%%%%%%%%%%%%%%%%%%%%%%%%%%%%%%%%%%%%%%%%%%%%%%%%%%%%%%%%%%%
\subsubsection{HC3 and HC3 Plus}
\label{app:hc3_details}

The \textbf{HC3} benchmark~\citep{Guo23-HC3} 
(Human ChatGPT Comparison Corpus) is one of the earliest large-scale datasets for AI-generated text detection. It contains approximately 40,000 questions and responses from human experts collected through social media and Wikipedia, with ChatGPT (GPT-3.5 series) used to generate corresponding answers. HC3 is available in both English and Chinese, with the English subset covering five domains and the Chinese subset covering seven domains,  spanning areas including open-domain question answering, medicine, finance, and law.

\textbf{HC3 Plus}~\citep{Su23-HC3Plus} extends HC3 with semantic-invariant tasks beyond question answering. The semantic-invariant split incorporates summarisation, translation, and paraphrasing tasks, with human-written text sourced from CNN/DailyMail, XSum, LCSTS, the CLUE benchmark, and datasets from the Workshop on Machine Translation (WMT). AI-generated outputs were produced using GPT-3.5-Turbo for all tasks in the semantic-invariant split. Unlike the original HC3 question-answering tasks, semantic-invariant tasks are substantially more challenging for detection because the AI output preserves the full semantic content of the human source, eliminating the stylistic differences that make standard QA detection tractable.

We use only the English subset throughout this work. The official HC3 Plus training split contains 148,040 samples, while the SI split is reserved exclusively for semantic robustness evaluation.

\begin{table*}[t]
\centering
\small
\fitpage{%
\begin{tabular}{lccccc}
\toprule
Split & Subset & Total & Human & AI & Purpose \\
\midrule
Train      & HC3 Plus QA & 148,040 & 83,261 & 64,779 & Main training set \\
Validation & HC3 QA      & 5,839   & 2,919  & 2,920  & Standard validation \\
Validation & HC3 SI      & 5,839   & 2,919  & 2,920  & Semantic-invariant validation \\
Test       & HC3 QA      & 24,969  & 16,951 & 8,018  & Standard evaluation \\
Test       & HC3 SI      & 38,110  & 19,047 & 19,063 & Semantic-invariant evaluation \\
\bottomrule
\end{tabular}}
\caption{
Detailed composition of the English splits of HC3 Plus. All counts denote number of text samples. The HC3 Plus QA test split is human-heavy following the official benchmark configuration.}
\label{tab:hc3_splits}
\end{table*}

%%%%%%%%%%%%%%%%%%%%%%%%%%%%%%%%%%%%%%%%%%%%%%%%%%%%%%%%%%%%
\subsubsection{M4}
\label{app:m4_details}

M4~\citep{Wang24-M4} (Multi-Generator, Multi-Domain, and Multi-Lingual Black-Box Machine-Generated Text Detection) is designed specifically for evaluating cross-generator and cross-domain robustness, and received the Best Resource Paper Award at EACL 2024.

The full M4 corpus spans \textbf{seven languages} (Arabic, Bulgarian, Chinese, English, Indonesian, Russian, and Urdu) and contains approximately 147,000 human-machine parallel text pairs, of which 102,000 are English and 45,000 cover the remaining six languages. In addition, the corpus contains over 10 million non-parallel human-written texts. AI-generated outputs are produced by six generators: GPT-4, ChatGPT, GPT-3.5 (\texttt{text-davinci-003}), Cohere, Dolly-v2, and BLOOMz~176B. Human-written English texts are sourced from Wikipedia (March 2022 snapshot), WikiHow, Reddit ELI5, arXiv, and PeerRead.

We use the \textbf{English subset only}, covering five domains (Wikipedia, WikiHow, Reddit ELI5, PeerRead, and arXiv) and eight AI generators (ChatGPT, text-davinci-003, Cohere, BLOOMZ, Dolly, Dolly-2, Flan-T5, and LLaMA), as detailed in Table~\ref{tab:m4_statistics}. After preprocessing, filtering, and deduplication, the English subset contains 78,766 samples (12,583 human-written and 66,183 AI-generated). Because the English subset is heavily AI-dominated (approximately 84\,\% AI-generated), we rebalance the training corpus by supplementing its human-written side with additional human-written texts drawn exclusively from the Wikipedia (March 2022 snapshot) and WikiHow subsets of M4. These sources predate the public release of ChatGPT (November 2022) and are therefore verifiably human-authored under a temporal guarantee. This filler contributes $438{,}305$ rows, the single largest component of the $1.55$M-row corpus, and is reported as its own line in Table~\ref{tab:datasets-overview} rather than folded into the M4 total. Full split statistics are provided in Table~\ref{tab:m4_statistics}.

\begin{table*}[t]
\centering
\small
\setlength{\tabcolsep}{5pt}
\renewcommand{\arraystretch}{1.05}

\fitpage{%
\begin{tabular}{lccccccccc}
\toprule
Domain & Human & ChatGPT & Davinci003 & Cohere & BLOOMZ & Dolly & Dolly2 & FlanT5 & LLaMA \\
\midrule

Wikipedia
& 3,000
& 2,995
& 3,000
& 2,336
& 3,000
& 2,702
&   
&   
&    \\

Reddit ELI5
& 3,000
& 3,000
& 3,000
& 3,000
& 3,000
&   
&   
& 4,220
&    \\

WikiHow
& 3,001
& 3,000
& 3,000
& 3,000
& 3,000
&   
& 3,000
&   
&    \\

PeerRead
& 586
& 586
& 586
& 586
&   
& 586
&   
&   
& 586 \\

arXiv
& 2,996
& 3,000
& 3,000
& 1,220
& 3,000
& 3,000
&   
& 1,780
&    \\

\midrule
\textbf{Total}
& \textbf{12,583}
& \textbf{12,581}
& \textbf{12,586}
& \textbf{10,142}
& \textbf{12,000}
& \textbf{6,288}
& \textbf{3,000}
& \textbf{6,000}
& \textbf{586} \\

\bottomrule
\end{tabular}}

\caption{Sample counts for the English subset of M4 used in this work, covering five domains and eight AI generators. A blank cell means M4 provides no text for that generator in that domain.}
\label{tab:m4_statistics}
\end{table*}

%%%%%%%%%%%%%%%%%%%%%%%%%%%%%%%%%%%%%%%%%%%%%%%%%%%%%%%%%%%%
\Needspace{6\baselineskip}
\subsubsection{MAGE}
\label{app:mage_details}

MAGE~\citep{Li24-MAGE} (MAchine-GEnerated text detection in the Wild) is a large-scale benchmark designed to evaluate AI-generated text detection under realistic, highly heterogeneous conditions resembling practical deployment scenarios.

The benchmark collects human-written texts from \textbf{seven distinct writing tasks}, including story generation, news writing, scientific abstract writing, peer review, and biomedical question answering, among others. For each human-written text, corresponding machine-generated texts are produced using \textbf{27 large language models} under three representative prompt types (continuation, topical, and specified), covering a wide spectrum of model families including OpenAI GPT (text-davinci-002/003, GPT-3.5-Turbo), LLaMA (6B/13B/30B/65B), GLM-130B, FLAN-T5 (small/base/large/xl/xxl), OPT (125M to 30B), and BigScience models (T0, BLOOM-7B1)~\citep{Li24-MAGE}.

The data is organised into \textbf{eight testbeds} of progressively increasing detection complexity, ranging from in-domain detection of a single white-box generator to the most challenging setting involving texts from entirely unseen domains and generators under paraphrasing attacks. The full corpus contains 447,674 samples. Following the official split configuration, we use 319,071 training samples.

Compared with HC3 Plus and M4, MAGE exhibits substantially weaker stylistic separation between human and AI text, particularly in formal and polished writing domains. This is partly because the dataset explicitly captures the ``in-the-wild'' regime where surface-level lexical cues are insufficient for reliable detection. This makes MAGE especially challenging under deployment-realistic fixed-threshold evaluation.

%%%%%%%%%%%%%%%%%%%%%%%%%%%%%%%%%%%%%%%%%%%%%%%%%%%%%%%%%%%%
\subsubsection{RAID}
\label{app:raid_details}

\begin{table*}[t]
\centering
\small
\setlength{\tabcolsep}{5pt}
\renewcommand{\arraystretch}{1.05}

\fitpage{%
\begin{tabular}{l p{4.2cm} p{4.2cm} c}
\toprule
Attack Type & Example (Attacked) & After Preprocessing & Effectiveness \\
\midrule

Homoglyph substitution
&
Th\i{}s \i{}s \~a test
&
This is a test
&
Fully neutralised
\\

Zero-width insertion
&
This[ZWS]is[ZWS]a test
&
This is a test
&
Fully neutralised
\\

Whitespace perturbation
&
This~~~is~~~a~~~test
&
This is a test
&
Fully neutralised
\\

Case perturbation
&
tHiS iS a TeSt
&
this is a test
&
Mostly neutralised
\\

Number/letter swap
&
Th1s 1s a test
&
This is a test
&
Partially neutralised
\\

Alternative spelling
&
colour $\rightarrow$ color
&
normalised
&
Partial
\\

Synonym substitution
&
test $\rightarrow$ examination
&
semantic change remains
&
Limited
\\

Paraphrasing
&
rewritten sentence
&
semantic change remains
&
Limited
\\

Article deletion
&
sentence without articles
&
unchanged
&
Limited
\\

Misspelling
&
artificail inteligence
&
partially corrected
&
Partial
\\

Paragraph insertion
&
sentence + inserted para.
&
unchanged
&
Limited
\\

\bottomrule
\end{tabular}}

\caption{Representative RAID attacks and the effect of the proposed preprocessing pipeline. Surface-form attacks are largely neutralised before tokenisation, while semantic attacks remain challenging.}

\label{tab:attack_taxonomy}
\end{table*}

RAID~\citep{Dugan24-RAID} (Robust AI-Text Detection 
Benchmark) is the primary benchmark used in this work 
for adversarial robustness evaluation, and the largest 
machine-generated text detection benchmark available 
at the time of writing.

% All changed now to be words (from 1 to 10)
\Needspace{8\baselineskip}
\paragraph{Corpus Structure.}
The full RAID corpus contains over \textbf{6.2 million AI-generated texts}, constructed by generating 2,000 continuations for every combination of domain, model, decoding strategy, repetition penalty, and adversarial attack. Human-written reference texts are sourced from publicly available pre-2022 datasets across the same eight domains.

\textbf{Generators (11):} ChatGPT, GPT-4, GPT-3 (\texttt{text-davinci-003}), GPT-2 XL, LLaMA-2-70B (Chat), Cohere, Cohere (Chat), MPT-30B, MPT-30B (Chat), Mistral-7B, and Mistral-7B (Chat).

\textbf{Domains (8):} arXiv abstracts, book summaries, news articles (NYT), poetry, recipes, Reddit posts, IMDb movie reviews, and Wikipedia.

\textbf{Adversarial attacks (11):} paraphrasing, synonym substitution, homoglyph replacement, zero-width character insertion, whitespace addition, upper/lower case swap, article deletion, number swap, alternative spelling, perplexity-based misspelling, and paragraph insertion.

\textbf{Decoding strategies (4):} greedy ($T{=}0$), sampling ($T{=}1$), greedy with repetition penalty ($T{=}0$, $\Theta{=}1.2$), and sampling with repetition penalty ($T{=}1$, $\Theta{=}1.2$). Repetition penalty variants are applied to open-source models only.

\paragraph{Train/Test split.}
Ten per cent of the RAID corpus is withheld without labels as the \textbf{official hidden test set}, used exclusively through the public leaderboard at \texttt{raid-bench.xyz}. The remaining 90\,\% is released with labels and used for training. All leaderboard results are evaluated on the official hidden test set.

\paragraph{RAID rows used in this work.}
The training corpus of Table~\ref{tab:datasets-overview} draws $398{,}834$ AI-generated texts from the labelled RAID split, sampled across the attack\,$\times$\,domain cells, together with all $131{,}700$ human reference texts, $530{,}534$ rows in all. The submitted version used smaller subsets of the same construction with a fixed number of AI texts per cell and Cohere-focused oversampling (v2.2: $\sim$205k rows, 1,671 per cell; v2.4-3k: $\sim$322k, 3,000; v2.4-5k: $\sim$498k, 5,000; Appendix~\ref{app:versions}). The full-split run of Section~\ref{subsec:scaling} uses every labelled RAID training row ($9.38$M balanced rows).

\paragraph{Data currency note.}
The RAID benchmark represents the state of adversarial AI-text detection as of its 2024 release, covering 11 generators and 11 attack categories available at that time. AI text generation capabilities are advancing rapidly, including the emergence of post-processing humanisation filters designed to reduce the statistical detectability of AI-generated text~\citep{Chakraborty24-Position, Wu25-Survey}. Detectors trained on current benchmarks may therefore face accelerating distribution shift as newer, more human-like generators enter deployment. This is a known open challenge shared across the AI-text detection community, and motivates continued benchmark refreshment alongside the targeted adversarial data curation strategy introduced in this work.

%%%%%%%%%%%%%%%%%%%%%%%%%%%%%%%%%%%%%%%%%%%%%%%%%%%%%%%%%%%%
%─────────────────────────────────────────────────
% B. METHOD DETAILS
%─────────────────────────────────────────────────
\subsection{Additional Method Details}
\label{app:method_details}

This section provides additional implementation, preprocessing, feature-engineering, and training details to facilitate reproducibility. Additional analyses are included to clarify the behaviour of the proposed attack-aware preprocessing pipeline, feature-fusion mechanism, and semantic-invariance augmentation strategies under adversarial and cross-domain conditions.

%=================================================
\Needspace{12\baselineskip}
\subsubsection{Attack Taxonomy and Preprocessing Pipeline}
\label{app:attack_taxonomy}
\label{app:preprocessing}

RAID includes 11 adversarial attack categories covering both surface-form perturbations and semantic-preserving rewriting attacks. Many of these attacks exploit vulnerabilities in Unicode handling, tokenisation, whitespace processing, and stylometric feature extraction rather than high-level semantic modelling itself. Table~\ref{tab:attack_taxonomy} summarises all attack categories and their response to the proposed preprocessing pipeline.

The attack-aware preprocessing pipeline mitigates surface-form perturbations before tokenisation and feature extraction, explicitly targeting Unicode-level and formatting-level manipulations that destabilise tokenisation, entropy estimation, burstiness statistics, and stylometric feature extraction. The pipeline performs six stages applied consistently during both training and inference. Table~\ref{tab:preprocessing_pipeline} details each stage; Table~\ref{tab:preprocessing_examples} provides representative normalisation examples.

\begin{table}[t]
\centering
\small
\setlength{\tabcolsep}{4pt}
\renewcommand{\arraystretch}{1.05}
\fitcol{%
\begin{tabular}{p{1.9cm} p{5.6cm}}
\toprule
Stage & Purpose \\
\midrule
1.\ Homoglyph substitution &
    Cyrillic, Greek and fullwidth characters mapped to their Latin look-alikes through a curated confusables table. NFKD cannot repair these, as they are canonically distinct. \\
2.\ Typographic normalisation &
    Smart quotes, dashes, fractions and related symbols mapped to ASCII. \\
3.\ NFKD decomposition &
    Decomposition followed by removal of combining marks (Mn), reducing accented Latin to ASCII (\'e $\to$ e) and expanding compatibility forms; the result is re-composed with NFC. \\
4.\ Invisible char.\ removal &
    Removal of every Unicode format character (category Cf): zero-width space, ZWNJ, ZWJ, BOM, directional marks and soft hyphen. \\
5.\ Whitespace collapsing &
    Runs of spaces collapsed to one. At inference all whitespace is collapsed and the string trimmed. \\
\bottomrule
\end{tabular}}
\caption{Stages of the attack-aware preprocessing pipeline, in the order applied. The pipeline is deterministic, idempotent and carries no learned parameters.}
\label{tab:preprocessing_pipeline}
\end{table}

\begin{table}[t]
\centering
\small
\setlength{\tabcolsep}{3pt}
\renewcommand{\arraystretch}{1.05}
\fitcol{%
\begin{tabular}{@{}>{\raggedright\arraybackslash}p{2.3cm} >{\raggedright\arraybackslash}p{2.1cm} >{\raggedright\arraybackslash}p{1.6cm}@{}}
\toprule
Original Adversarial Text & After Preprocessing & Attack Type \\
\midrule
{[}Cyrillic{]} AI-generated text & AI-generated text & Homoglyph \\
\footnotesize This{[}ZWS{]}is\-{[}ZWS{]}a test   & This is a test    & Zero-width \\
T~h~i~s~~~i~s~~~A~I               & This is AI        & Whitespace \\
Th1s t3xt c0nta1ns subst.         & This text\ldots   & Char.\ subst. \\
\small tHiS Is An aDvErSaRiAl SaMpLe & this is an adv.\ & Case pert. \\
\bottomrule
\end{tabular}}
\caption{Representative adversarial-text normalisation examples performed by the preprocessing pipeline.}
\label{tab:preprocessing_examples}
\end{table}

The pipeline additionally stabilises linguistic-feature distributions. Without normalisation, Unicode perturbations artificially inflate entropy, burstiness, and vocabulary-richness estimates, leading to unstable detector behaviour under adversarial attacks.

%=================================================
\subsubsection{Linguistic Features}
\label{app:features}

The handcrafted feature space contains 62 linguistic and statistical features organised into eight categories. The top 30 features are selected for feature-attention fusion via Mutual Information~\citep{CoverThomas06-InfoTheory} computed on source-side training data, as detailed below.

\begin{table}[t]
\centering
\small
\setlength{\tabcolsep}{4pt}
\renewcommand{\arraystretch}{1.05}
\fitcol{%
\begin{tabular}{@{}l c p{4.0cm}@{}}
\toprule
\textbf{Category} & \textbf{N} & \textbf{Description} \\
\midrule
Perplexity   & 8  & GPT-2 perplexity stats (doc/token) \\
Entropy      & 10 & Char., word, n-gram, POS entropy \\
Burstiness   & 6  & Repetition and interval burstiness \\
Repetition   & 10 & Self-BLEU, unique n-grams, compression \\
Vocabulary   & 8  & Zipf, Yule, Hapax, TTR, MATTR \\
Coherence    & 6  & Sentence similarity, topic drift \\
Readability  & 6  & Flesch, FK-grade, ARI, SMOG \\
Stylometric  & 8  & Punctuation, pronouns, caps ratio \\
\midrule
\textbf{Total}    & \textbf{62} & \\
\textbf{Selected} & \textbf{30} & Via MI on training data \\
\bottomrule
\end{tabular}}
\caption{Handcrafted feature categories ($m{=}62$ total; $k{=}30$ selected via Mutual Information).}
\label{tab:features_categories}
\end{table}

\Needspace{9\baselineskip}
\paragraph{Feature Normalisation and Selection.}
\label{app:feature_selection}
Feature selection is performed exclusively on source-side training data to avoid target-domain leakage. All features are normalised using \texttt{RobustScaler} fitted on source training data only. The final subset ($k{=}30$) maximises Mutual Information with the binary human/AI label on the full training mix.

\paragraph{Complete Feature List (62).}
Table~\ref{tab:full_feature_list} lists all 62 features with descriptions, organised by category.

\begin{table*}[p]
\centering
\small
\setlength{\tabcolsep}{4pt}
\renewcommand{\arraystretch}{1.06}

\begin{minipage}[t]{0.48\textwidth}
\centering
\fitpage{%
\begin{tabular}{l >{\raggedright\arraybackslash}p{3.9cm}}
\toprule
\textbf{Feature} & \textbf{Description} \\
\midrule
\multicolumn{2}{l}{\textit{Perplexity (8)}} \\
\texttt{ppl\_text}         & Document-level GPT-2 perplexity \\
\texttt{ppl\_mean\_token}  & Mean token-level perplexity \\
\texttt{ppl\_max\_token}   & Maximum token perplexity \\
\texttt{ppl\_std\_token}   & Std.\ of token perplexity \\
\texttt{ppl\_skew\_token}  & Skewness of token perplexity \\
\texttt{ppl\_low\_ratio}   & Fraction of low-perplexity tokens \\
\texttt{ppl\_high\_ratio}  & Fraction of high-perplexity tokens \\
\texttt{ppl\_trend}        & Perplexity trend across sentences \\
\midrule
\multicolumn{2}{l}{\textit{Entropy (10)}} \\
\texttt{ent\_char}         & Character-level Shannon entropy \\
\texttt{ent\_word}         & Word-level Shannon entropy \\
\texttt{ent\_bigram}       & Bigram entropy \\
\texttt{ent\_trigram}      & Trigram entropy \\
\texttt{ent\_conditional}  & Conditional word entropy \\
\texttt{ent\_rate}         & Entropy rate across document \\
\texttt{ent\_pos}          & Part-of-speech tag entropy \\
\texttt{ent\_punct}        & Punctuation entropy \\
\texttt{ent\_sent\_len}    & Sentence-length entropy \\
\texttt{ent\_word\_len}    & Word-length entropy \\
\midrule
\multicolumn{2}{l}{\textit{Burstiness (6)}} \\
\texttt{burst\_word}       & Word occurrence burstiness \\
\texttt{burst\_sent\_len}  & Sentence-length burstiness \\
\texttt{burst\_punct}      & Punctuation burstiness \\
\texttt{burst\_memory}     & Inter-arrival memory coefficient \\
\texttt{burst\_fano}       & Fano factor of word frequency \\
\texttt{burst\_iat\_var}   & Inter-arrival time variance \\
\midrule
\multicolumn{2}{l}{\textit{Repetition (10)}} \\
\texttt{rep\_self\_bleu2}      & Self-BLEU at bigram level \\
\texttt{rep\_self\_bleu3}      & Self-BLEU at trigram level \\
\texttt{rep\_self\_bleu4}      & Self-BLEU at 4-gram level \\
\texttt{rep\_unique\_bigram}   & Unique bigram ratio \\
\texttt{rep\_unique\_trigram}  & Unique trigram ratio \\
\texttt{rep\_bigram\_rate}     & Bigram occurrence rate \\
\texttt{rep\_trigram\_rate}    & Trigram occurrence rate \\
\texttt{rep\_max\_ngram\_freq} & Maximum n-gram frequency \\
\texttt{rep\_compression}      & Text compression ratio (zlib) \\
\texttt{rep\_dup\_sent}        & Duplicate sentence ratio \\
\bottomrule
\end{tabular}}
\end{minipage}
\hfill
\begin{minipage}[t]{0.48\textwidth}
\centering
\begin{tabular}{@{}l >{\raggedright\arraybackslash}p{3.5cm}@{}}
\toprule
\textbf{Feature} & \textbf{Description} \\
\midrule
\multicolumn{2}{l}{\textit{Vocabulary (8)}} \\
\texttt{vocab\_zipf\_coef}       & Zipf law coefficient \\
\texttt{vocab\_yules\_k}         & Yule's K richness measure \\
\texttt{vocab\_hapax}            & Hapax legomena ratio \\
\texttt{vocab\_dis}              & Dis legomena ratio \\
\texttt{vocab\_heaps}            & Heaps' law exponent \\
\texttt{vocab\_ttr}              & Type-token ratio (TTR) \\
\texttt{vocab\_mattr}            & Moving-average TTR (MATTR) \\
\texttt{vocab\_lexical\_density} & Content word ratio \\
\midrule
\multicolumn{2}{l}{\textit{Coherence (6)}} \\
\texttt{coh\_sent\_sim\_mean}    & Mean adjacent sentence similarity \\
\texttt{coh\_sent\_sim\_var}     & Variance of sentence similarity \\
\texttt{coh\_topic\_drift}       & Topic shift across sentences \\
\texttt{coh\_self\_sim}          & Global document self-similarity \\
\texttt{coh\_semantic\_density}  & Average semantic density \\
\texttt{coh\_consistency}        & Cross-sentence consistency score \\
\midrule
\multicolumn{2}{l}{\textit{Readability (6)}} \\
\texttt{read\_flesch}      & Flesch Reading Ease \\
\texttt{read\_fk\_grade}   & Flesch-Kincaid grade level \\
\texttt{read\_ari}         & Automated Readability Index \\
\texttt{read\_coleman}     & Coleman-Liau index \\
\texttt{read\_smog}        & SMOG grade \\
\texttt{read\_dale\_chall} & Dale-Chall readability score \\
\midrule
\multicolumn{2}{l}{\textit{Stylometric (8)}} \\
\texttt{style\_func\_word}    & Function word ratio \\
\texttt{style\_pronoun}       & Pronoun usage ratio \\
\texttt{style\_conjunction}   & Conjunction density \\
\texttt{style\_avg\_word\_len}& Mean word length (characters) \\
\texttt{style\_avg\_sent\_len}& Mean sentence length (words) \\
\texttt{style\_sent\_len\_var}& Sentence length variance \\
\texttt{style\_punct\_ratio}  & Punctuation density \\
\texttt{style\_cap\_ratio}    & Capitalisation ratio \\
\bottomrule
\end{tabular}
\end{minipage}

\caption{Complete list of all 62 handcrafted features with
descriptions, organised by category. Left: Perplexity,
Entropy, Burstiness, Repetition (38 features). Right:
Vocabulary, Coherence, Readability, Stylometric (28 features).
The top $k{=}30$ are selected via Mutual Information
(Section~\ref{app:feature_selection}).}
\label{tab:full_feature_list}
\label{tab:features_left}
\label{tab:features_right}
\end{table*}

%=================================================
\Needspace{7\baselineskip}
\subsubsection{Contrastive Learning and Paraphrase Augmentation}
\label{app:conpara}

To improve robustness against semantic-preserving rewriting, we additionally investigate paraphrase augmentation and supervised contrastive learning (SupCon). Paraphrases are generated offline using \texttt{flan-t5-base}~\citep{Chung22-FlanT5}. Generated samples are filtered using a cosine-similarity threshold of $\geq 0.85$ over sentence embeddings. Following~\citet{Krishna23-Paraphrase}, this value was selected to balance semantic fidelity against lexical diversity: thresholds below 0.80 admitted meaning-altering rewrites that degraded classification signal, while thresholds above 0.90 produced near-identical paraphrases with insufficient diversity to improve robustness. For supervised contrastive learning, positive pairs consist of original and paraphrased samples from the same class, while in-batch samples from opposite classes serve as negatives. The SupCon objective is combined with weighted cross-entropy loss:
\begin{equation}
\mathcal{L}
=
\mathcal{L}_{\mathrm{CE}}
+
\lambda\,\mathcal{L}_{\mathrm{SupCon}},
\label{eq:total_loss}
\end{equation}
where $\lambda \in \{0.05,0.1,0.2\}$ is tuned experimentally.

%=================================================
\subsubsection{Training Hyperparameters}
\label{app:hyperparameters}

All experiments use the same configuration. Hyperparameters are selected and kept fixed to ensure consistent comparison.

\begin{table}[H]
\centering
\small
\setlength{\tabcolsep}{5pt}
\renewcommand{\arraystretch}{1.0}
\fitcol{%
\begin{tabular}{lc}
\toprule
Parameter & Value \\
\midrule
Backbone             & DeBERTa-v3-large \\
Max sequence length  & 512 \\
Batch size           & 32 \\
Optimiser            & AdamW \\
Learning rate        & $2\times10^{-5}$ \\
Weight decay         & 0.01 \\
Scheduler            & OneCycleLR + cosine annealing \\
Dropout              & 0.1 \\
Gradient clipping    & 1.0 \\
SupCon temperature   & 0.07 \\
Hardware             & NVIDIA A100-SXM4-40GB \\
Framework            & PyTorch + Transformers \\
\bottomrule
\end{tabular}}
\caption{Training Hyperparameters.}
\label{tab:hyperparameters}
\end{table}

%%%%%%%%%%%%%%%%%%%%%%%%%%%%%%%%%%%%%%%%%%%%%%%%%%%%%%%%%%%%
%─────────────────────────────────────────────────
% C. ADDITIONAL EXPERIMENTAL RESULTS
%─────────────────────────────────────────────────
\subsection{Additional Experimental Results}
\label{app:experiments}

This section reports the out-of-distribution evaluation, the semantic-invariance (ConPara) ablation and the multi-seed check carried out on the submitted version of the system; the version sequence itself is in Appendix~\ref{app:versions}.

%=================================================

\begin{table*}[t]
\centering
\small
\setlength{\tabcolsep}{6pt}
\renewcommand{\arraystretch}{1.05}
\fitpage{%
\begin{tabular}{lrcccc}
\toprule
Dataset & $N$ & BA (\%) & AI Rec. (\%) 
        & Hum.\ Rec. (\%) & AUROC (\%) \\
\midrule
Gemini 2.0 Flash HC3~\citep{Mady25-HC3Gemini}
    & 23{,}463  & 98.64 & 98.64 & n/a & n/a \\
Academic Text~\citep{Mady25-AcademicText}
    & 669{,}008 & 94.69 & 99.98 & 89.41 & 99.90 \\
\bottomrule
\end{tabular}}
\caption{Out-of-distribution evaluation under 
fixed-threshold protocol ($\tau^*=0.920$, 
DeBERTa-ConPara v2.4-3k) without target-domain 
re-calibration. The Gemini set contains AI-generated text only, so human recall and AUROC are undefined (n/a).}
\label{tab:ood_results}
\end{table*}

%=================================================

\subsubsection{Extended Out-of-Distribution Evaluation}
\label{app:ood_extended}
\label{app:external}
\label{app:academic_text}

To further assess generalisation beyond the training 
distribution, we evaluate DeBERTa-ConPara v2.4-3k 
($\tau^*=0.920$) on two large-scale benchmarks under 
the fixed-threshold protocol without any target-domain 
re-calibration. Table~\ref{tab:ood_results} summarises 
the results.

DeBERTa-ConPara achieves strong generalisation across 
diverse unseen distributions. On 23K Gemini 2.0 Flash 
HC3-style responses~\citep{Mady25-HC3Gemini}, the 
detector achieves 98.64\,\% recall despite Gemini 
outputs being entirely absent from training data. 
On the 669K academic benchmark spanning human arXiv 
paragraphs and GPT-3.5/Gemini-generated scientific 
prose~\citep{Mady25-AcademicText}, the model achieves 
94.69\,\% balanced accuracy and 99.90\,\% AUROC, with 
near-perfect AI recall (99.98\,\%) and 89.41\,\% 
human recall.

%=================================================
\subsubsection{ConPara Ablations}
\label{app:conpara_ablation}

To evaluate the contribution of semantic-invariance objectives, we investigate supervised contrastive learning (SupCon) and paraphrasing augmentation under the heterogeneous multi-domain setting used throughout this work.

\begin{table*}[t]
\centering
\small
\setlength{\tabcolsep}{5pt}
\renewcommand{\arraystretch}{1.05}

\fitpage{%
\begin{tabular}{lccccc}
\toprule
Configuration & HC3-QA & HC3-SI & MAGE & Average & RAID TPR@5\,\% \\
\midrule

v2 (no ConPara)
& 99.43\,\%
& 82.18\,\%
& 92.84\,\%
& 91.91\,\%
& 97.11\,\%
\\

v2.2 (baseline)
& 99.56\,\%
& 84.04\,\%
& 93.15\,\%
& 92.25\,\%
& \textbf{97.65\,\%}
\\

$+$ SupCon ($\lambda=0.05$)
& 99.24\,\%
& 85.45\,\%
& 90.10\,\%
& 91.60\,\%
& n/s
\\

$+$ SupCon ($\lambda=0.1$)
& 99.42\,\%
& 85.01\,\%
& 92.59\,\%
& 92.34\,\%
& 97.18\,\%
\\

$+$ SupCon ($\lambda=0.2$)
& 99.32\,\%
& 84.97\,\%
& 90.78\,\%
& 91.69\,\%
& n/s
\\

$+$ Offline paraphrasing (T5)
& 99.20\,\%
& 84.51\,\%
& 90.35\,\%
& 91.35\,\%
& n/s
\\

v3 (SupCon $+$ paraphrasing)
& 99.14\,\%
& 84.35\,\%
& 89.30\,\%
& 90.93\,\%
& 95.88\,\%
\\

\bottomrule
\end{tabular}}

\caption{Ablation analysis of supervised contrastive learning and paraphrasing augmentation. Semantic-invariance objectives moderately improve HC3-SI but consistently reduce MAGE and RAID robustness under heterogeneous adversarial training conditions. n/s: variant not submitted to the RAID hidden test.}

\label{tab:conpara_ablation}
\end{table*}

\Needspace{9\baselineskip}
Supervised contrastive objectives produce only modest gains on semantic-invariant HC3 evaluation (+0.47 to +1.41\,pp), while consistently degrading robustness on MAGE and RAID. The strongest HC3-SI performance (85.45\,\%) is achieved with SupCon ($\lambda=0.05$), but this configuration simultaneously reduces MAGE performance by more than three percentage points.

Similarly, offline paraphrasing augmentation slightly improves HC3-SI robustness but substantially decreases cross-domain adversarial robustness. The combined v3 configuration (SupCon $+$ paraphrasing) produces the largest overall degradation, reducing RAID TPR@5\,\% FPR from 97.65\,\% to 95.88\,\%.

These results suggest that semantic-invariance objectives partially conflict with the AI-vs-human discrimination objective under heterogeneous multi-generator settings: contrastive clustering of semantically related samples can reduce sensitivity to subtle stylistic and statistical signals that remain informative for adversarial detection.

%=================================================
\subsubsection{Multi-Seed Stability}
\label{app:multiseed}

To evaluate robustness against random initialisation and stochastic optimisation effects, we repeat v2.2 training across five random seeds and evaluate on a held-out RAID subset containing 53,738 samples.

\begin{table}[H]
\centering
\small
\setlength{\tabcolsep}{6pt}
\renewcommand{\arraystretch}{1.05}

\fitcol{%
\begin{tabular}{lcc}
\toprule
Metric & Mean & Std. Dev. \\
\midrule

Balanced Accuracy & 92.20\,\% & $\pm0.75$\,pp \\
AUROC             & 94.53\,\% & $\pm1.20$\,pp \\

\bottomrule
\end{tabular}}

\caption{
Multi-seed stability results for DeBERTa-ConPara v2.2 on a 53k held-out RAID evaluation subset.}

\label{tab:multiseed}
\end{table}

The results demonstrate that the observed robustness 
improvements are stable across repeated training runs, 
with relatively low variance despite the highly 
heterogeneous multi-domain and multi-generator setting.

In particular, the low balanced-accuracy variance ($\pm0.75$\,pp) suggests that attack-aware preprocessing and adversarial data scaling produce consistent robustness gains that are not strongly dependent on favourable random initialisation. The AUROC variance ($\pm1.20$\,pp) is slightly higher, consistent with known sensitivity of ranking-based metrics to probability calibration across seeds. Together, these results indicate that the primary robustness gains are attributable to architectural and data-centric design choices rather than fortunate random initialisation.

%=================================================
%=================================================
%=================================================
%=================================================

\Needspace{12\baselineskip}
\section{Factorial Ablation: Extended Results}
\label{app:factorial_detail}

\paragraph{Submission provenance.}
All eight cells were submitted individually to the RAID hidden test between
16 and 25 August 2026. The naming convention in the leaderboard history is:
\texttt{v218} denotes a model trained on the normalised corpus and
\texttt{v218b} one trained on raw text; \texttt{nofeat} denotes removal of the
feature-fusion branch; and the suffix records the inference-time pipeline.
Table~\ref{tab:factorial_full} maps each submission to its cell.

\begin{table}[H]
\centering\small
\setlength{\tabcolsep}{3pt}
\fitcol{%
\begin{tabular}{@{}lccc@{}}
\toprule
\textbf{Submission} & \textbf{Feat} & \textbf{Train} & \textbf{Infer} \\
\midrule
\texttt{v2.18}                  & yes & norm & norm \\
\texttt{v218-rawinf}            & yes & norm & raw  \\
\texttt{v2.18b-raw}             & yes & raw  & raw  \\
\texttt{v2.18b-norm}            & yes & raw  & norm \\
\texttt{v218-nofeat-norm}       & no  & norm & norm \\
\texttt{v218-nofeat-rawinf}     & no  & norm & raw  \\
\texttt{v218b-nofeat-raw}       & no  & raw  & raw  \\
\texttt{v218b-nofeat-norminf}   & no  & raw  & norm \\
\bottomrule
\end{tabular}}
\caption{Mapping from leaderboard submission to factorial cell.}
\label{tab:factorial_full}
\end{table}

\Needspace{6\baselineskip}
\paragraph{Interaction between the two placements.}
The two preprocessing placements are not independent. Averaged over the
feature factor, inference-time normalisation is worth $+13.45$\,pp after
normalised-corpus training but only $+1.94$\,pp after raw training;
equivalently, raw training is worth $+16.18$\,pp under raw inference and
$+4.66$\,pp under normalised inference.

Each intervention substitutes for the
other: whichever is applied first captures most of the available gain, and the
second adds comparatively little. What is not viable is applying normalisation
at training time only, which removes the adversarial supervision and leaves
nothing at inference to compensate.

\paragraph{Clean-text performance.}
The unattacked column separates attack robustness from baseline quality. On
unattacked rows alone the eight cells span $88.38$ to $97.17\,\%$ TPR@1\,\%
FPR, and the ordering differs from the aggregate: the best clean score belongs
to raw training with raw inference ($97.17$), not to the configuration we
report ($96.98$). The configuration we report wins on the aggregate because
its attack penalty is the smallest of the eight ($0.41$\,pp).

\paragraph{The feature branch under every condition.}
The branch reduces TPR@1\,\% FPR in all four train/inference combinations, by
$7.07$, $3.89$, $1.85$ and $0.14$\,pp. Its cost is largest in the fully
normalised configuration, the one used in the submitted version of this
paper, and shrinks as raw training is introduced, consistent with the
branch supplying a signal that adversarial supervision provides more reliably.

\label{app:factorial}

\paragraph{Neither half of the defence is sufficient alone.}
The two components decompose cleanly. For homoglyph attacks, detection rises
from $11.05\,\%$ with no defence to $85.78\,\%$ with raw-corpus training alone,
and to $96.98\,\%$ when inference-time normalisation is added. For zero-width
space the corresponding figures are $1.12\,\%$, $73.80\,\%$ and $96.98\,\%$.
Adversarial supervision does most of the work; normalisation closes the
remainder. Notably, inference-time normalisation slightly \emph{reduces}
performance on nine of the twelve attack classes (by $0.09$ to $0.24$\,pp); it
is worth applying only because the two classes it repairs are worth $11$ and
$23$ points.

\paragraph{The mismatch penalty is asymmetric.}
Withholding preprocessing at test time from a model trained with it costs
$0.108$ of mean submitted score across the hidden test. Adding preprocessing
at test time to a model trained without it costs $0.007$, fifteen times
less and in the beneficial direction. This asymmetry is the signature of
adversarial training: a model that has seen perturbations is undisturbed when
they are removed, because the clean text was in its training set too, while a
model that has never seen them is badly disturbed when they appear.

\paragraph{Generalisation beyond this architecture.}
The same two-attack signature appears in a detector with no trained
classification head and a different tokeniser. Scoring the identical
validation rows with Fast-DetectGPT~\citep{Bao24-FastDetectGPT} using a
GPT-Neo-2.7B scoring model, inference-time normalisation raises AUROC on
zero-width-space attacks from $0.604$ to $0.875$ and on homoglyph attacks from
$0.683$ to $0.873$, while the remaining ten classes move by between $-0.025$
and $+0.005$. The effect therefore belongs to the attacks and to the
tokenisation failures they induce, not to our architecture.

\section{Backbone Comparison: Extended Results}

\begin{table}[t]
\centering\small
\fitcol{%
\begin{tabular}{@{}lccccc@{}}
\toprule
\textbf{Backbone} & \textbf{AUROC} & \textbf{T5} & \textbf{T1}
 & \textbf{T1$_{\text{cl}}$} & \textbf{pen} \\
\midrule
DeBERTa-v3-large & \pc{99.61} & \pc{99.01} & \pc{96.57} & \pc{96.98} & 0.41 \\
BERT-large       & \pc{99.07} & \pc{96.77} & \pc{90.54} & \pc{92.31} & 1.77 \\
RoBERTa-large    & \pc{94.32} & \pc{87.08} & \pc{71.23} & \pc{69.68} & $-$1.55 \\
\bottomrule
\end{tabular}}
\caption{Three backbones, identical corpus, hyperparameters and seed, all
raw-trained with normalised inference. T5 and T1 are TPR at $5\,\%$ and
$1\,\%$ FPR over all rows; T1$_{\text{cl}}$ is TPR@1\,\% FPR over unattacked
rows only; pen $=$ T1$_{\text{cl}}-$T1 is the attack penalty, i.e.\ how much
adversarial perturbation costs. RoBERTa is the only system whose penalty is
negative, scoring \emph{worse} on unattacked text than overall, which
follows from its collapse on one domain rather than from any attack.}
\label{tab:backbones}
\end{table}

Table~\ref{tab:backbones} extends the backbone comparison of
Section~\ref{subsec:backbones}; the two observations below explain why the
attack is not specific to DeBERTa and why validation-set ranking is an
unreliable guide to the hidden test.

\label{app:backbones}

\paragraph{The attack mechanism is architecture-general.}
All three tokenisers fragment homoglyph-substituted words into subword pieces
rather than emitting an unknown-token symbol: measured token inflation on a
matched sample is $3.40\times$ for BERT, $4.00\times$ for RoBERTa and
$4.12\times$ for DeBERTa, with zero unknown tokens in any of them. The
vulnerability that inference-time normalisation repairs is therefore a
property of subword tokenisation itself.

\paragraph{Validation ranking does not transfer to the hidden test.}
On our own RAID validation pool, RoBERTa reaches $85.29\,\%$ TPR@1\,\% FPR
against BERT's $76.09\,\%$, a nine-point lead. On the hidden test the
ordering reverses and the gap widens in the opposite direction: BERT reaches
$90.54\,\%$ and RoBERTa $71.23\,\%$. Selecting a backbone on validation would
have chosen the worse system by nineteen points.

Two measurements explain the instability. RoBERTa is the most overconfident of
the three, with a fitted temperature of $2.946$ against $2.204$ for DeBERTa
and $2.154$ for BERT, and correspondingly the widest bootstrap interval on
validation TPR@1\,\% FPR ($15.7$ points against $2.1$ for DeBERTa). Its
hidden-test failure is also domain-localised rather than uniform: per-domain
AUROC is $0.9913$ on news and $0.9996$ on recipes but $0.6562$ on abstracts,
and on books the ranking is sound ($0.9820$ AUROC) while the strict operating
point collapses ($35.45\,\%$ TPR@1\,\% FPR against $99.13\,\%$ at $5\,\%$).
A single global threshold is not viable for a model whose scores are that
compressed. We report this as a caution about single-split model selection at
low false-positive rates, and as motivation for the fixed-threshold protocol
used throughout this paper.

\section{Threshold Sensitivity}

\begin{figure*}[t]
\centering
\includegraphics[width=0.92\textwidth]{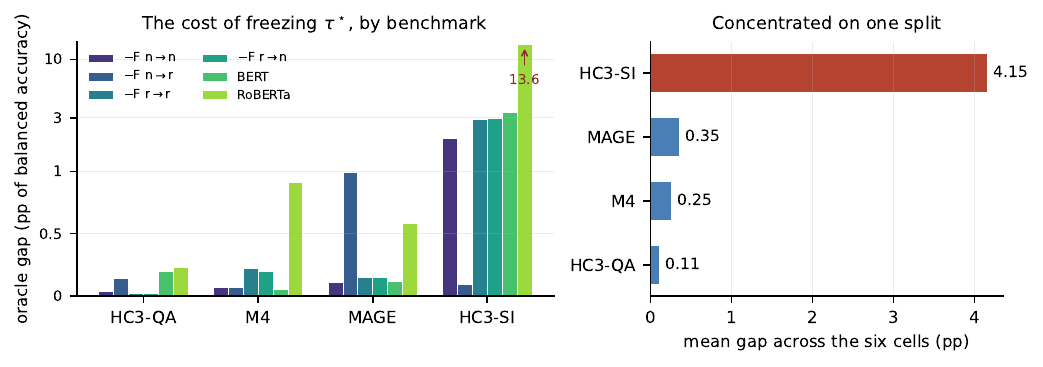}
\caption{Oracle gap, the balanced accuracy given up by fixing $\tau^\star$ in
advance rather than tuning per target, across $24$ cell-benchmark pairs.
The cost is under one point everywhere except HC3-SI. Per-cell sweep curves
are in Figure~\ref{fig:threshold_sweep}.}
\label{fig:threshold_gaps}
\end{figure*}

\begin{figure*}[t]
\centering
\includegraphics[width=0.92\textwidth]{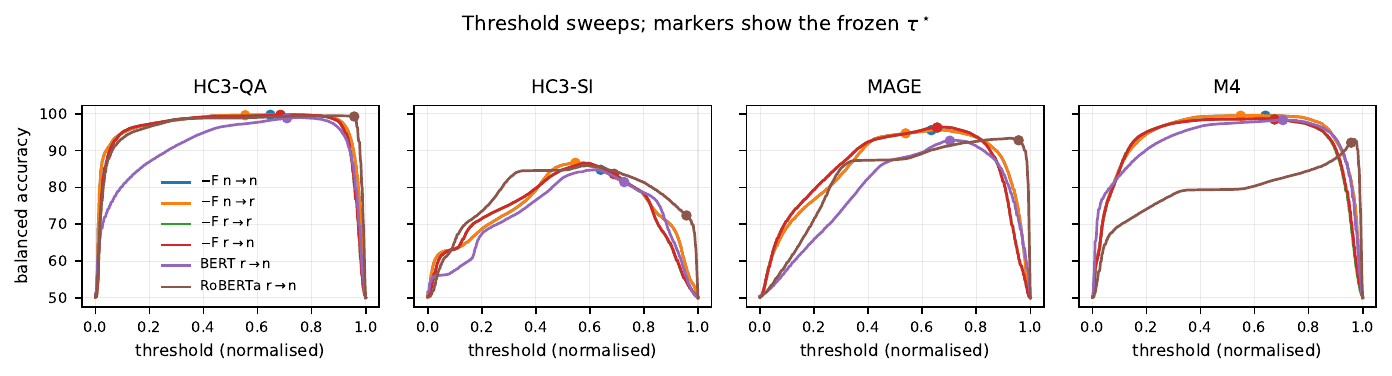}
\caption{Balanced accuracy as the decision threshold is swept across its
range, for each cell on each benchmark. Markers show where the frozen
$\tau^\star$ falls. HC3-QA and M4 give broad plateaux, so the frozen
threshold lands almost anywhere without cost; HC3-SI is sharply peaked for
every cell, which is where the oracle gap comes from.}
\label{fig:threshold_sweep}
\end{figure*}

\label{app:threshold}

\paragraph{What freezing the operating point costs.}
Because $\tau^\star$ is fixed before any target data is seen, the relevant
question is what that constraint costs. Sweeping the threshold on each
benchmark and comparing balanced accuracy at the frozen $\tau^\star$ against
the best achievable on that target gives an oracle gap for each of the $24$
cell-benchmark pairs (Figure~\ref{fig:threshold_gaps}).

The cost is negligible almost everywhere and concentrated on one split. Mean
gaps are $0.11$\,pp on HC3-QA, $0.25$ on M4 and $0.35$ on MAGE; on HC3-SI the
mean is $4.15$\,pp and the worst case $13.58$. Excluding HC3-SI, all $18$
remaining pairs fall between $0.02$ and $0.99$\,pp. Fixed-threshold evaluation
is therefore close to free on the benchmarks where scores are high, and
expensive exactly where meaning-preserving rewriting has already compressed
the score distribution.

We tested whether the gap is explained by the shape of the threshold curve,
as reported for a different model family~\citep{Mady26-FeatAug}: the width of
the band keeping balanced accuracy within one point of its maximum correlates
at $r=-0.35$ across all pairs, but the sign is inconsistent within
benchmarks ($-0.19$, $+0.48$, $+0.52$, $-0.90$ for the four splits), so that
account does not carry over here. The benchmark, not the curve, governs the
cost.

\paragraph{The configuration we report.}
For the released configuration the gap is $0.01$\,pp on HC3-QA, $0.10$\,pp on
MAGE and $0.25$\,pp on M4, so oracle access to the operating point is worth
almost nothing on three of the four benchmarks. HC3-SI is again the exception
at $2.97$\,pp, which localises the cost of the protocol to the split that is
hardest in absolute terms.

\Needspace{3\baselineskip}
\section{Version History}
\label{app:versions}

\begin{table*}[t]
\centering
\footnotesize
\setlength{\tabcolsep}{4pt}
\renewcommand{\arraystretch}{1.0}
\begin{tabular}{@{}lcccp{5.4cm}@{}}
\toprule
Version & TPR@5\,\% & TPR@1\,\% & AUROC & Change \\
\midrule
\textbf{Reported (Sec.~\ref{subsec:ablation})} & \textbf{99.01\,\%} & \textbf{96.57\,\%} & \textbf{99.61\,\%} & \texttt{deberta-v3-large}, no features, raw training, normalised inference, 1.55M rows \\
Full RAID (Sec.~\ref{subsec:scaling}) & 99.13\,\% & 97.12\,\% & 99.63\,\% & Same configuration on the complete RAID split (9.38M rows); loses 3.42\,pp TPR@1\,\% FPR on MAGE \\
\midrule
v1 & 94.29\,\% & 87.36\,\% & \textbf{98.58\,\%} & \texttt{deberta-v3-base}, feature fusion \\
v2 & 97.11\,\% & 93.34\,\% & 97.38\,\% & Attack-diverse training data \\
v2.2 & 97.65\,\% & 93.99\,\% & 97.87\,\% & Cohere balancing, revised RAID validation \\
v2.2-Seed42 & 97.29\,\% & 93.55\,\% & 96.86\,\% & v2.2, second seed \\
v2.3 (GDWL) & 97.69\,\% & 91.64\,\% & 97.41\,\% & Generator-difficulty weighted loss \\
v2.4-3k & \textbf{97.79\,\%} & \textbf{94.95\,\%} & 97.35\,\% & RAID scaled to 3,000 rows per cell \\
v2.4-5k & 96.70\,\% & 90.98\,\% & 96.01\,\% & 5,000 rows per cell, proportional validation (App.~\ref{app:validation}) \\
\midrule
A3 (norm.\ train, norm.\ inference) & 97.39\,\% & 92.05\,\% & 96.79\,\% & Submitted-version ablation arm, no features \\
A1 (norm.\ train, raw inference) & 84.00\,\% & 78.76\,\% & 89.78\,\% & A3 without inference-time normalisation \\
\bottomrule
\end{tabular}
\caption{RAID hidden-test results of the configuration we report, of the same configuration trained on the complete RAID split, and of the development sequence behind the submitted version (all on \texttt{deberta-v3-base}). TPR at 5\,\% and 1\,\% FPR. Bold marks the configuration we report and, within the development sequence, the best value in each column. A1 and A3 differ only by a train/inference mismatch; the controlled comparison of the two preprocessing placements is the eight-cell factorial of Section~\ref{subsec:ablation}.}
\label{tab:extended_raid}
\end{table*}

\begin{table}[!tb]
\centering\footnotesize\setlength{\tabcolsep}{4pt}\renewcommand{\arraystretch}{0.84}
\fitcol{%
\begin{tabular}{@{}llcc@{}}
\toprule
\textbf{Breakdown} & \textbf{Level} & \textbf{TPR@1\,\%} & \textbf{TPR@5\,\%} \\
\midrule
\multirow{12}{*}{Attack} & alternative\_spelling & \pc{97.13} & \pc{99.03} \\
 & article\_deletion & \pc{97.87} & \pc{99.09} \\
 & homoglyph & \pc{96.98} & \pc{99.04} \\
 & insert\_paragraphs & \pc{96.98} & \pc{99.04} \\
 & none & \pc{96.98} & \pc{99.04} \\
 & number & \pc{97.08} & \pc{99.09} \\
 & paraphrase & \pc{89.61} & \pc{98.33} \\
 & perplexity\_misspelling & \pc{97.41} & \pc{99.03} \\
 & synonym & \pc{96.78} & \pc{99.10} \\
 & upper\_lower & \pc{98.10} & \pc{99.27} \\
 & whitespace & \pc{96.98} & \pc{99.04} \\
 & zero\_width\_space & \pc{96.98} & \pc{99.04} \\
\midrule
\multirow{8}{*}{Domain} & abstracts & \pc{96.12} & \pc{98.59} \\
 & books & \pc{98.72} & \pc{99.50} \\
 & news & \pc{95.62} & \pc{99.02} \\
 & poetry & \pc{94.48} & \pc{98.75} \\
 & recipes & \pc{99.45} & \pc{99.84} \\
 & reddit & \pc{93.45} & \pc{98.12} \\
 & reviews & \pc{97.83} & \pc{99.46} \\
 & wiki & \pc{96.91} & \pc{98.83} \\
\midrule
\multirow{11}{*}{Generator} & chatgpt & \pc{99.52} & \pc{99.72} \\
 & cohere & \pc{89.83} & \pc{96.86} \\
 & cohere-chat & \pc{92.48} & \pc{98.04} \\
 & gpt2 & \pc{97.62} & \pc{99.50} \\
 & gpt3 & \pc{94.57} & \pc{98.29} \\
 & gpt4 & \pc{98.94} & \pc{99.79} \\
 & llama-chat & \pc{99.64} & \pc{99.96} \\
 & mistral & \pc{93.90} & \pc{98.32} \\
 & mistral-chat & \pc{98.74} & \pc{99.67} \\
 & mpt & \pc{94.96} & \pc{98.11} \\
 & mpt-chat & \pc{98.35} & \pc{99.71} \\
\midrule
\multirow{2}{*}{Decoding} & greedy & \pc{98.26} & \pc{99.51} \\
 & sampling & \pc{94.89} & \pc{98.52} \\
\midrule
\multirow{2}{*}{Rep.\ penalty} & no & \pc{96.03} & \pc{98.88} \\
 & yes & \pc{97.57} & \pc{99.25} \\
\bottomrule
\end{tabular}}
\caption{Cell \textbf{$-$F\,\,r$\to$n}: without the feature branch, raw training, normalised inference; the configuration we report.}
\label{tab:cell_Nrn}
\end{table}

% Floats of Appendix G, placed early so they fill page tops above the running text.
\begin{table*}[tb]
\centering\small\setlength{\tabcolsep}{2.5pt}\renewcommand{\arraystretch}{1.05}
\fitpage{%
\begin{tabular}{@{}lcccccccc@{}}
\toprule
\textbf{Attack} & \textbf{$+$F\,\,n$\to$n} & \textbf{$+$F\,\,n$\to$r} & \textbf{$+$F\,\,r$\to$r} & \textbf{$+$F\,\,r$\to$n} & \textbf{$-$F\,\,n$\to$n} & \textbf{$-$F\,\,n$\to$r} & \textbf{$-$F\,\,r$\to$r} & \textbf{$-$F\,\,r$\to$n} \\
\midrule
alternative\_spelling & \pc{88.8} & \pc{88.4} & \pc{95.5} & \pc{95.0} & \pc{95.6} & \pc{95.5} & \textbf{\pc{97.3}} & \pc{97.1} \\
article\_deletion & \pc{89.1} & \pc{88.8} & \pc{95.9} & \pc{95.3} & \pc{96.3} & \pc{96.1} & \textbf{\pc{98.0}} & \pc{97.9} \\
homoglyph & \pc{88.8} & \pc{32.8} & \pc{89.7} & \pc{95.3} & \pc{95.4} & \pc{11.0} & \pc{85.8} & \textbf{\pc{97.0}} \\
insert\_paragraphs & \pc{89.3} & \pc{88.7} & \pc{96.4} & \pc{96.2} & \pc{95.4} & \pc{95.3} & \textbf{\pc{97.2}} & \pc{97.0} \\
none & \pc{88.8} & \pc{88.4} & \pc{95.8} & \pc{95.3} & \pc{95.4} & \pc{95.3} & \textbf{\pc{97.2}} & \pc{97.0} \\
number & \pc{88.7} & \pc{88.2} & \pc{95.8} & \pc{95.3} & \pc{95.4} & \pc{95.3} & \textbf{\pc{97.3}} & \pc{97.1} \\
paraphrase & \pc{77.6} & \pc{75.9} & \pc{90.1} & \textbf{\pc{90.4}} & \pc{83.1} & \pc{82.8} & \pc{88.9} & \pc{89.6} \\
perplexity\_misspelling & \pc{89.0} & \pc{88.5} & \pc{95.8} & \pc{95.3} & \pc{95.9} & \pc{95.7} & \textbf{\pc{97.5}} & \pc{97.4} \\
synonym & \pc{83.7} & \pc{83.0} & \pc{94.7} & \pc{93.8} & \pc{93.5} & \pc{93.3} & \textbf{\pc{97.0}} & \pc{96.8} \\
upper\_lower & \pc{87.8} & \pc{87.5} & \pc{95.2} & \pc{94.3} & \pc{97.2} & \pc{97.1} & \textbf{\pc{98.2}} & \pc{98.1} \\
whitespace & \pc{88.8} & \pc{86.0} & \pc{95.5} & \pc{95.3} & \pc{95.4} & \pc{95.3} & \textbf{\pc{97.2}} & \pc{97.0} \\
zero\_width\_space & \pc{88.8} & \pc{10.9} & \pc{83.3} & \pc{95.3} & \pc{95.4} & \pc{1.1} & \pc{73.8} & \textbf{\pc{97.0}} \\
\bottomrule
\end{tabular}}
\caption{TPR@1\,\% FPR by attack, all eight factorial cells (cell labels as defined in Appendix~\ref{app:factorial_full}). Best per row in bold.}
\label{tab:fact_attack_1}
\end{table*}

\begin{table*}[tb]
\centering\small\setlength{\tabcolsep}{2.5pt}\renewcommand{\arraystretch}{1.05}
\fitpage{%
\begin{tabular}{@{}lcccccccc@{}}
\toprule
\textbf{Domain} & \textbf{$+$F\,\,n$\to$n} & \textbf{$+$F\,\,n$\to$r} & \textbf{$+$F\,\,r$\to$r} & \textbf{$+$F\,\,r$\to$n} & \textbf{$-$F\,\,n$\to$n} & \textbf{$-$F\,\,n$\to$r} & \textbf{$-$F\,\,r$\to$r} & \textbf{$-$F\,\,r$\to$n} \\
\midrule
abstracts & \pc{97.5} & \pc{93.3} & \textbf{\pc{99.3}} & \pc{99.1} & \pc{95.2} & \pc{82.6} & \pc{95.4} & \pc{96.1} \\
books & \pc{90.2} & \pc{78.3} & \pc{91.5} & \pc{93.1} & \pc{96.2} & \pc{80.8} & \pc{95.5} & \textbf{\pc{98.7}} \\
news & \pc{94.5} & \pc{79.5} & \pc{94.1} & \textbf{\pc{96.4}} & \pc{94.5} & \pc{79.2} & \pc{89.2} & \pc{95.6} \\
poetry & \pc{38.1} & \pc{30.4} & \pc{83.7} & \pc{82.4} & \pc{87.9} & \pc{71.3} & \pc{91.5} & \textbf{\pc{94.5}} \\
recipes & \pc{99.5} & \pc{91.7} & \pc{99.5} & \textbf{\pc{99.6}} & \pc{99.1} & \pc{84.0} & \pc{99.3} & \pc{99.5} \\
reddit & \pc{89.8} & \pc{70.2} & \pc{94.1} & \textbf{\pc{95.4}} & \pc{90.9} & \pc{76.4} & \pc{91.5} & \pc{93.5} \\
reviews & \pc{94.7} & \pc{79.9} & \pc{94.5} & \pc{96.4} & \pc{96.0} & \pc{80.5} & \pc{95.1} & \textbf{\pc{97.8}} \\
wiki & \pc{95.4} & \pc{81.4} & \pc{92.4} & \pc{95.4} & \pc{96.4} & \pc{81.0} & \pc{92.8} & \textbf{\pc{96.9}} \\
\bottomrule
\end{tabular}}
\caption{TPR@1\,\% FPR by domain, all eight factorial cells. Best per row in bold.}
\label{tab:fact_domain_1}
\end{table*}

\begin{table*}[tb]
\centering\small\setlength{\tabcolsep}{2.5pt}\renewcommand{\arraystretch}{1.05}
\fitpage{%
\begin{tabular}{@{}lcccccccc@{}}
\toprule
\textbf{Generator} & \textbf{$+$F\,\,n$\to$n} & \textbf{$+$F\,\,n$\to$r} & \textbf{$+$F\,\,r$\to$r} & \textbf{$+$F\,\,r$\to$n} & \textbf{$-$F\,\,n$\to$n} & \textbf{$-$F\,\,n$\to$r} & \textbf{$-$F\,\,r$\to$r} & \textbf{$-$F\,\,r$\to$n} \\
\midrule
chatgpt & \pc{87.5} & \pc{75.6} & \pc{96.4} & \pc{95.1} & \pc{99.4} & \pc{83.1} & \pc{97.6} & \textbf{\pc{99.5}} \\
cohere & \pc{75.3} & \pc{64.3} & \pc{82.3} & \pc{84.4} & \pc{85.5} & \pc{71.1} & \pc{84.9} & \textbf{\pc{89.8}} \\
cohere-chat & \pc{79.5} & \pc{67.1} & \pc{89.7} & \pc{91.4} & \pc{90.4} & \pc{75.6} & \pc{89.3} & \textbf{\pc{92.5}} \\
gpt2 & \pc{95.1} & \pc{83.6} & \pc{94.3} & \pc{97.4} & \pc{93.6} & \pc{79.1} & \pc{93.6} & \textbf{\pc{97.6}} \\
gpt3 & \pc{82.6} & \pc{70.5} & \pc{91.5} & \pc{92.7} & \pc{90.0} & \pc{74.9} & \pc{92.6} & \textbf{\pc{94.6}} \\
gpt4 & \pc{84.4} & \pc{73.3} & \pc{95.1} & \pc{95.7} & \textbf{\pc{99.1}} & \pc{82.7} & \pc{94.8} & \pc{98.9} \\
llama-chat & \pc{88.1} & \pc{76.3} & \pc{97.5} & \pc{97.5} & \pc{99.4} & \pc{83.2} & \pc{97.8} & \textbf{\pc{99.6}} \\
mistral & \pc{88.1} & \pc{77.7} & \pc{90.0} & \pc{91.8} & \pc{90.3} & \pc{78.1} & \pc{91.2} & \textbf{\pc{93.9}} \\
mistral-chat & \pc{88.2} & \pc{75.5} & \pc{96.9} & \pc{97.6} & \pc{98.3} & \pc{82.3} & \pc{95.9} & \textbf{\pc{98.7}} \\
mpt & \pc{90.0} & \pc{78.3} & \pc{92.5} & \pc{93.5} & \pc{92.0} & \pc{77.6} & \pc{93.0} & \textbf{\pc{95.0}} \\
mpt-chat & \pc{89.2} & \pc{75.8} & \pc{97.2} & \pc{97.6} & \pc{97.7} & \pc{81.6} & \pc{96.0} & \textbf{\pc{98.4}} \\
\bottomrule
\end{tabular}}
\caption{TPR@1\,\% FPR by generator, all eight factorial cells. Best per row in bold.}
\label{tab:fact_model_1}
\end{table*}

\begin{table*}[tb]
\centering\small\setlength{\tabcolsep}{2.5pt}\renewcommand{\arraystretch}{1.05}
\fitpage{%
\begin{tabular}{@{}lcccccccc@{}}
\toprule
\textbf{Condition} & \textbf{$+$F\,\,n$\to$n} & \textbf{$+$F\,\,n$\to$r} & \textbf{$+$F\,\,r$\to$r} & \textbf{$+$F\,\,r$\to$n} & \textbf{$-$F\,\,n$\to$n} & \textbf{$-$F\,\,n$\to$r} & \textbf{$-$F\,\,r$\to$r} & \textbf{$-$F\,\,r$\to$n} \\
\midrule
\multicolumn{9}{@{}l}{\textit{TPR@1\,\% FPR, decoding}} \\
greedy & \pc{90.8} & \pc{79.3} & \pc{96.6} & \pc{96.8} & \pc{96.1} & \pc{81.3} & \pc{96.6} & \textbf{\pc{98.3}} \\
sampling & \pc{84.1} & \pc{71.9} & \pc{90.7} & \pc{92.6} & \pc{92.9} & \pc{77.7} & \pc{91.0} & \textbf{\pc{94.9}} \\
\midrule
\multicolumn{9}{@{}l}{\textit{TPR@1\,\% FPR, repetition penalty}} \\
no & \pc{85.1} & \pc{74.0} & \pc{92.4} & \pc{93.4} & \pc{93.3} & \pc{78.7} & \pc{93.2} & \textbf{\pc{96.0}} \\
yes & \pc{91.8} & \pc{78.5} & \pc{95.9} & \pc{97.2} & \pc{96.7} & \pc{81.0} & \pc{94.9} & \textbf{\pc{97.6}} \\
\midrule
\multicolumn{9}{@{}l}{\textit{TPR@5\,\% FPR, decoding}} \\
greedy & \pc{99.5} & \pc{94.2} & \pc{99.5} & \textbf{\pc{99.6}} & \pc{99.3} & \pc{87.7} & \pc{99.3} & \pc{99.5} \\
sampling & \pc{98.2} & \pc{91.3} & \pc{98.1} & \pc{98.5} & \pc{97.7} & \pc{84.6} & \pc{97.5} & \textbf{\pc{98.5}} \\
\midrule
\multicolumn{9}{@{}l}{\textit{TPR@5\,\% FPR, repetition penalty}} \\
no & \pc{98.7} & \pc{93.4} & \pc{98.6} & \pc{98.9} & \pc{98.2} & \pc{86.3} & \pc{98.2} & \textbf{\pc{98.9}} \\
yes & \pc{99.2} & \pc{91.6} & \pc{99.2} & \textbf{\pc{99.3}} & \pc{99.0} & \pc{85.8} & \pc{98.8} & \pc{99.2} \\
\bottomrule
\end{tabular}}
\caption{TPR by decoding strategy and repetition penalty at $1\,\%$ and $5\,\%$ FPR, all eight factorial cells. Best per row in bold.}
\label{tab:fact_decoding_1}
\end{table*}

\begin{table*}[tb]
\centering\small\setlength{\tabcolsep}{2.5pt}\renewcommand{\arraystretch}{1.05}
\fitpage{%
\begin{tabular}{@{}lcccccccc@{}}
\toprule
\textbf{Attack} & \textbf{$+$F\,\,n$\to$n} & \textbf{$+$F\,\,n$\to$r} & \textbf{$+$F\,\,r$\to$r} & \textbf{$+$F\,\,r$\to$n} & \textbf{$-$F\,\,n$\to$n} & \textbf{$-$F\,\,n$\to$r} & \textbf{$-$F\,\,r$\to$r} & \textbf{$-$F\,\,r$\to$n} \\
\midrule
alternative\_spelling & \pc{99.0} & \pc{99.0} & \textbf{\pc{99.2}} & \pc{99.1} & \pc{98.7} & \pc{98.7} & \pc{99.1} & \pc{99.0} \\
article\_deletion & \pc{99.0} & \pc{99.0} & \textbf{\pc{99.2}} & \pc{99.2} & \pc{98.7} & \pc{98.7} & \pc{99.2} & \pc{99.1} \\
homoglyph & \pc{99.0} & \pc{87.9} & \pc{97.9} & \textbf{\pc{99.1}} & \pc{98.7} & \pc{43.0} & \pc{96.4} & \pc{99.0} \\
insert\_paragraphs & \pc{98.9} & \pc{99.0} & \textbf{\pc{99.2}} & \pc{99.1} & \pc{98.7} & \pc{98.7} & \pc{99.1} & \pc{99.0} \\
none & \pc{99.0} & \pc{99.0} & \textbf{\pc{99.2}} & \pc{99.1} & \pc{98.7} & \pc{98.7} & \pc{99.1} & \pc{99.0} \\
number & \pc{99.0} & \pc{99.0} & \textbf{\pc{99.2}} & \pc{99.1} & \pc{98.7} & \pc{98.7} & \pc{99.1} & \pc{99.1} \\
paraphrase & \pc{97.5} & \pc{97.5} & \pc{98.2} & \pc{98.0} & \pc{96.6} & \pc{96.6} & \textbf{\pc{98.4}} & \pc{98.3} \\
perplexity\_misspelling & \pc{99.0} & \pc{99.0} & \textbf{\pc{99.2}} & \pc{99.1} & \pc{98.7} & \pc{98.7} & \pc{99.1} & \pc{99.0} \\
synonym & \pc{98.8} & \pc{98.8} & \textbf{\pc{99.3}} & \pc{99.2} & \pc{98.3} & \pc{98.3} & \pc{99.2} & \pc{99.1} \\
upper\_lower & \pc{99.1} & \pc{99.1} & \textbf{\pc{99.3}} & \pc{99.2} & \pc{98.9} & \pc{98.9} & \pc{99.3} & \pc{99.3} \\
whitespace & \pc{99.0} & \pc{98.9} & \textbf{\pc{99.1}} & \pc{99.1} & \pc{98.7} & \pc{98.7} & \pc{99.1} & \pc{99.0} \\
zero\_width\_space & \pc{99.0} & \pc{36.8} & \pc{96.9} & \textbf{\pc{99.1}} & \pc{98.7} & \pc{6.0} & \pc{93.7} & \pc{99.0} \\
\bottomrule
\end{tabular}}
\caption{TPR@5\,\% FPR by attack, all eight factorial cells. Best per row in bold.}
\label{tab:fact_attack_5}
\end{table*}

\begin{table*}[tb]
\centering\small\setlength{\tabcolsep}{2.5pt}\renewcommand{\arraystretch}{1.05}
\fitpage{%
\begin{tabular}{@{}lcccccccc@{}}
\toprule
\textbf{Domain} & \textbf{$+$F\,\,n$\to$n} & \textbf{$+$F\,\,n$\to$r} & \textbf{$+$F\,\,r$\to$r} & \textbf{$+$F\,\,r$\to$n} & \textbf{$-$F\,\,n$\to$n} & \textbf{$-$F\,\,n$\to$r} & \textbf{$-$F\,\,r$\to$r} & \textbf{$-$F\,\,r$\to$n} \\
\midrule
abstracts & \pc{98.8} & \pc{97.9} & \textbf{\pc{99.6}} & \pc{99.4} & \pc{98.0} & \pc{88.8} & \pc{98.7} & \pc{98.6} \\
books & \pc{99.2} & \pc{88.5} & \pc{97.9} & \pc{99.0} & \pc{98.9} & \pc{83.6} & \pc{97.9} & \textbf{\pc{99.5}} \\
news & \pc{98.8} & \pc{90.6} & \pc{99.5} & \textbf{\pc{99.6}} & \pc{98.4} & \pc{86.7} & \pc{97.6} & \pc{99.0} \\
poetry & \pc{98.5} & \pc{94.2} & \pc{97.8} & \pc{98.4} & \pc{98.6} & \pc{85.3} & \pc{98.4} & \textbf{\pc{98.8}} \\
recipes & \textbf{\pc{99.9}} & \pc{97.5} & \pc{99.9} & \pc{99.9} & \pc{99.9} & \pc{92.6} & \pc{99.8} & \pc{99.8} \\
reddit & \pc{97.8} & \pc{91.3} & \textbf{\pc{98.3}} & \pc{97.9} & \pc{97.2} & \pc{83.9} & \pc{97.6} & \pc{98.1} \\
reviews & \pc{99.2} & \pc{91.8} & \pc{98.9} & \pc{99.2} & \pc{98.7} & \pc{84.0} & \pc{99.0} & \textbf{\pc{99.5}} \\
wiki & \pc{98.5} & \pc{90.3} & \pc{98.8} & \textbf{\pc{98.9}} & \pc{98.3} & \pc{84.1} & \pc{98.1} & \pc{98.8} \\
\bottomrule
\end{tabular}}
\caption{TPR@5\,\% FPR by domain, all eight factorial cells. Best per row in bold.}
\label{tab:fact_domain_5}
\end{table*}

\begin{table*}[tb]
\centering\small\setlength{\tabcolsep}{2.5pt}\renewcommand{\arraystretch}{1.05}
\fitpage{%
\begin{tabular}{@{}lcccccccc@{}}
\toprule
\textbf{Generator} & \textbf{$+$F\,\,n$\to$n} & \textbf{$+$F\,\,n$\to$r} & \textbf{$+$F\,\,r$\to$r} & \textbf{$+$F\,\,r$\to$n} & \textbf{$-$F\,\,n$\to$n} & \textbf{$-$F\,\,n$\to$r} & \textbf{$-$F\,\,r$\to$r} & \textbf{$-$F\,\,r$\to$n} \\
\midrule
chatgpt & \pc{99.7} & \pc{93.1} & \textbf{\pc{99.7}} & \pc{99.7} & \pc{99.7} & \pc{86.0} & \pc{99.6} & \pc{99.7} \\
cohere & \pc{97.2} & \pc{91.1} & \pc{97.0} & \textbf{\pc{97.5}} & \pc{95.4} & \pc{82.3} & \pc{95.5} & \pc{96.9} \\
cohere-chat & \pc{97.8} & \pc{91.8} & \pc{97.9} & \pc{97.9} & \pc{97.1} & \pc{84.4} & \pc{97.3} & \textbf{\pc{98.0}} \\
gpt2 & \pc{99.5} & \pc{94.5} & \pc{98.8} & \pc{99.4} & \pc{99.2} & \pc{88.1} & \pc{98.3} & \textbf{\pc{99.5}} \\
gpt3 & \pc{98.1} & \pc{92.5} & \textbf{\pc{98.4}} & \pc{98.3} & \pc{97.2} & \pc{85.6} & \pc{98.2} & \pc{98.3} \\
gpt4 & \pc{99.8} & \pc{93.0} & \pc{99.7} & \textbf{\pc{99.9}} & \pc{99.7} & \pc{85.5} & \pc{99.1} & \pc{99.8} \\
llama-chat & \pc{99.9} & \pc{93.2} & \pc{99.9} & \pc{99.9} & \pc{99.9} & \pc{86.3} & \pc{99.8} & \textbf{\pc{100.0}} \\
mistral & \pc{97.6} & \pc{93.1} & \pc{97.7} & \pc{98.2} & \pc{97.2} & \pc{87.9} & \pc{97.2} & \textbf{\pc{98.3}} \\
mistral-chat & \pc{99.7} & \pc{93.0} & \pc{99.7} & \textbf{\pc{99.8}} & \pc{99.6} & \pc{86.3} & \pc{99.3} & \pc{99.7} \\
mpt & \pc{97.6} & \pc{92.0} & \pc{97.9} & \textbf{\pc{98.1}} & \pc{97.4} & \pc{85.8} & \pc{97.4} & \pc{98.1} \\
mpt-chat & \pc{99.6} & \pc{92.0} & \pc{99.6} & \pc{99.6} & \pc{99.4} & \pc{85.8} & \pc{99.4} & \textbf{\pc{99.7}} \\
\bottomrule
\end{tabular}}
\caption{TPR@5\,\% FPR by generator, all eight factorial cells. Best per row in bold.}
\label{tab:fact_model_5}
\end{table*}

The submitted version of this paper was built on \texttt{deberta-v3-base} and
went through the development sequence in Table~\ref{tab:extended_raid}, which
also lists the full-RAID run of Section~\ref{subsec:scaling}. The sequence is
reported for completeness; the factorial of Section~\ref{subsec:ablation}
supersedes it as ablation evidence. Three observations from the sequence still
hold.

\paragraph{The gains were data-centric.}
The largest single step, v1 to v2 ($+2.82$\,pp TPR@5\,\% FPR), changed
preprocessing, corpus composition and adversarial coverage together, and every
later step changed data or loss while the backbone stayed fixed. Scaling the
RAID training coverage from 1,671 to 3,000 rows per attack\,$\times$\,domain
cell (v2.2 to v2.4-3k) added $+0.14$\,pp at 5\,\% and $+0.96$\,pp at 1\,\%
FPR, with the largest gains on Reddit ($+1.78$\,pp) and the Cohere generators
($+0.99$ to $+1.27$\,pp).

\paragraph{Ranking and operating-point quality diverge.}
v2.2 has the highest AUROC of the sequence but v2.4-3k the best TPR at both
operating points, and the generator-weighted loss of v2.3 raised TPR@5\,\% FPR
slightly while cutting TPR@1\,\% FPR by $2.35$\,pp. AUROC alone hides
calibration behaviour at strict operating points, which is why the paper
reports fixed-threshold metrics throughout. The v2.4-5k row shows the
complementary trade-off: proportional validation improved HC3-SI
(Appendix~\ref{app:validation}) at the cost of RAID operating-point
robustness.

\paragraph{The A1 and A3 arms are not a measure of preprocessing.}
Both were trained on a corpus normalised at construction time, and A1 was then
evaluated on raw text, so their $13$\,pp gap is a mismatch between training
corpus and inference pipeline rather than the contribution of the pipeline.
The factorial isolates the same effect: its two cells trained on normalised
text and evaluated on raw lose $12.79$ and $15.80$\,pp to adversarial
perturbation.

\Needspace{14\baselineskip}
\section{Validation-Set Composition}
\label{app:validation}

\paragraph{Effect of validation-set composition.} Our initial v2.4-3k runs used a validation set in which MAGE made up 62.6\,\% of samples against $\sim$34\,\% of the training data; the resulting early-stopping bias cut HC3-SI from 84.04\,\% (v2.2) to 81.18\,\%. Correcting to proportional validation restored HC3-SI to \textbf{85.33\,\%} while preserving RAID robustness, so validation composition is a hyperparameter in its own right.

%%%%%%%%%%%%%%%%%%%%%%%%%%%%%%%%%%%%%%%%%%%%%%%%%%%%%%%%%%%%

\section{Factorial Ablation: Complete Breakdowns}
\label{app:factorial_full}

Cell labels: first letter, feature-fusion branch (\textbf{F}~present, \textbf{N}~absent); second and third, normalisation at training and at inference (\textbf{n}~normalised, \textbf{r}~raw). All numbers are TPR at the stated FPR on the $672{,}000$-row RAID hidden test.

\paragraph{Why the feature branch fails on poetry.}
Plausibly, the 30 features are surface statistics (sentence length,
punctuation density, burstiness, type-token ratio) whose
distributions poetry violates systematically, so the gated fusion supplies a
confidently wrong contribution.

\subsection{Per-Cell Detail}

Tables~\ref{tab:fact_attack_1} to~\ref{tab:fact_model_5} and~\ref{tab:fact_decoding_1} give all eight cells on every breakdown. Table~\ref{tab:cell_Nrn} reproduces, at full precision, only the configuration we report; the corresponding numbers for the other seven cells are the columns of those tables.

\section{Per-Cell Performance Matrices}
\label{app:percell}

The ten configurations are laid out two per page, each with three breakdowns
(generator$\times$domain, generator$\times$attack and domain$\times$attack) on
a colour scale fixed to $[0,100]$, so panels are comparable across pages. The
configuration we report appears again at the end, one breakdown per page at
full size.

\clearpage
\begin{figure*}[p]
\centering
\begin{minipage}[t]{0.485\textwidth}\centering
\includegraphics[width=\linewidth,keepaspectratio]{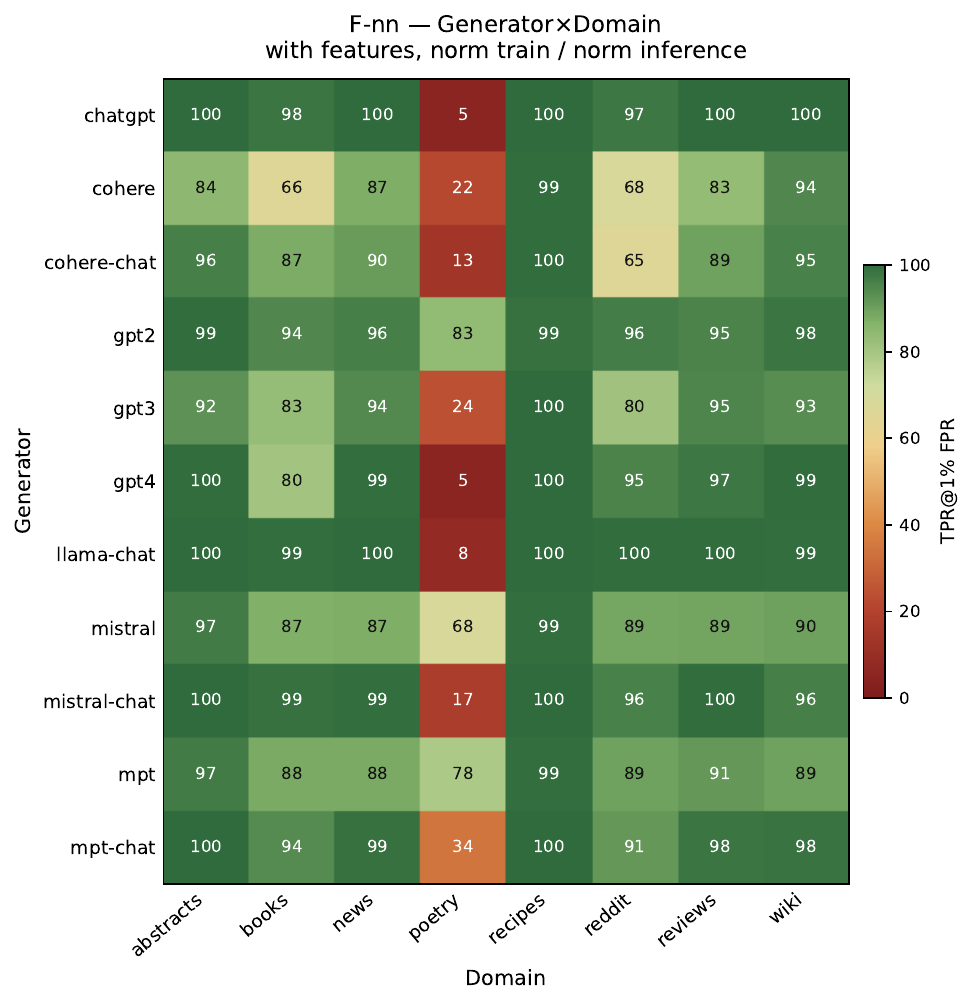}
\par\vspace{3pt}
\includegraphics[width=\linewidth,keepaspectratio]{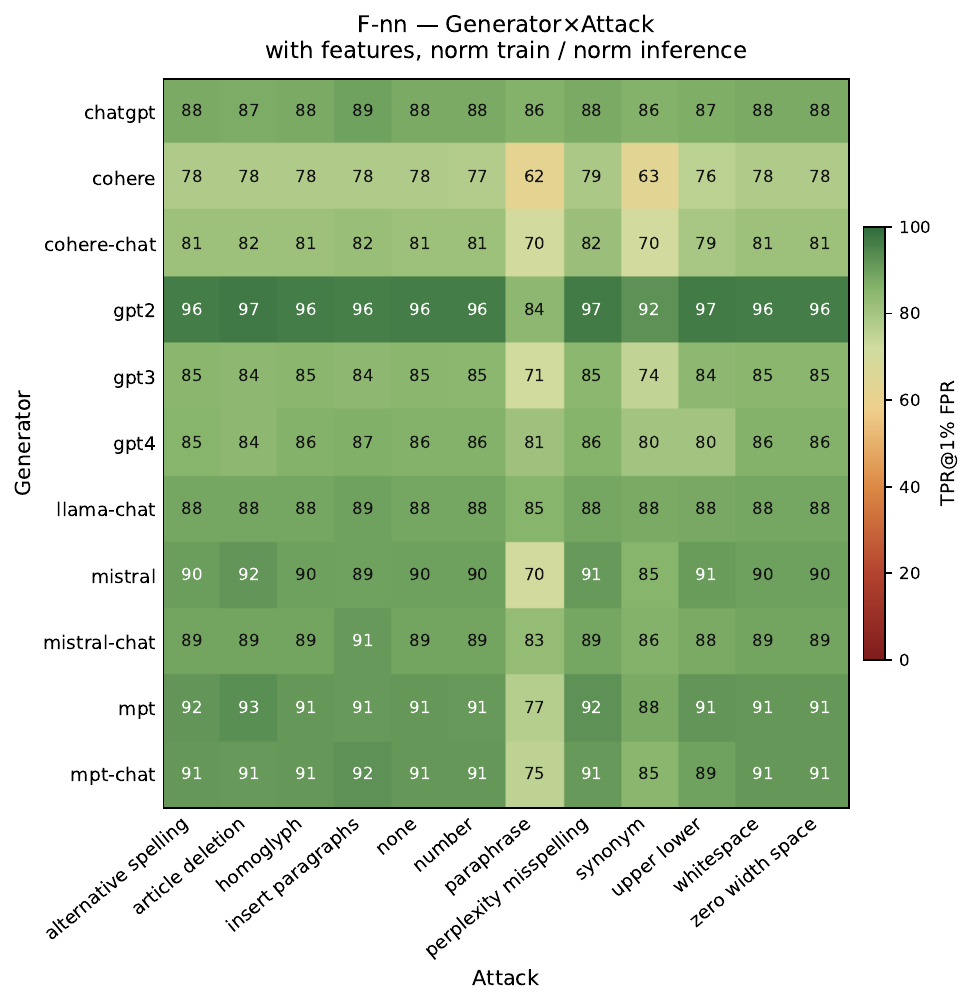}
\par\vspace{3pt}
\includegraphics[width=\linewidth,keepaspectratio]{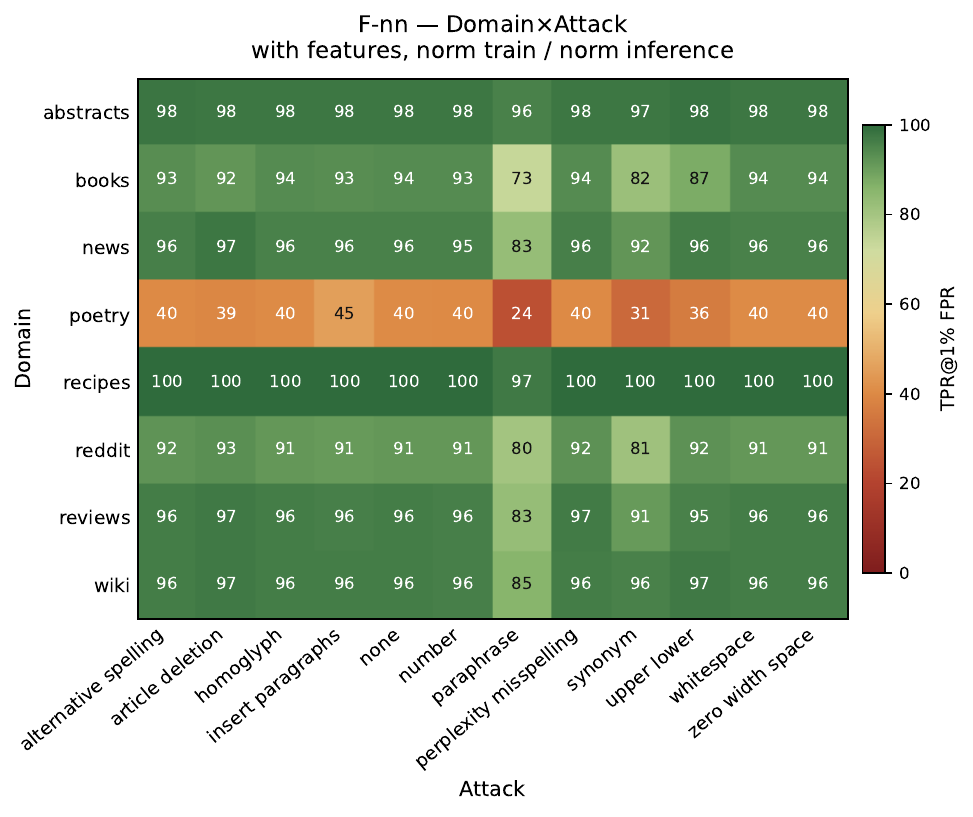}
\end{minipage}
\hfill
\begin{minipage}[t]{0.485\textwidth}\centering
\includegraphics[width=\linewidth,keepaspectratio]{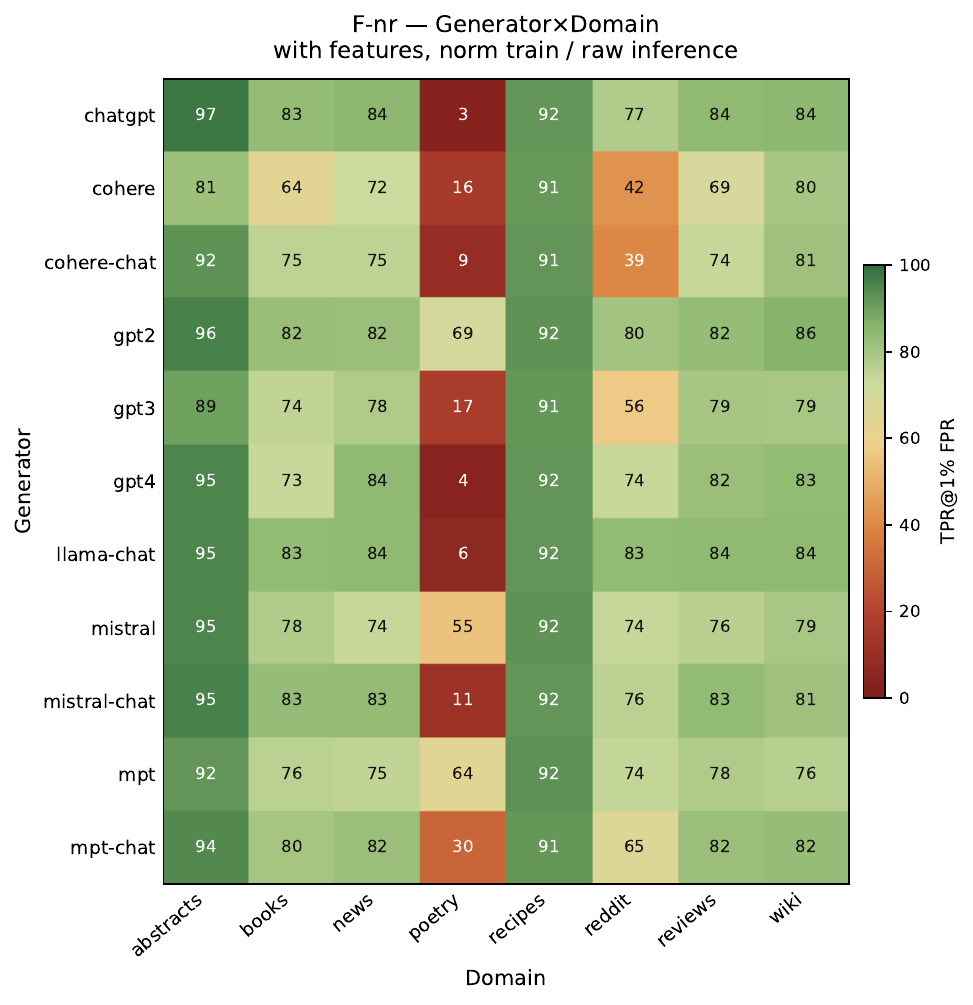}
\par\vspace{3pt}
\includegraphics[width=\linewidth,keepaspectratio]{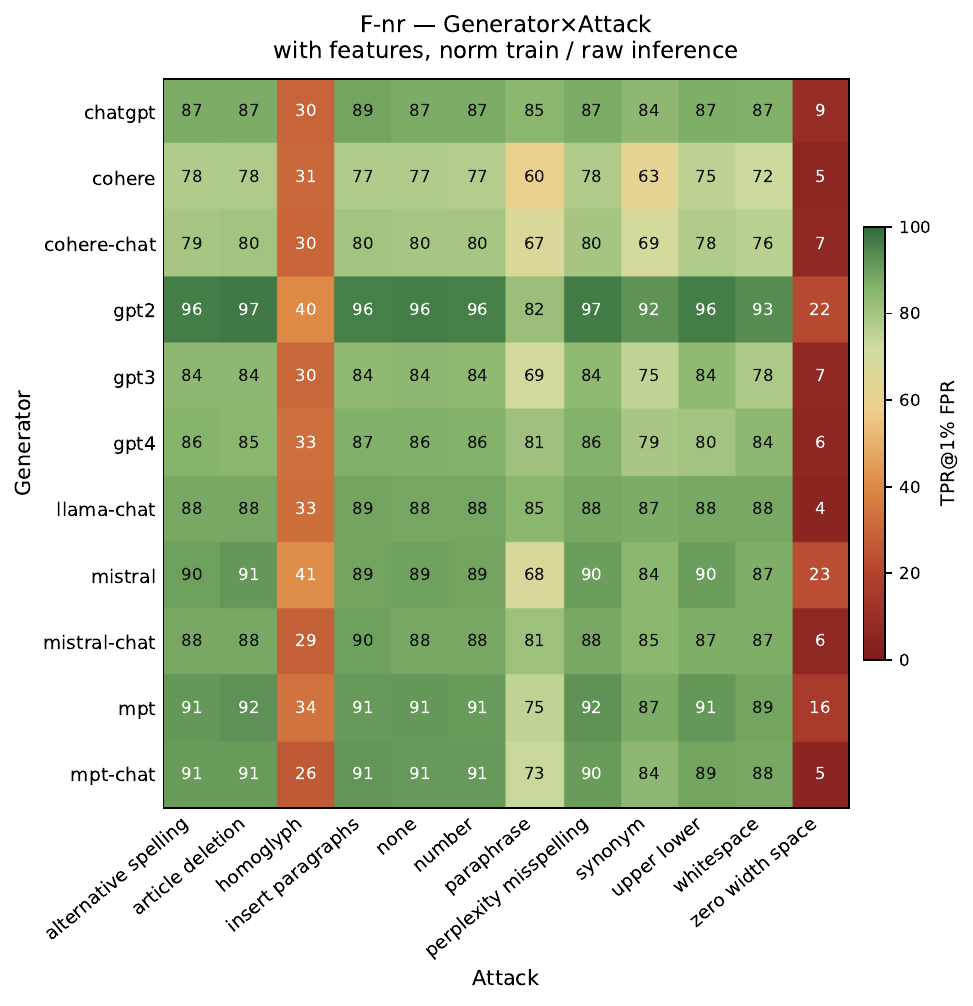}
\par\vspace{3pt}
\includegraphics[width=\linewidth,keepaspectratio]{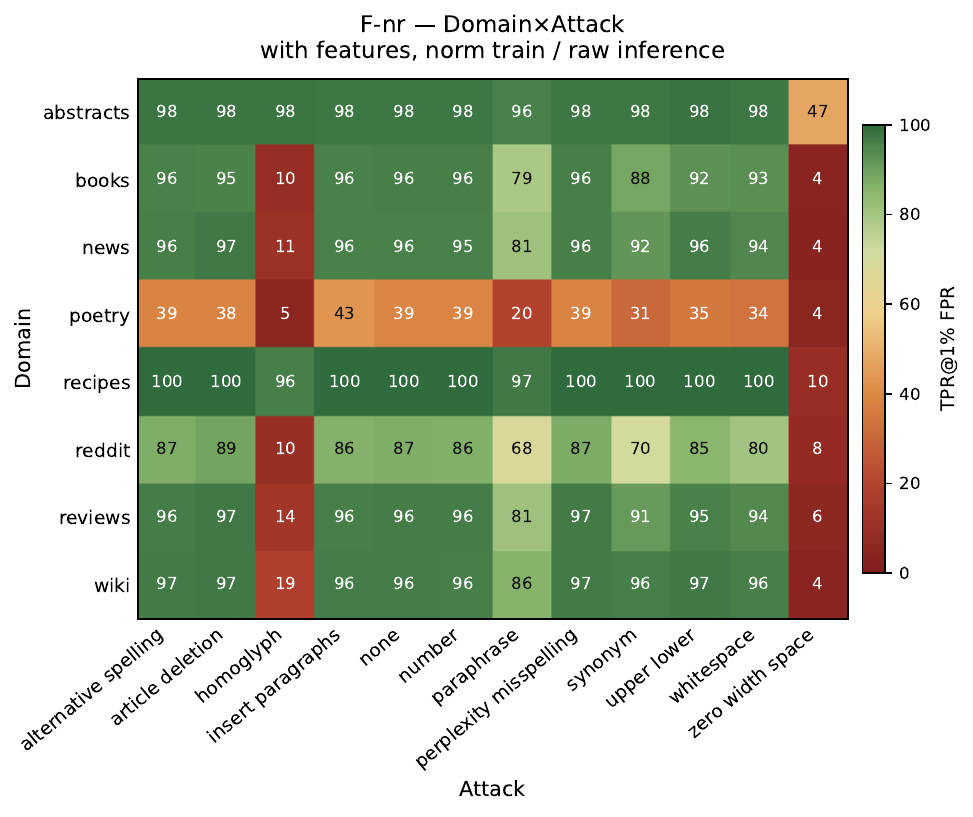}
\end{minipage}
\caption{Left: cell $+$F, norm$\to$norm. Right: cell $+$F, norm$\to$raw. Rows: generator$\times$domain (top), generator$\times$attack (middle), domain$\times$attack (bottom). TPR@1\,\% FPR.}
\label{fig:panelFnnFnr}
\end{figure*}

\begin{figure*}[p]
\centering
\begin{minipage}[t]{0.485\textwidth}\centering
\includegraphics[width=\linewidth,keepaspectratio]{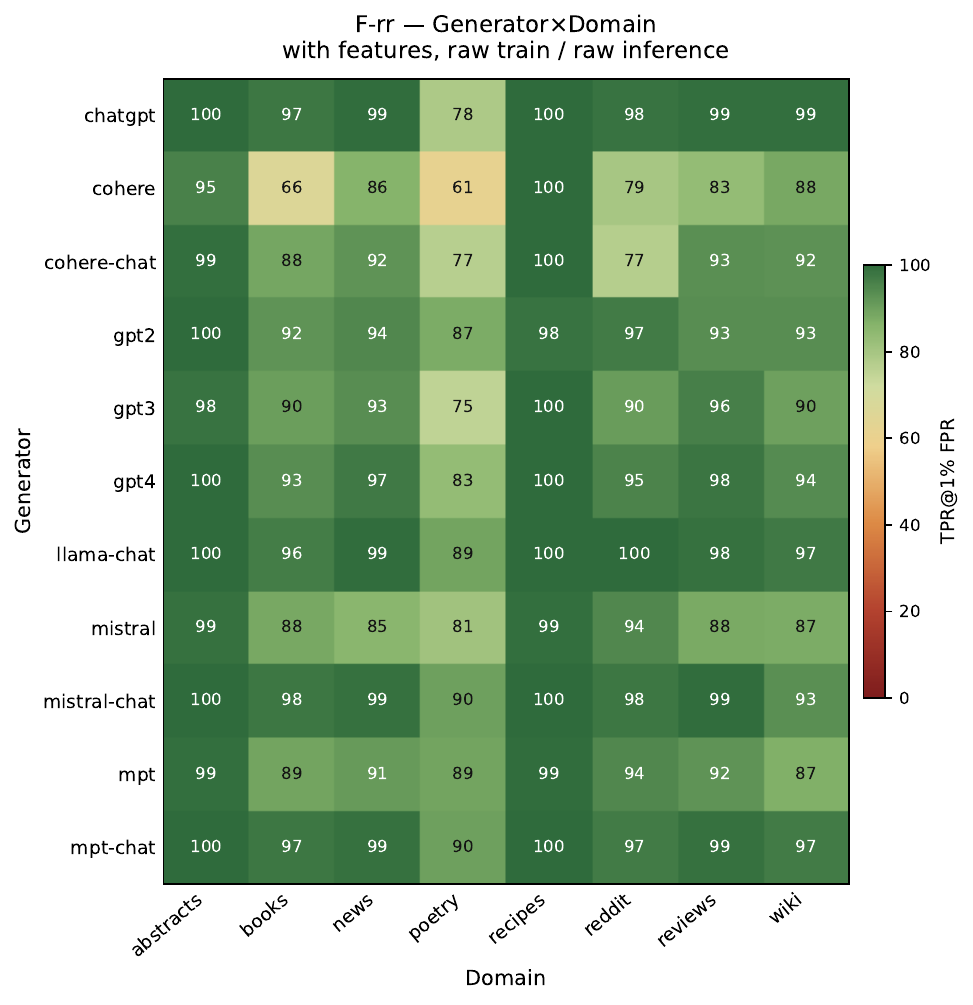}
\par\vspace{3pt}
\includegraphics[width=\linewidth,keepaspectratio]{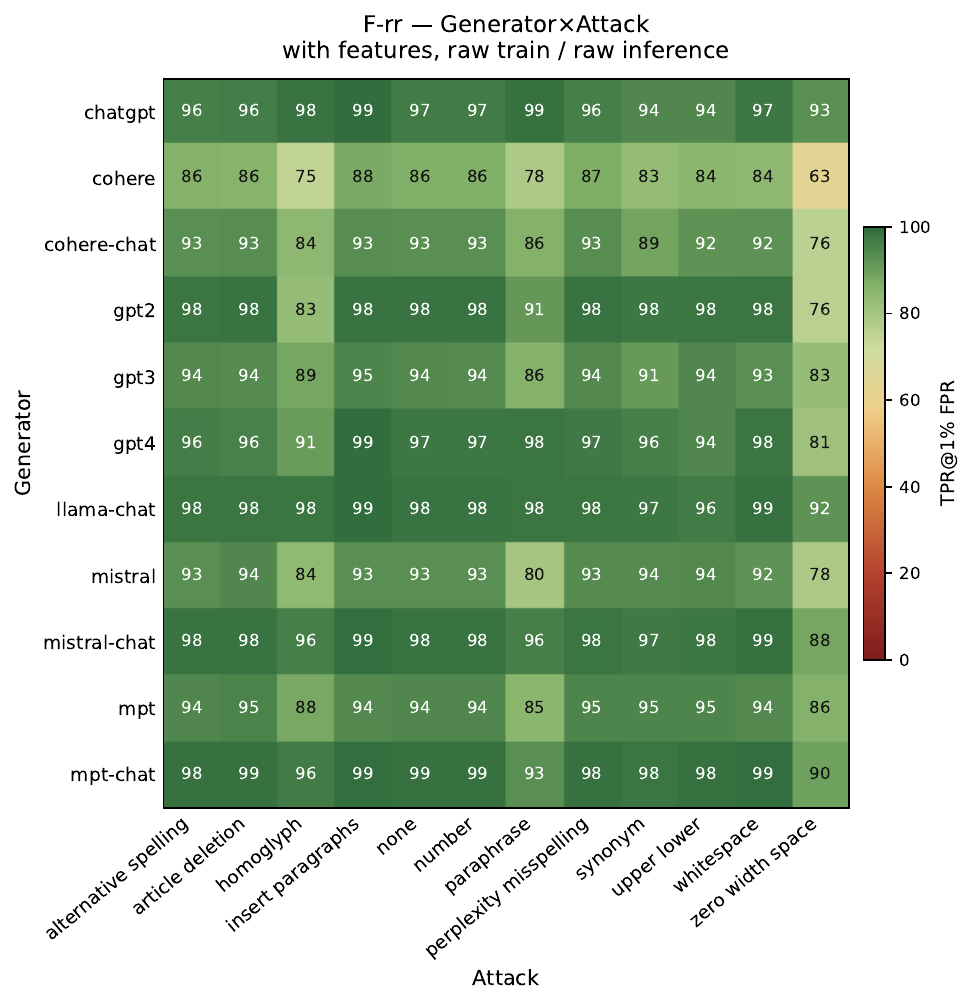}
\par\vspace{3pt}
\includegraphics[width=\linewidth,keepaspectratio]{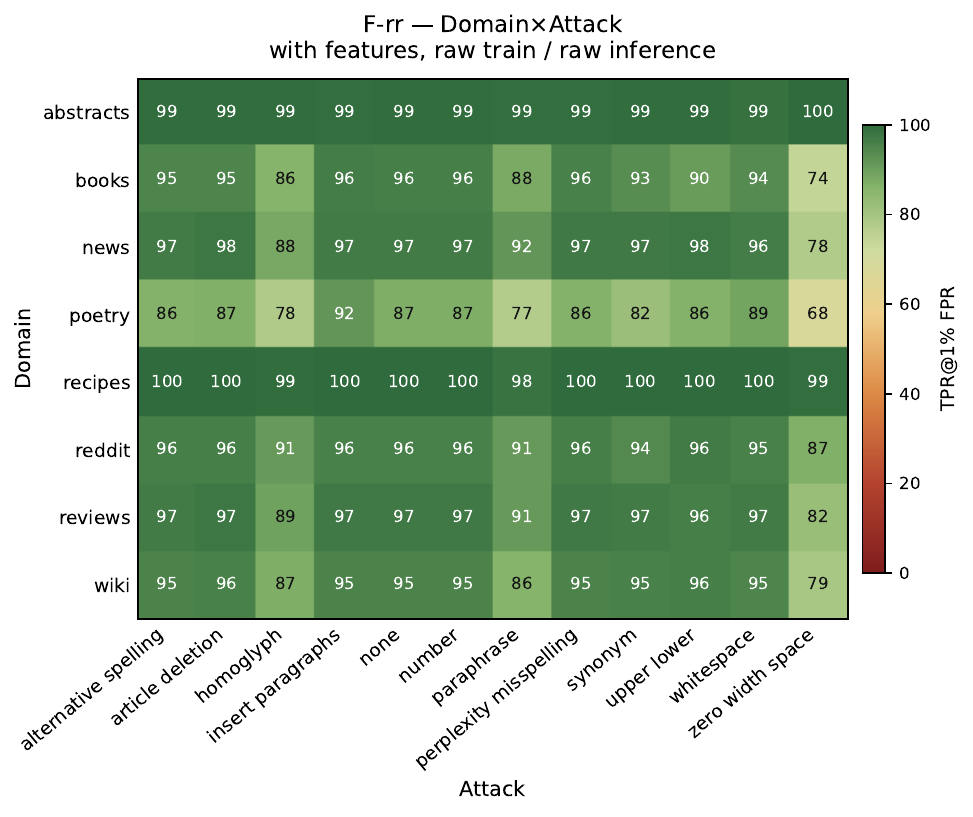}
\end{minipage}
\hfill
\begin{minipage}[t]{0.485\textwidth}\centering
\includegraphics[width=\linewidth,keepaspectratio]{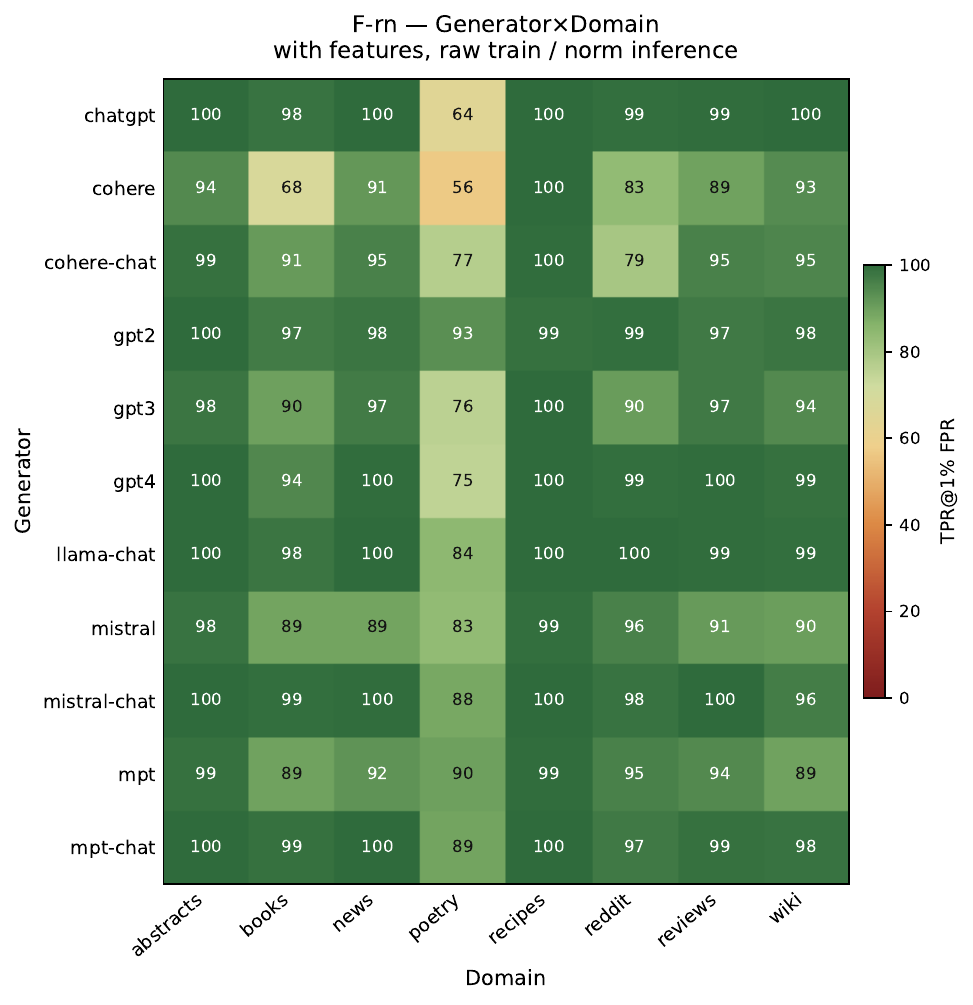}
\par\vspace{3pt}
\includegraphics[width=\linewidth,keepaspectratio]{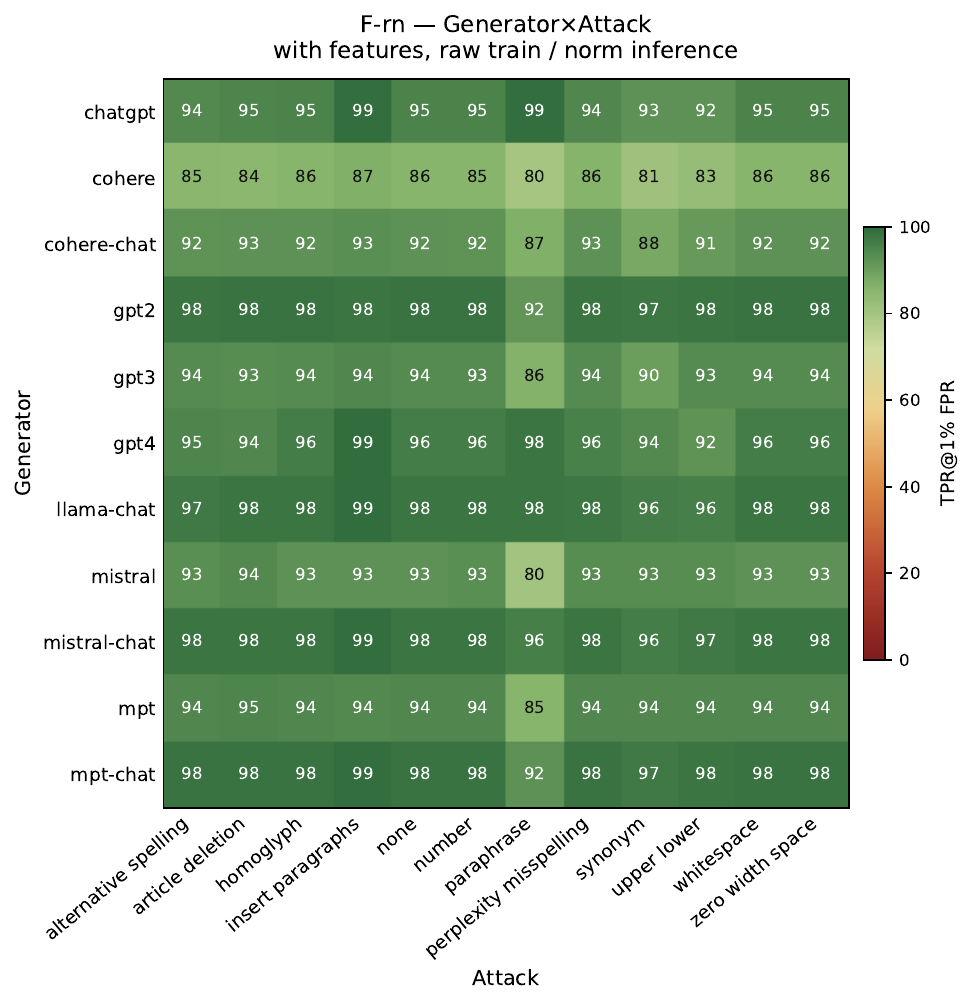}
\par\vspace{3pt}
\includegraphics[width=\linewidth,keepaspectratio]{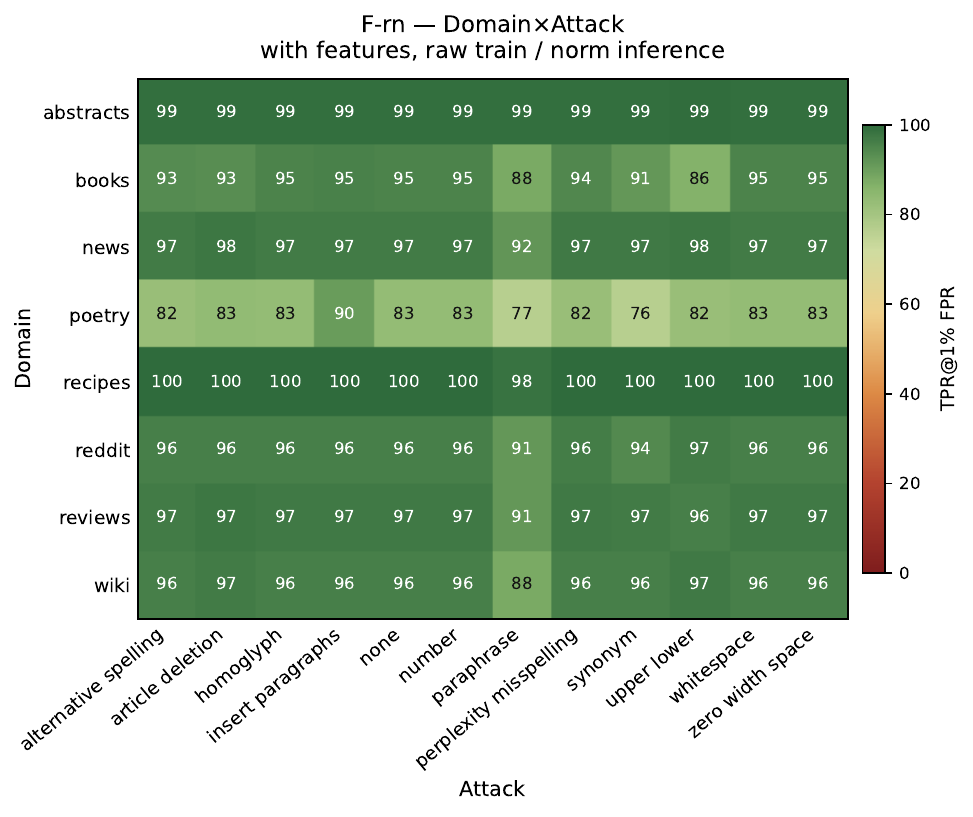}
\end{minipage}
\caption{Left: cell $+$F, raw$\to$raw. Right: cell $+$F, raw$\to$norm. Rows: generator$\times$domain (top), generator$\times$attack (middle), domain$\times$attack (bottom). TPR@1\,\% FPR.}
\label{fig:panelFrrFrn}
\end{figure*}

\begin{figure*}[p]
\centering
\begin{minipage}[t]{0.485\textwidth}\centering
\includegraphics[width=\linewidth,keepaspectratio]{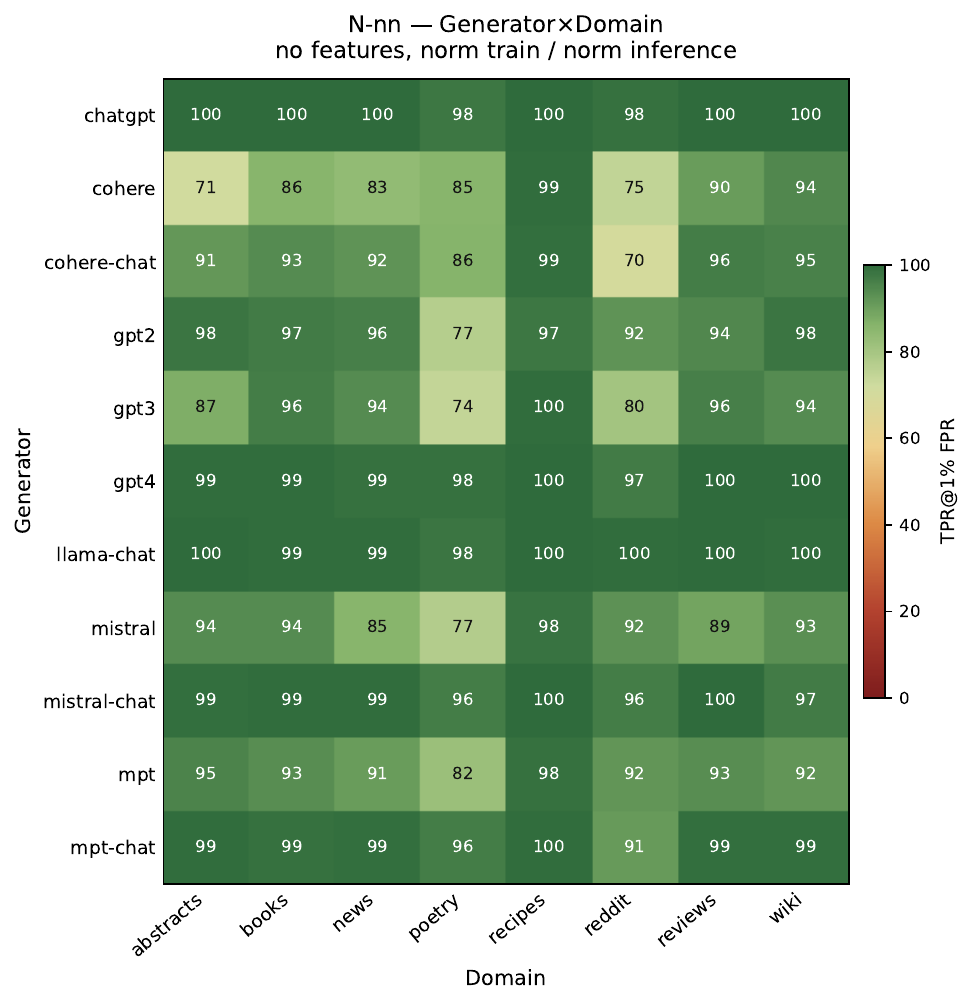}
\par\vspace{3pt}
\includegraphics[width=\linewidth,keepaspectratio]{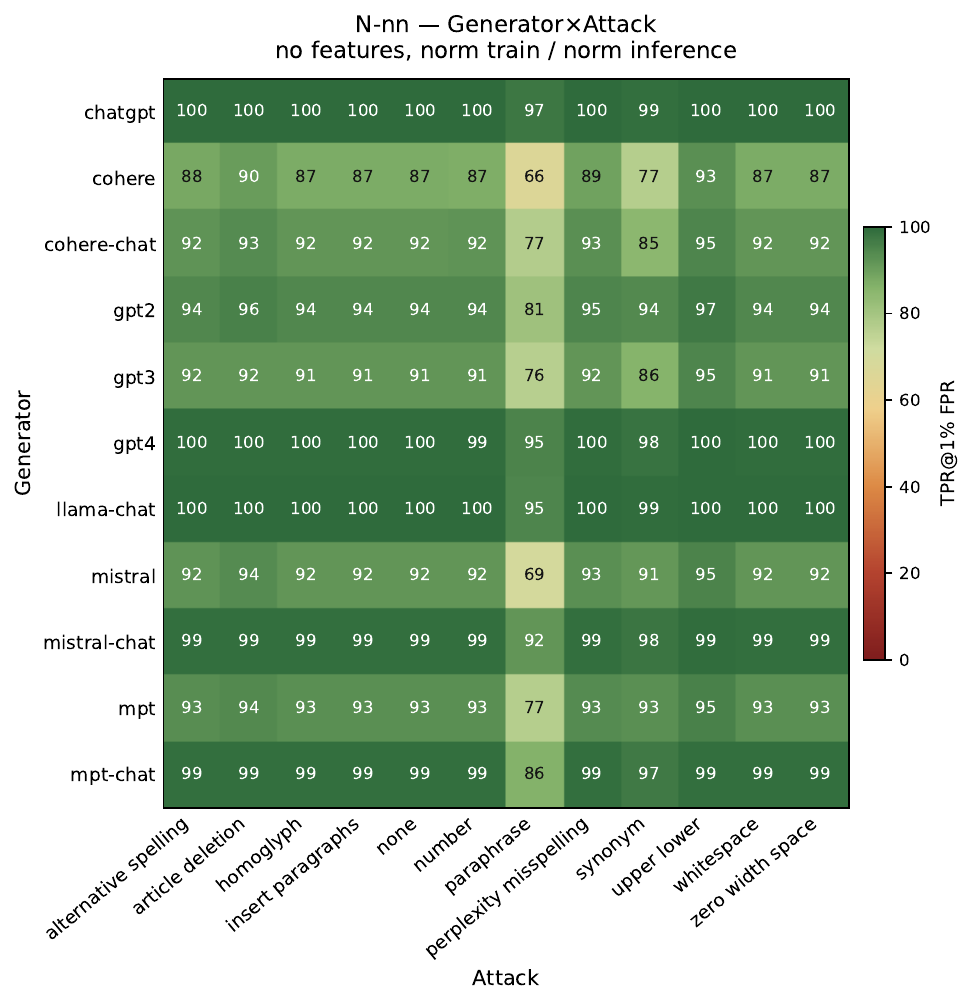}
\par\vspace{3pt}
\includegraphics[width=\linewidth,keepaspectratio]{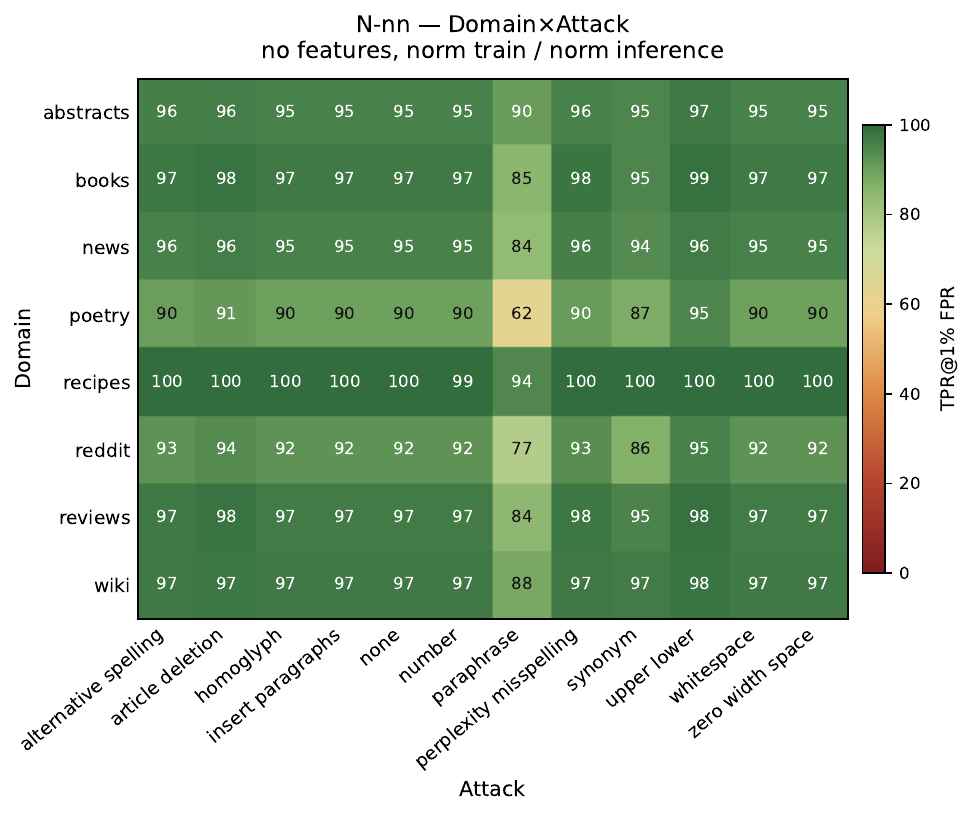}
\end{minipage}
\hfill
\begin{minipage}[t]{0.485\textwidth}\centering
\includegraphics[width=\linewidth,keepaspectratio]{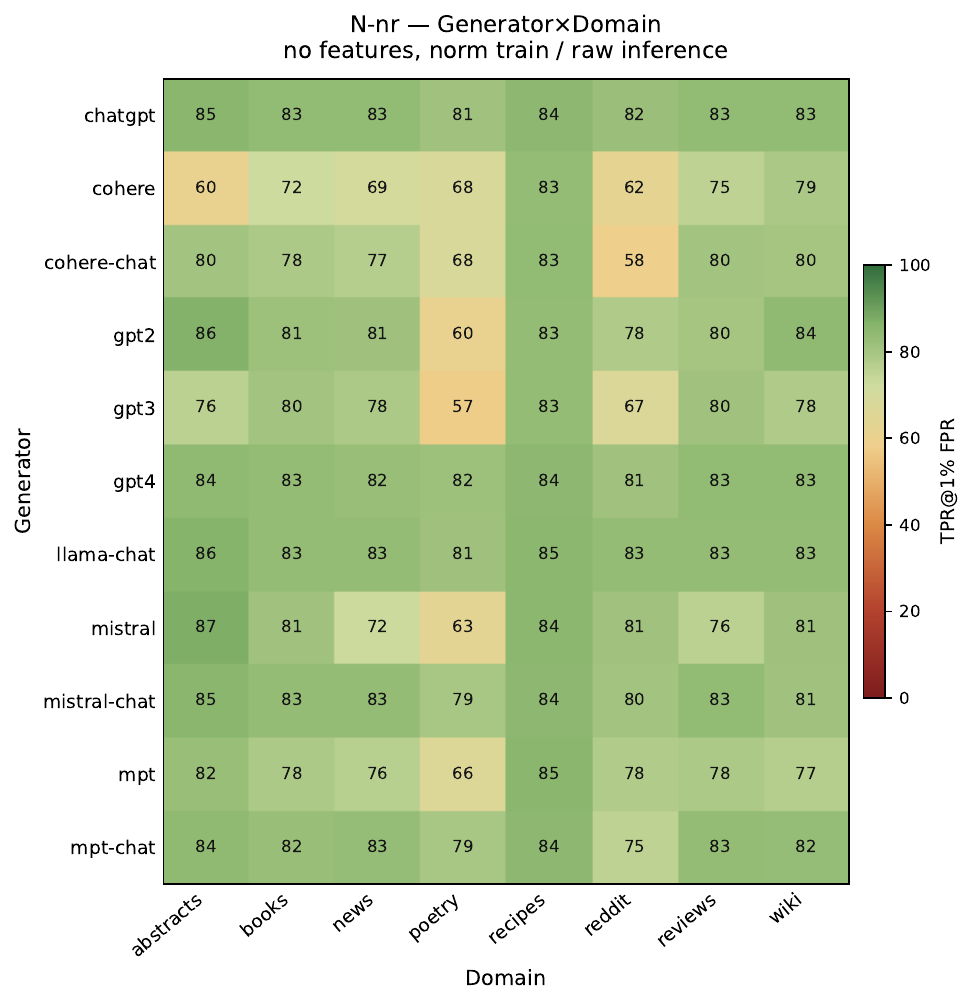}
\par\vspace{3pt}
\includegraphics[width=\linewidth,keepaspectratio]{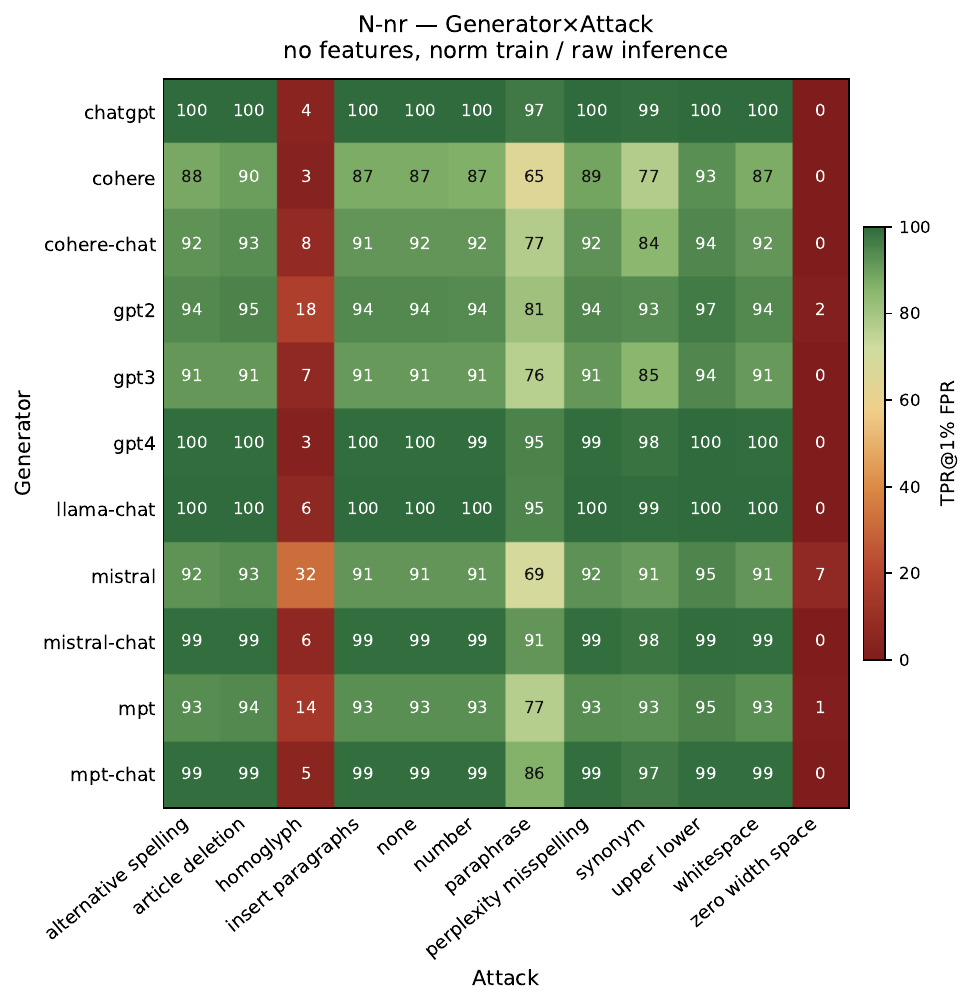}
\par\vspace{3pt}
\includegraphics[width=\linewidth,keepaspectratio]{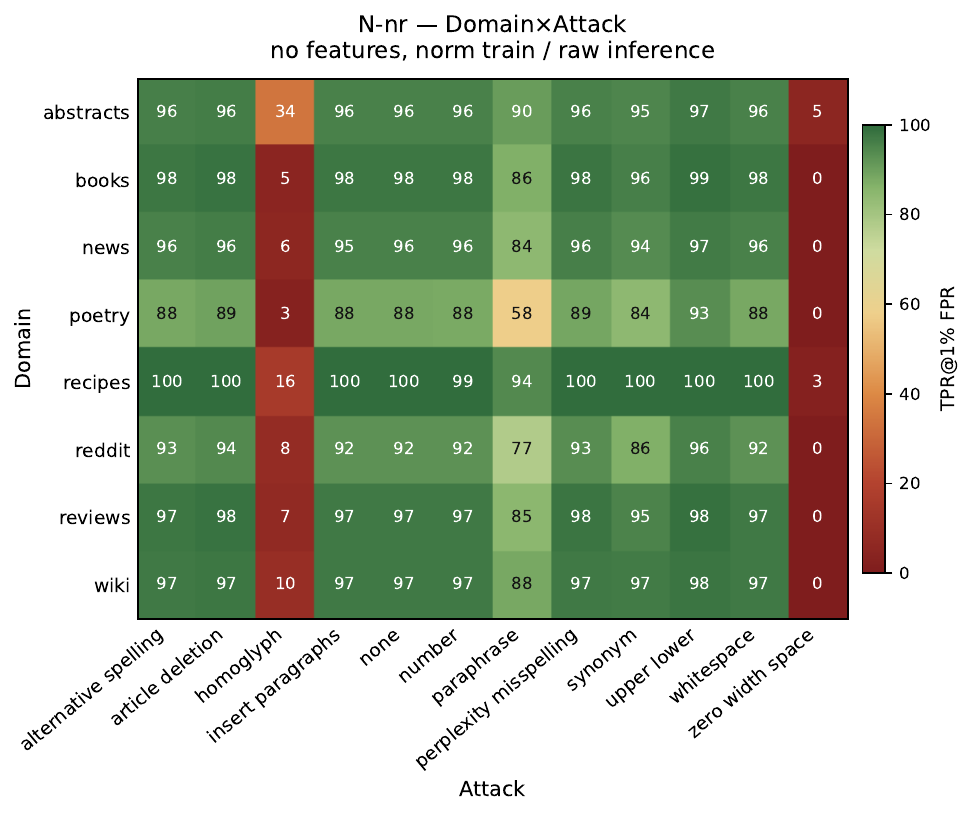}
\end{minipage}
\caption{Left: cell $-$F, norm$\to$norm. Right: cell $-$F, norm$\to$raw. Rows: generator$\times$domain (top), generator$\times$attack (middle), domain$\times$attack (bottom). TPR@1\,\% FPR.}
\label{fig:panelNnnNnr}
\end{figure*}

\begin{figure*}[p]
\centering
\begin{minipage}[t]{0.485\textwidth}\centering
\includegraphics[width=\linewidth,keepaspectratio]{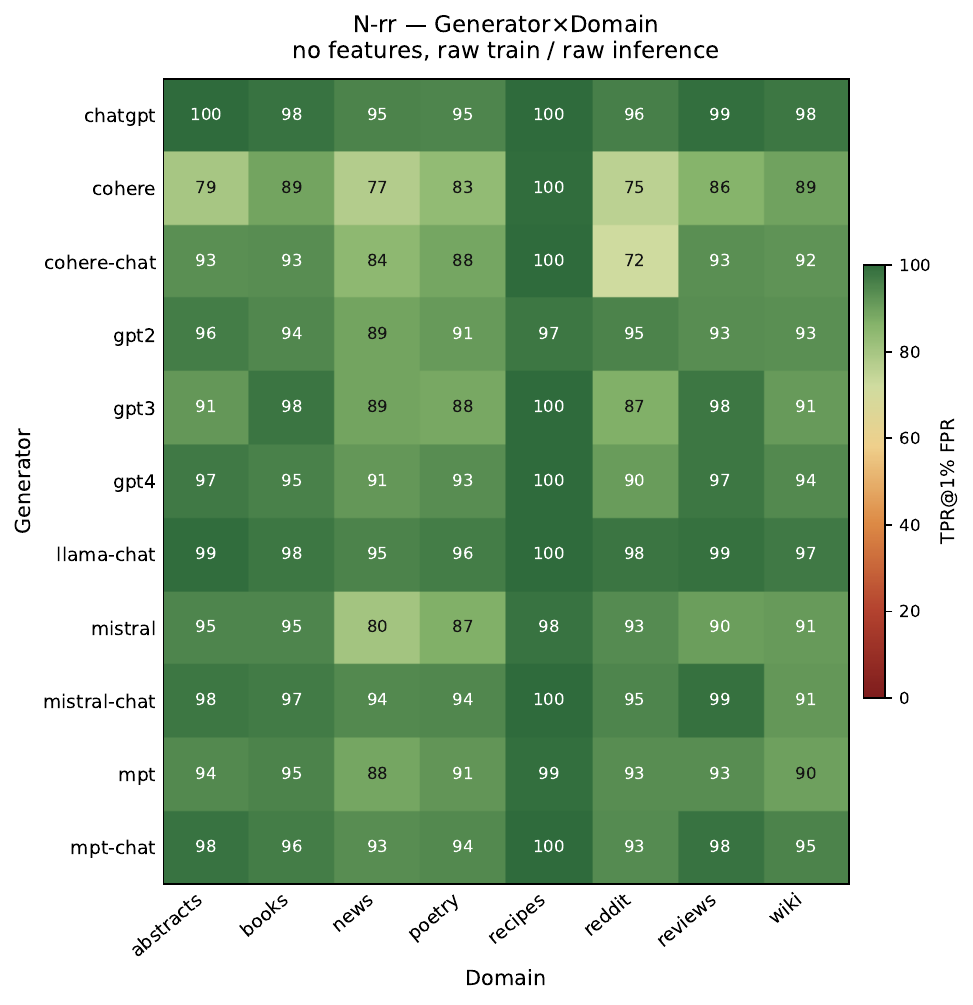}
\par\vspace{3pt}
\includegraphics[width=\linewidth,keepaspectratio]{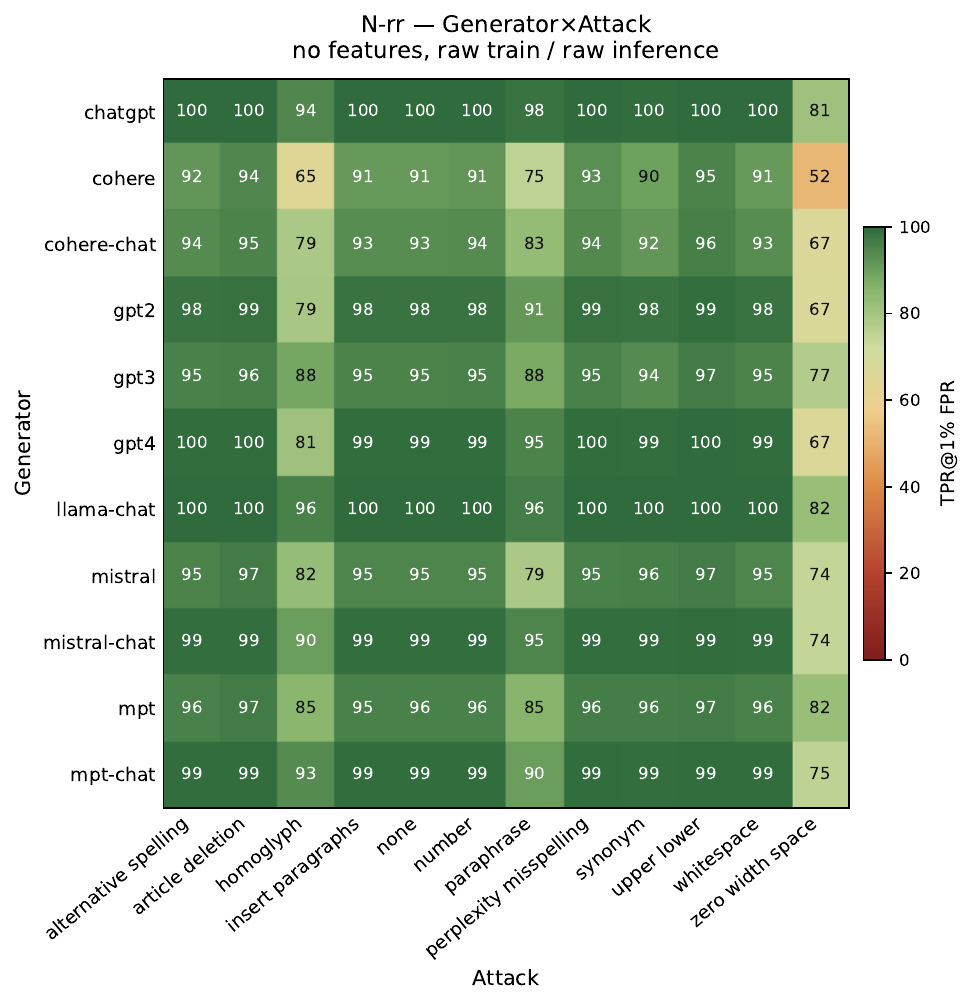}
\par\vspace{3pt}
\includegraphics[width=\linewidth,keepaspectratio]{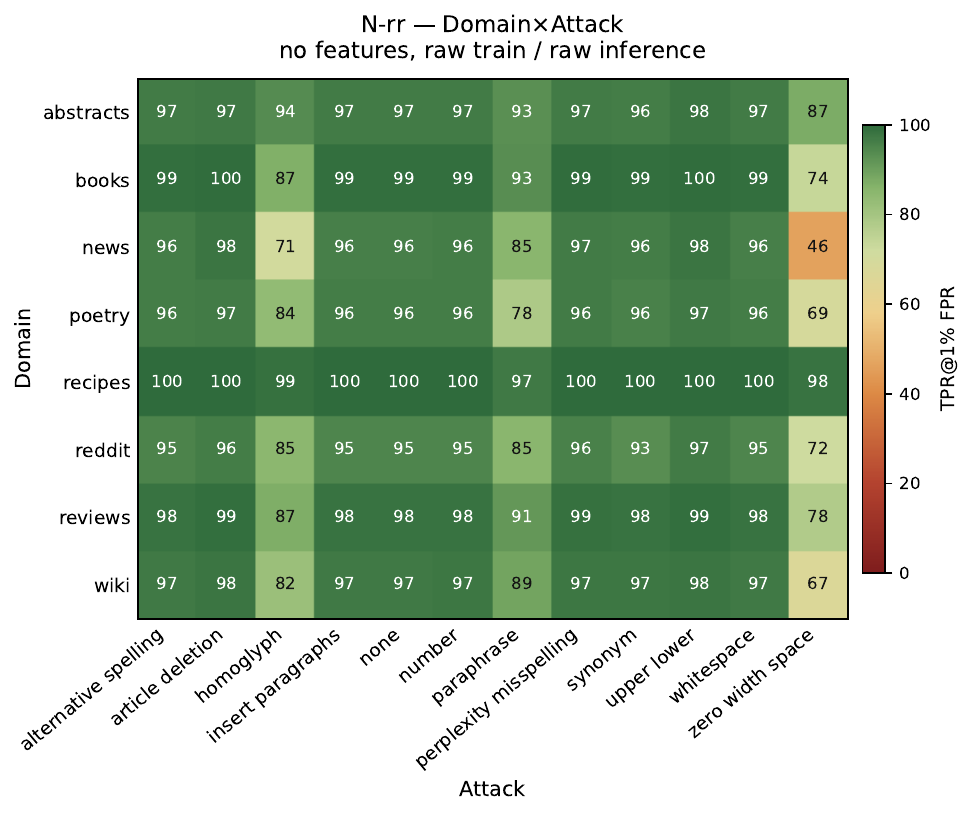}
\end{minipage}
\hfill
\begin{minipage}[t]{0.485\textwidth}\centering
\includegraphics[width=\linewidth,keepaspectratio]{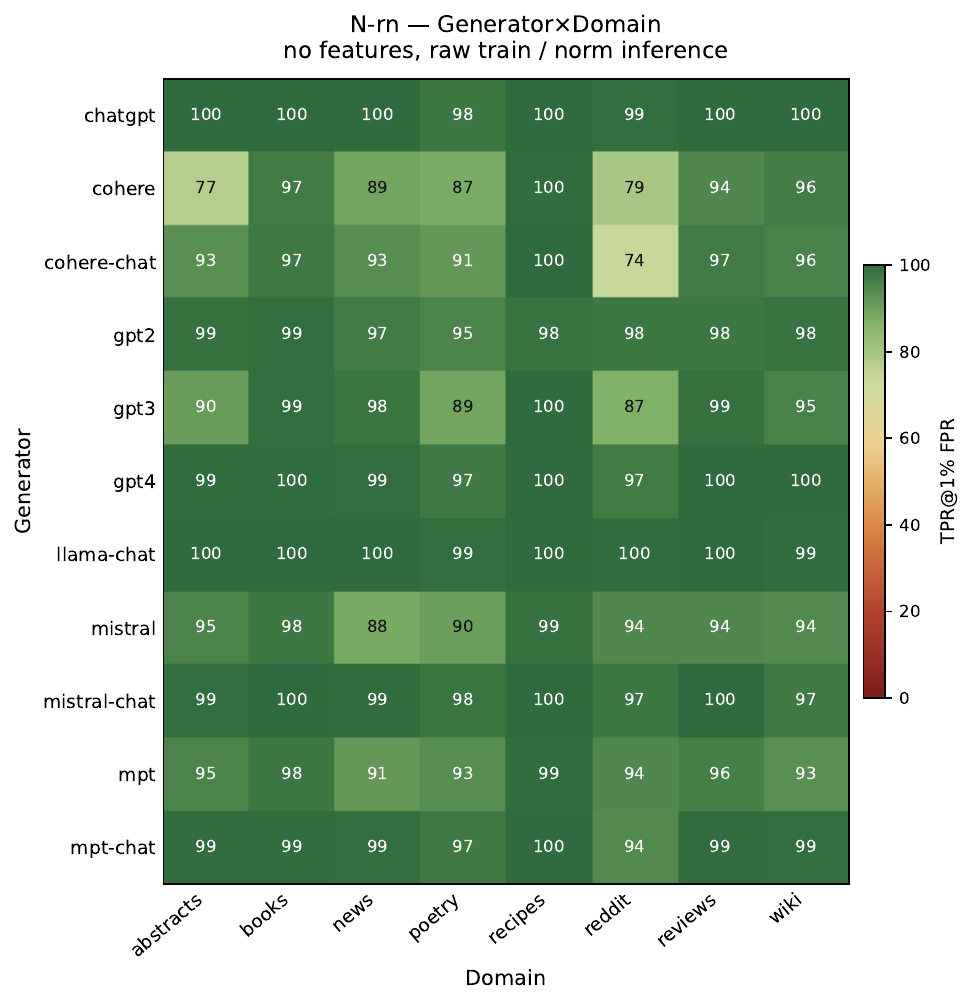}
\par\vspace{3pt}
\includegraphics[width=\linewidth,keepaspectratio]{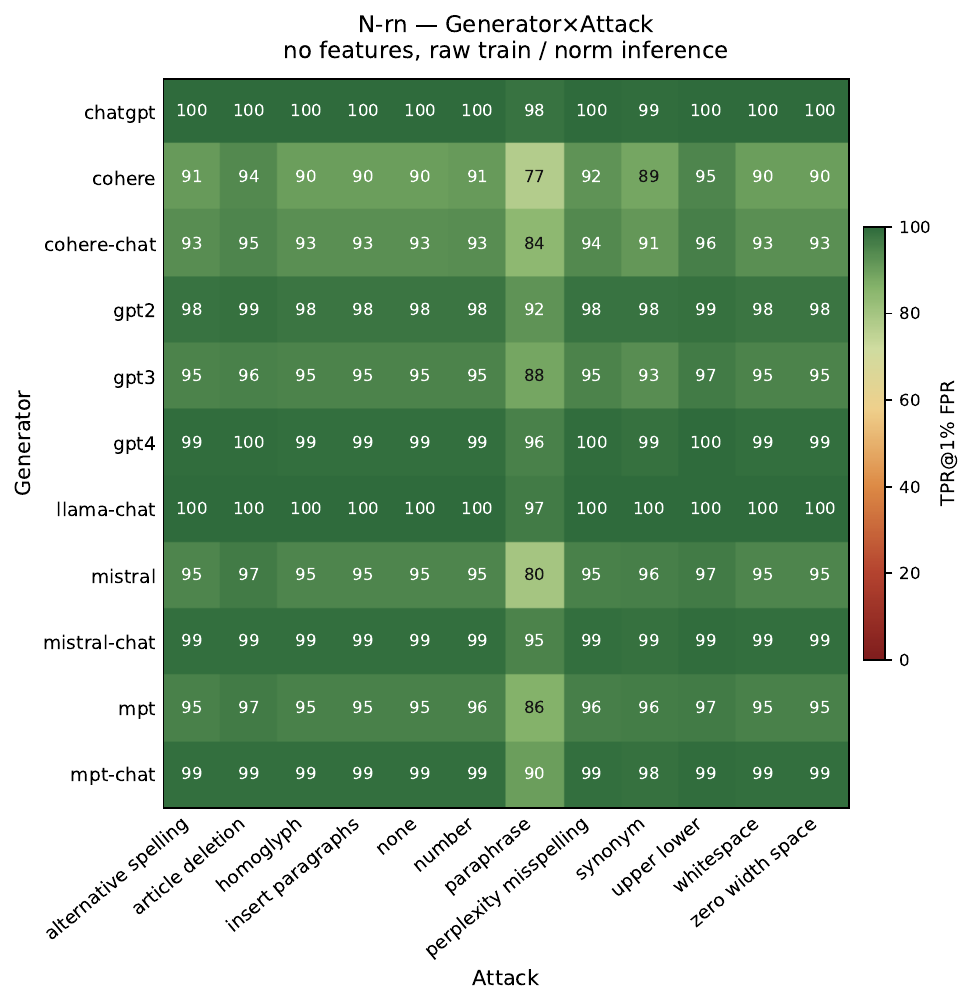}
\par\vspace{3pt}
\includegraphics[width=\linewidth,keepaspectratio]{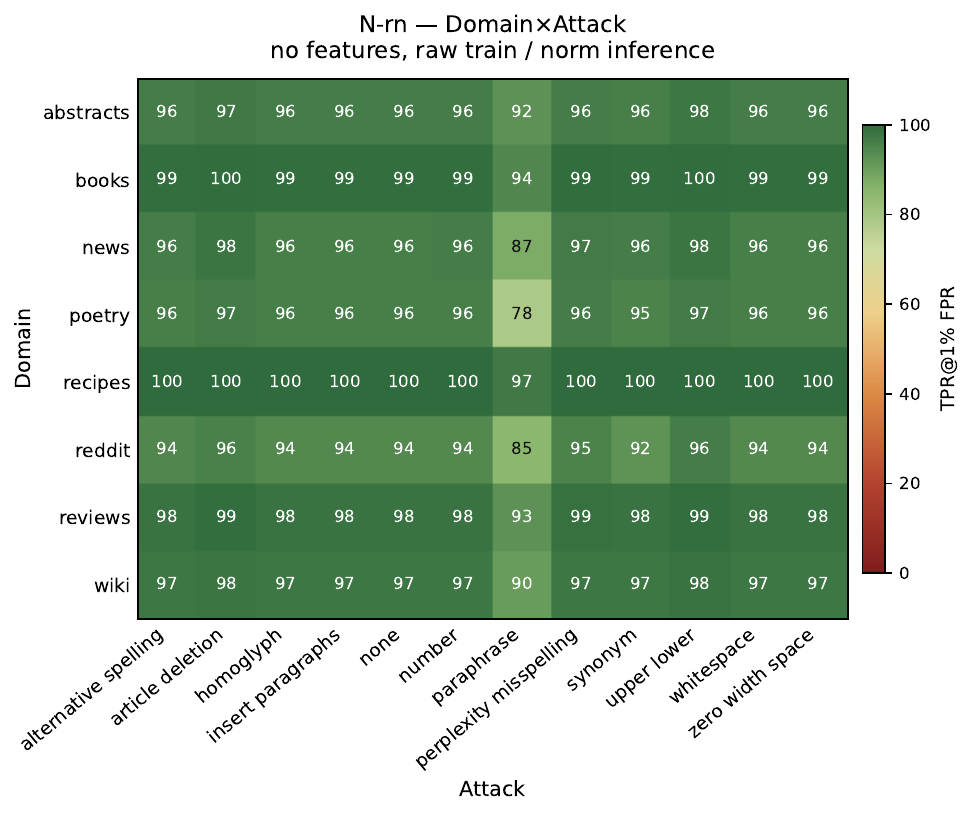}
\end{minipage}
\caption{Left: cell $-$F, raw$\to$raw. Right: cell $-$F, raw$\to$norm. Rows: generator$\times$domain (top), generator$\times$attack (middle), domain$\times$attack (bottom). TPR@1\,\% FPR.}
\label{fig:panelNrrNrn}
\end{figure*}

\begin{figure*}[p]
\centering
\begin{minipage}[t]{0.485\textwidth}\centering
\includegraphics[width=\linewidth,keepaspectratio]{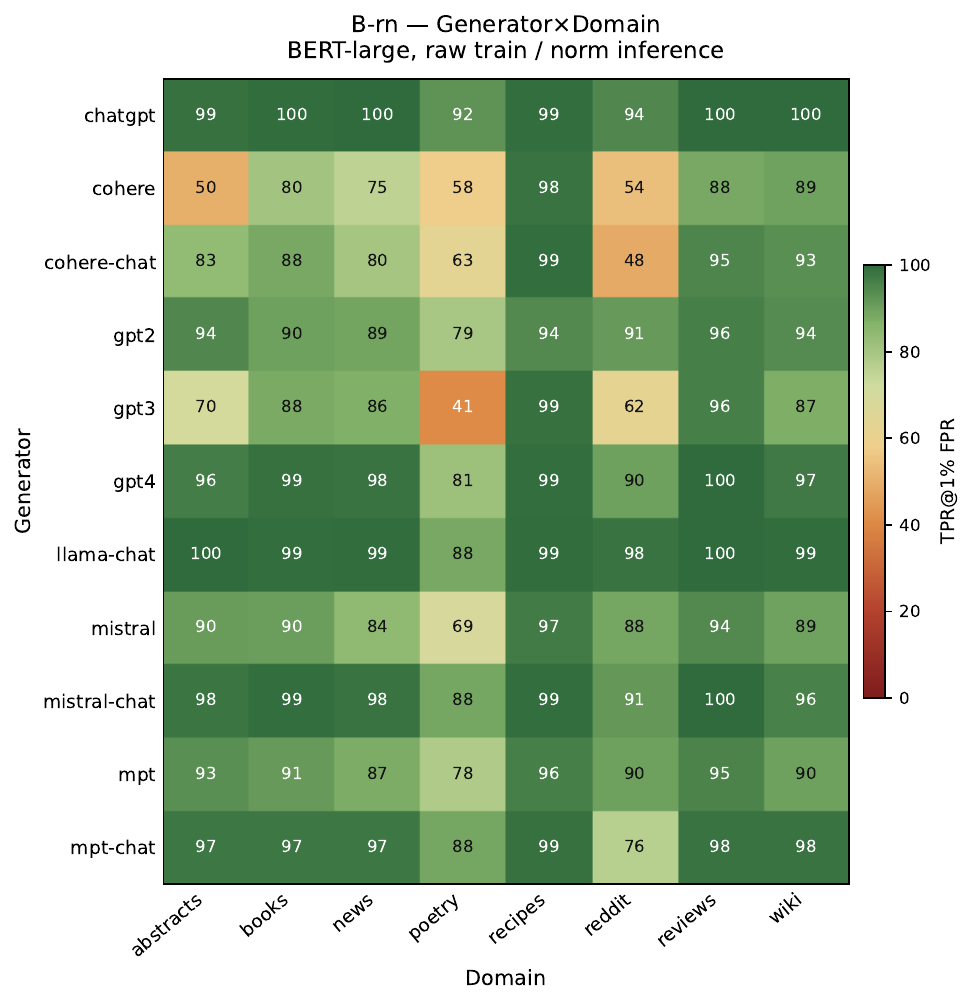}
\par\vspace{3pt}
\includegraphics[width=\linewidth,keepaspectratio]{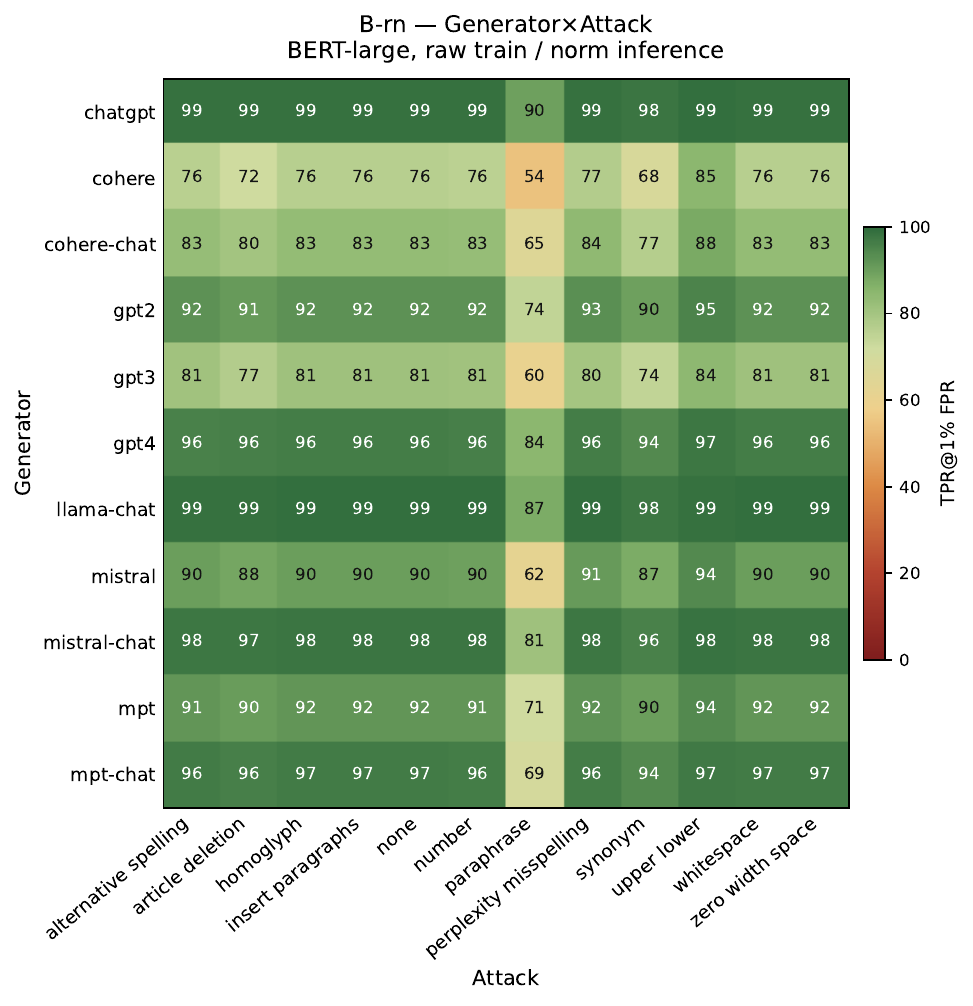}
\par\vspace{3pt}
\includegraphics[width=\linewidth,keepaspectratio]{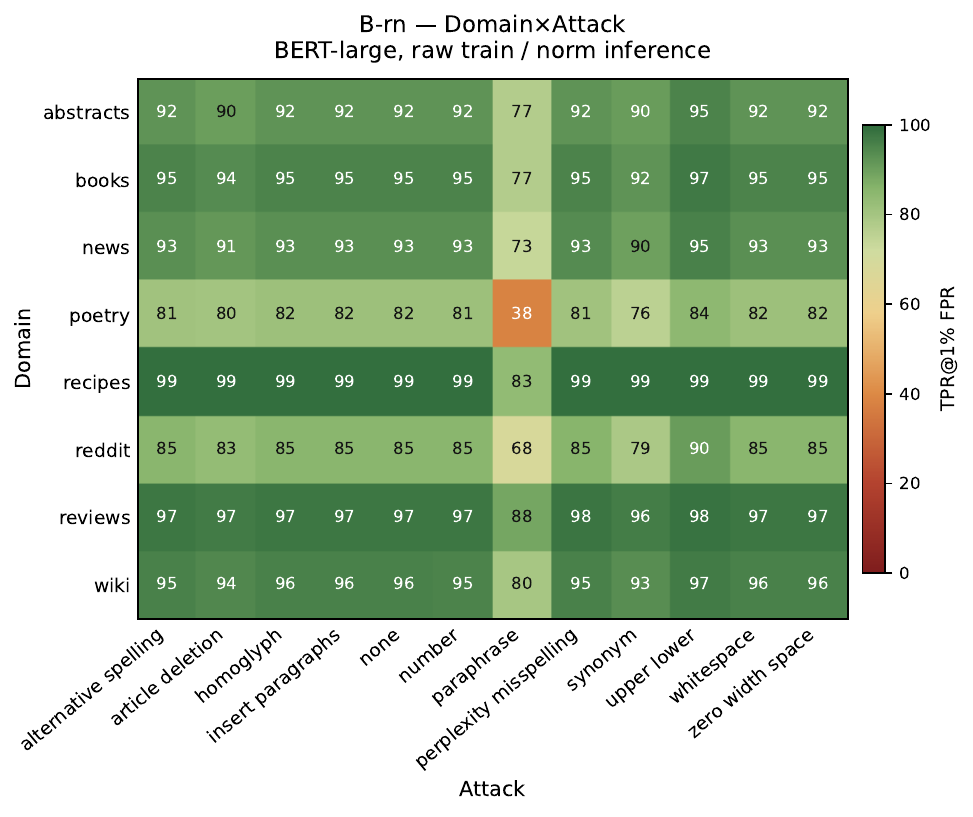}
\end{minipage}
\hfill
\begin{minipage}[t]{0.485\textwidth}\centering
\includegraphics[width=\linewidth,keepaspectratio]{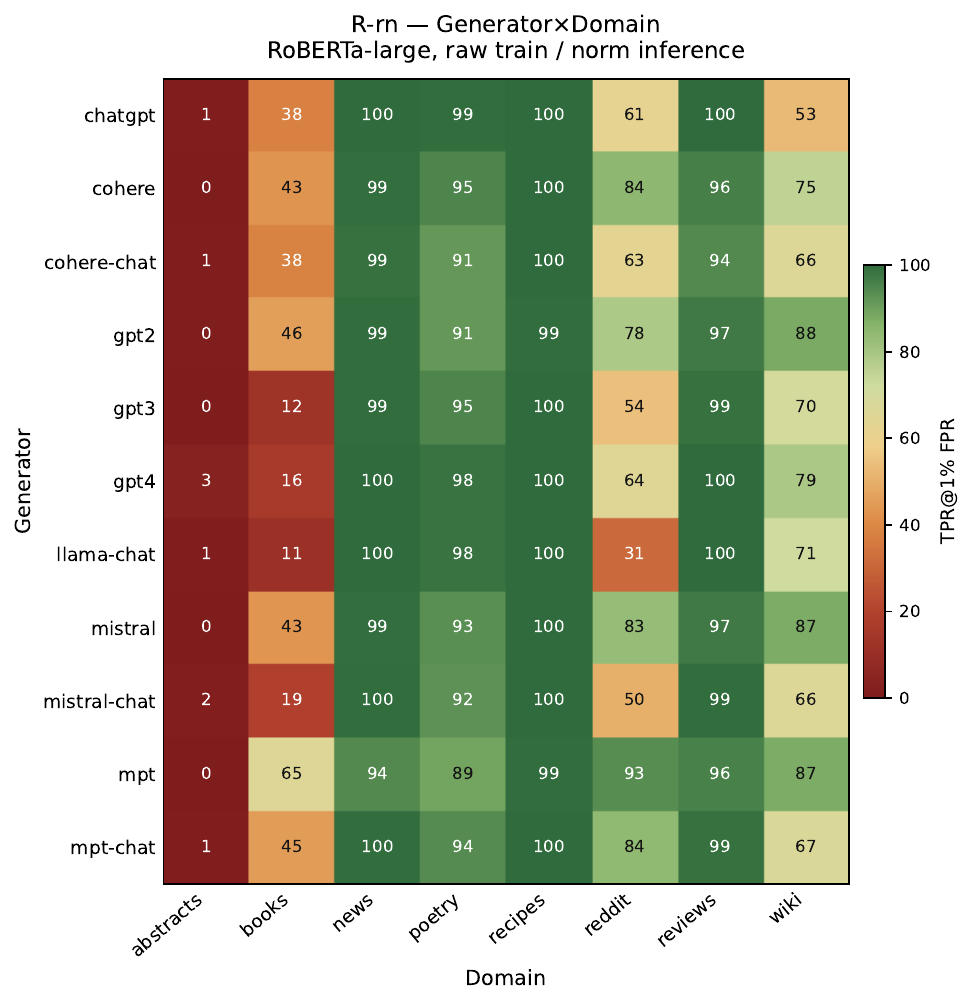}
\par\vspace{3pt}
\includegraphics[width=\linewidth,keepaspectratio]{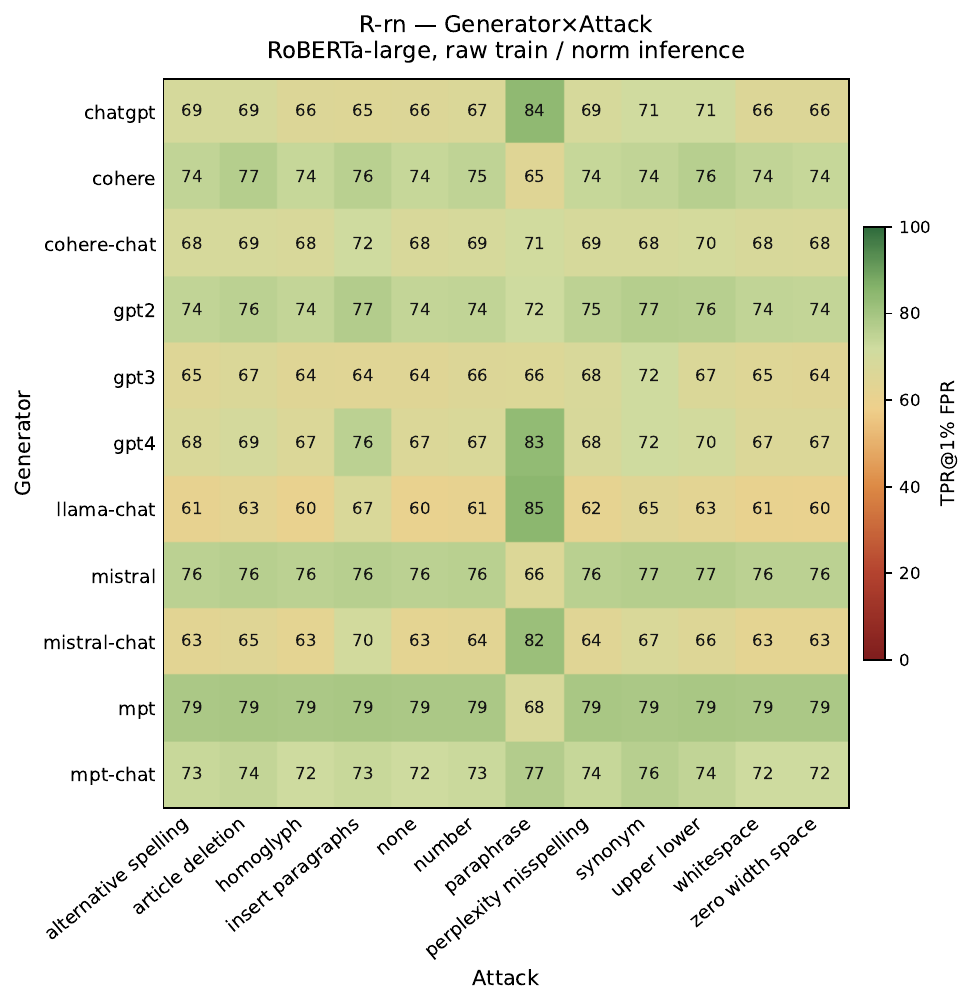}
\par\vspace{3pt}
\includegraphics[width=\linewidth,keepaspectratio]{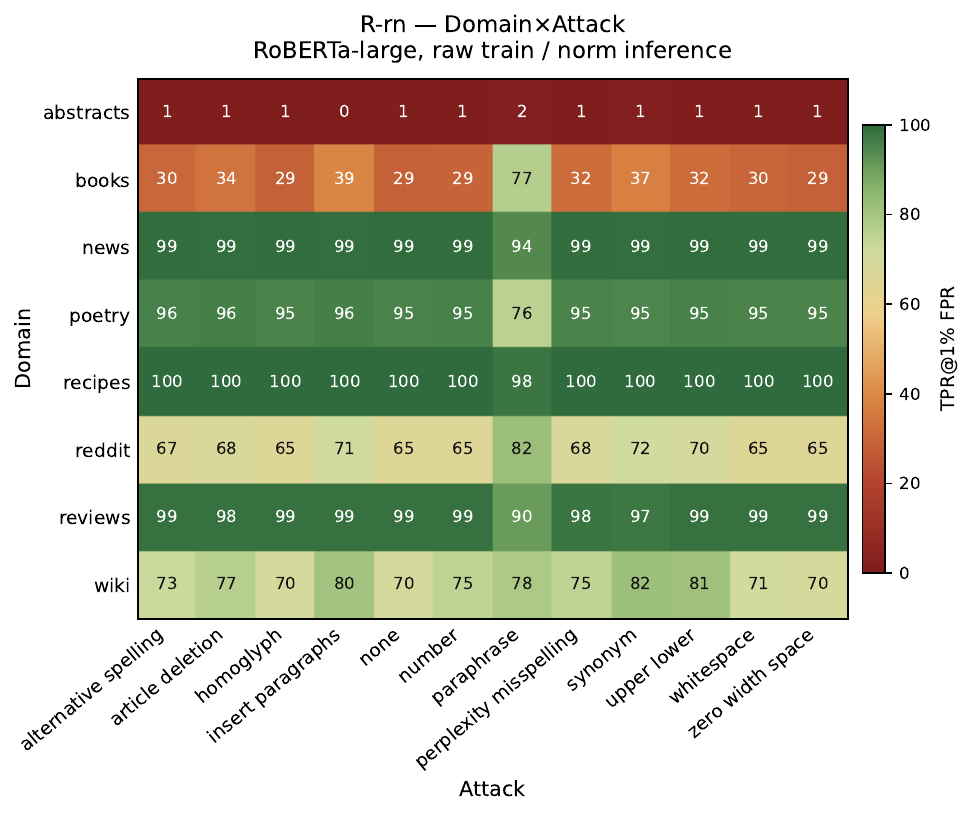}
\end{minipage}
\caption{Left: cell BERT, raw$\to$norm. Right: cell RoBERTa, raw$\to$norm. Rows: generator$\times$domain (top), generator$\times$attack (middle), domain$\times$attack (bottom). TPR@1\,\% FPR.}
\label{fig:panelBrnRrn}
\end{figure*}

\begin{figure*}[p]
\centering
\includegraphics[width=0.96\textwidth,keepaspectratio]{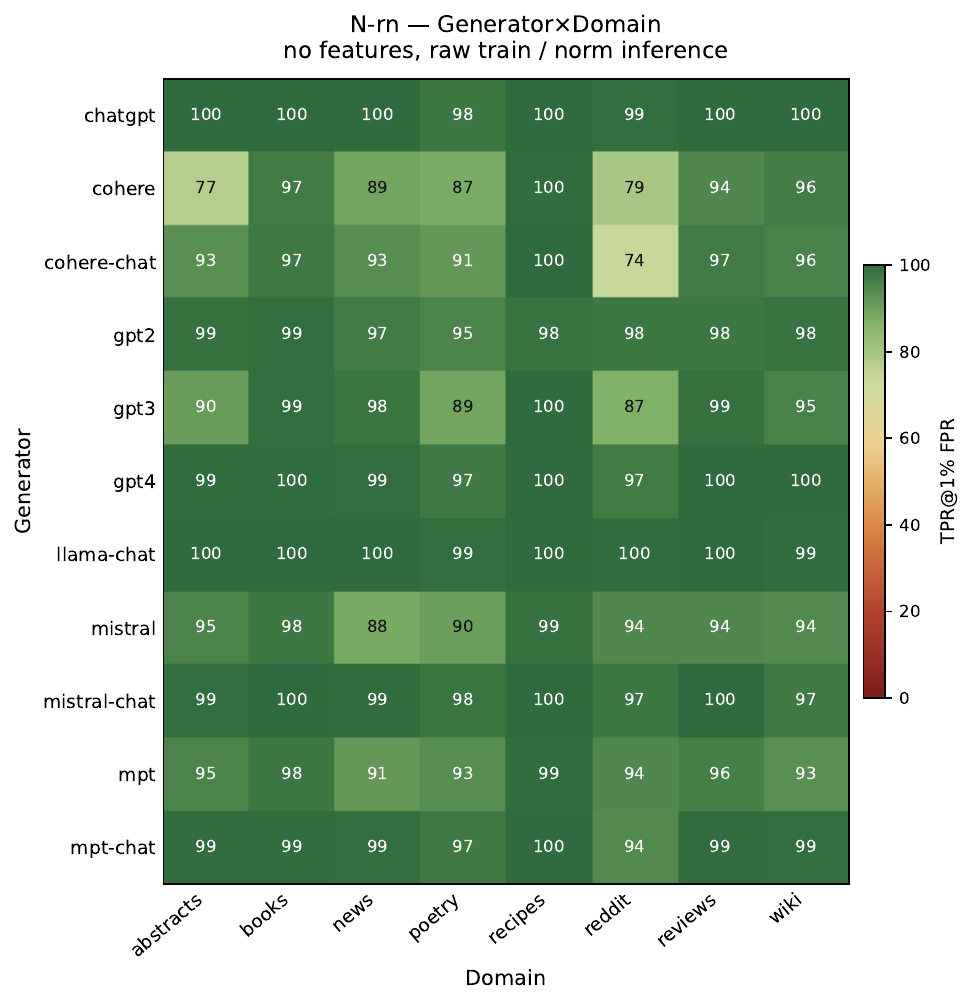}
\caption{The configuration we report ($-$F, raw$\to$norm) at full size: generator$\times$domain, TPR@1\,\% FPR.}
\label{fig:panelNrn_gen_dom}
\end{figure*}

\begin{figure*}[p]
\centering
\includegraphics[width=0.96\textwidth,keepaspectratio]{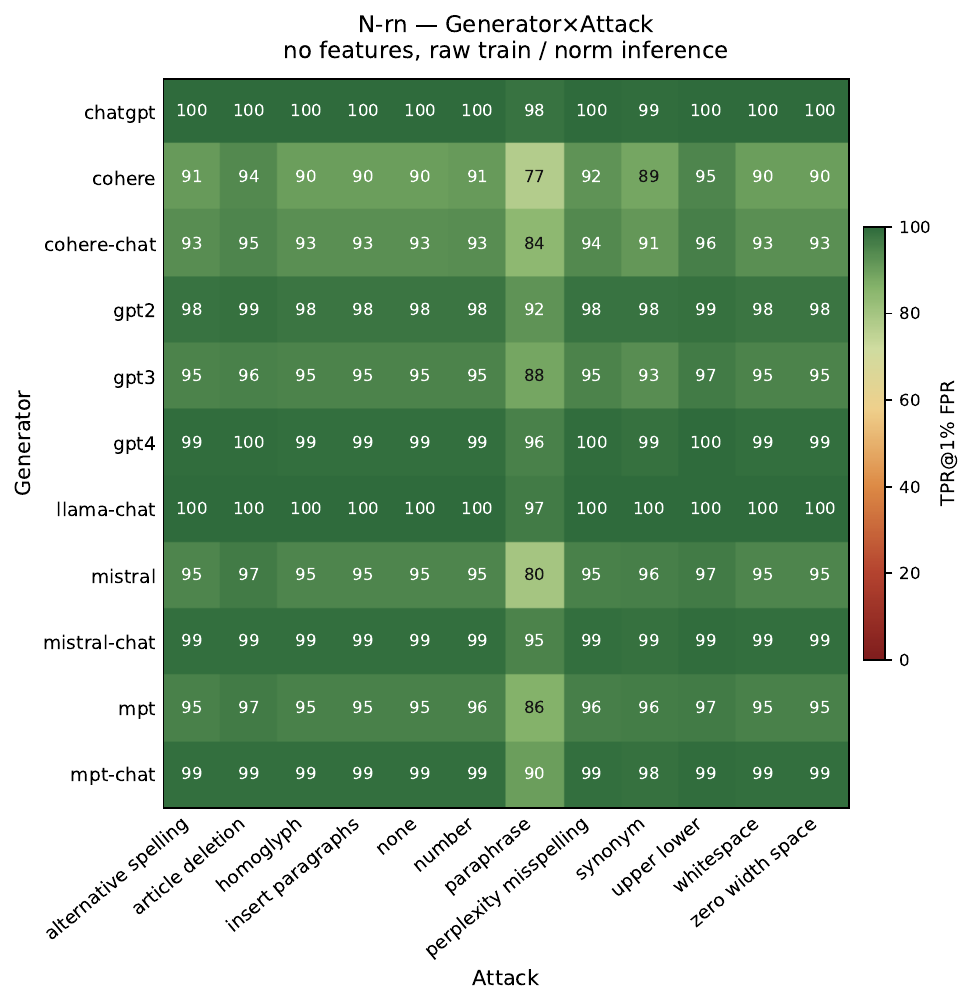}
\caption{The configuration we report ($-$F, raw$\to$norm) at full size: generator$\times$attack, TPR@1\,\% FPR.}
\label{fig:panelNrn_gen_atk}
\end{figure*}

\begin{figure*}[p]
\centering
\includegraphics[width=0.96\textwidth,keepaspectratio]{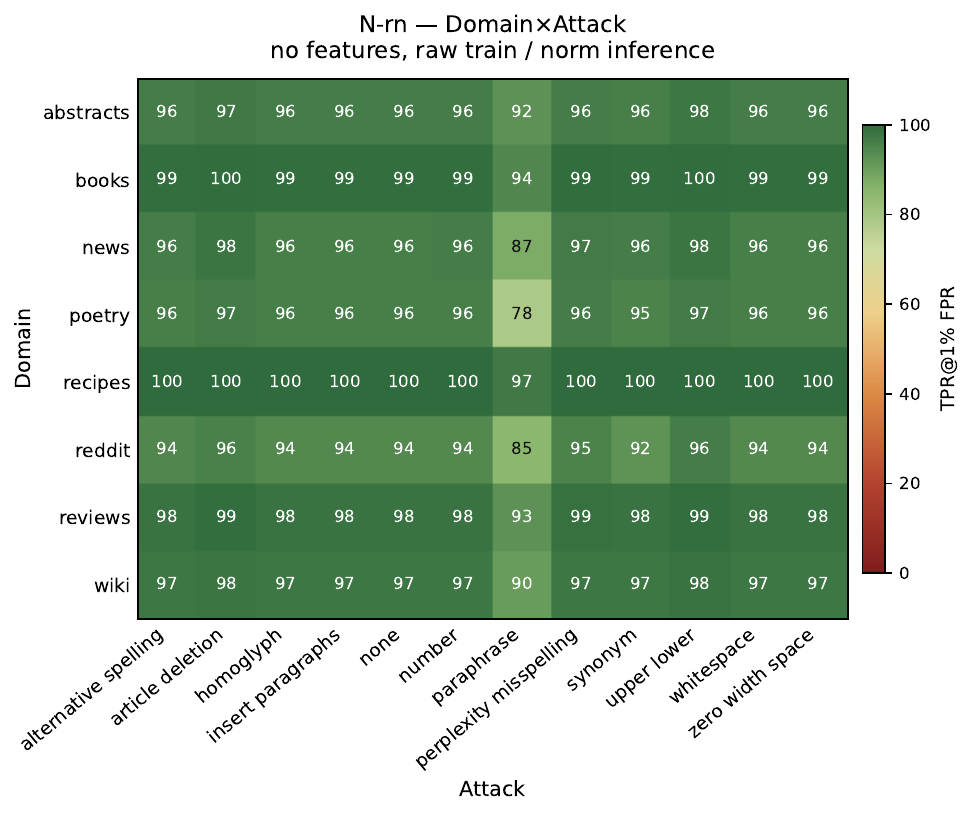}
\caption{The configuration we report ($-$F, raw$\to$norm) at full size: domain$\times$attack, TPR@1\,\% FPR.}
\label{fig:panelNrn_dom_atk}
\end{figure*}

\newpage

\end{document}